\documentclass[twocolumn
]{article}

\usepackage{graphicx}\usepackage{dcolumn}\usepackage{bm}
\usepackage[utf8]{inputenc}
\usepackage[T1]{fontenc}
\usepackage{mathptmx}
\usepackage{etoolbox}
\usepackage[utf8]{inputenc} \usepackage[T1]{fontenc}    \usepackage{hyperref}       \usepackage{url}            \usepackage{booktabs}       \usepackage{amsfonts}       \usepackage{amsmath}
\usepackage{amssymb}
\usepackage{amsthm}
\usepackage{dutchcal}
\usepackage{nicefrac}       \usepackage{microtype}      \usepackage{xspace}
\usepackage[dvipsnames]{xcolor}
\usepackage{colortbl}
\usepackage{hhline}
\usepackage{bbm}
\usepackage{tabularx}
\newcolumntype{e}{>{\hsize=.40\hsize}X}
\newcolumntype{E}{>{\hsize=.20\hsize}X}

\usepackage[nameinlink]{cleveref}
\usepackage[caption=false]{subfig}
\usepackage{tikz}
\usepackage[export]{adjustbox}
\usepackage{authblk}
\usepackage{blindtext}
\usepackage{placeins}

\usepackage{breakurl}

\usepackage[linesnumbered,ruled,vlined]{algorithm2e}

\usepackage{upgreek}
\newenvironment{conditions*}
  {\par\vspace{\abovedisplayskip}\noindent
   \tabularx{\columnwidth}{>{$}l<{$} @{${}:{}$}
   >{\raggedright\arraybackslash}X}} {\endtabularx\par\vspace{\belowdisplayskip}}

\Crefname{algocf}{Alg.}{Algs.}
\Crefname{algocf}{Alg.}{Algs.}
\Crefname{prop}{Prop.}{Props.}
\Crefname{table}{Table}{Tables}

\SetCommentSty{mycommfont}

\SetAlFnt{\sffamily}

\makeatletter
\renewcommand{\algocf@Vline}[1]{\strut\par\nointerlineskip\algocf@push{\skiprule}\hbox{\bgroup\color{cyan}\vrule\egroup\vtop{\algocf@push{\skiptext}\vtop{\algocf@addskiptotal #1}\bgroup\color{cyan}\Hlne\egroup}}\vskip\skiphlne\algocf@pop{\skiprule}\nointerlineskip}\renewcommand{\algocf@Vsline}[1]{\strut\par\nointerlineskip\algocf@bblockcode\algocf@push{\skiprule}\hbox{\bgroup\color{cyan}\vrule\egroup\vtop{\algocf@push{\skiptext}\vtop{\algocf@addskiptotal #1}}}\algocf@pop{\skiprule}\algocf@eblockcode}
\makeatother

\DeclareMathOperator*{\argmin}{argmin}
\DeclareMathOperator*{\argmax}{argmax}

\SetKwProg{Fn}{Function}{}{}
\SetKwInOut{Parameter}{Parameters}
\SetKw{KwVar}{Var}
\SetKw{KwSet}{Set}
\SetKw{KwList}{List}
\SetKw{KwInit}{Init:}
\SetKw{KwParams}{Params:}
\SetKw{KwTuple}{Tuple}
\SetKw{KwBreak}{break}
\SetKw{KwNot}{Not}

\newtheorem{prop}{Proposition}

\makeatletter
\newcommand{\myref}[1]{\Cref{#1}\mynameref{#1}{\csname r@#1\endcsname}}
\newcommand{\Myref}[1]{\Cref{#1}\mynameref{#1}{\csname r@#1\endcsname}}

\def\mynameref#1#2{\begingroup
    \edef\@mytxt{#2}\edef\@mytst{\expandafter\@thirdoffive\@mytxt}\ifx\@mytst\empty\else
    \space(\nameref{#1})\fi
  \endgroup
}
\makeatother

\newcommand\vmath[1]{\ensuremath{#1}\xspace}

\newcommand\ts[1]{\vmath{\mathrm{\MakeUppercase{#1}}}}

\newcommand{\matr}[1]{\MakeUppercase{\mathbf\mathit\bm{#1}}\protect}     
\newcommand\tv[1]{\vmath{\pmb{\MakeLowercase{#1}}}}

\newcommand{\norm}[1]{\left\lVert#1\right\rVert}
\newcommand{\target}[1]{\vmath{{{#1}^{*}}}}
\newcommand{\nnc}[1]{\vmath{\hat{#1}}}

\newcommand{\iu}{{i\mkern1mu}}

\newcolumntype{R}[1]{>{\hsize=#1\hsize\raggedright\arraybackslash}X}
\newcolumntype{L}[1]{>{\hsize=#1\hsize\raggedleft\arraybackslash}X}
\newcolumntype{Y}[1]{>{\hsize=#1\hsize\arraybackslash}X}

\newcommand{\approptoinn}[2]{\mathrel{\vcenter{
  \offinterlineskip\halign{\hfil$##$\cr
    #1\propto\cr\noalign{\kern2pt}#1\sim\cr\noalign{\kern-2pt}}}}}
\newcommand{\appropto}{\mathpalette\approptoinn\relax}

\makeatletter
\newcommand*{\addFileDependency}[1]{\typeout{(#1)}
  \@addtofilelist{#1}
  \IfFileExists{#1}{}{\typeout{No file #1.}}
}
\makeatother

\makeatletter
\newcommand{\removelatexerror}{\let\@latex@error\@gobble}
\makeatother

\newcommand{\tomcom}[1]{#1}

\newcommand{\statevar}[0]{x}
\newcommand{\statevec}[0]{\ensuremath{\tv{statevar}}\xspace}

\usepackage{etoolbox}
\makeatletter
\patchcmd\@combinedblfloats{\box\@outputbox}{\unvbox\@outputbox}{}{\GenericWarning{\noexpand patch failed}}
\makeatother
\begin{document}

\title{Neural Ordinary Differential Equation Control of Dynamics on Graphs}

\author[1]{Thomas Asikis}
\author[2]{Lucas B\"ottcher}
\author[1]{Nino Antulov-Fantulin}
\affil[1]{ETH Z\"urich, Switzerland}
\affil[2]{UCLA, Los Angeles, USA}
\affil[1]{\textit {\{asikist,anino\}@ethz.ch}}
\affil[2]{\textit {lucasb@g.ucla.edu}}

\maketitle

\begin{abstract}
We study the ability of neural networks to calculate feedback control signals that steer trajectories of continuous time non-linear dynamical systems on graphs, which we represent with neural ordinary differential equations (neural ODEs). To do so, we present a neural-ODE control (NODEC) framework and find that it can learn feedback control signals that drive graph dynamical systems into desired target states. While we use loss functions that do not constrain the control energy, our results show, in accordance with related work, that NODEC produces low energy control signals. Finally, we evaluate the performance and versatility of NODEC against well-known feedback controllers and deep reinforcement learning. We use NODEC to generate feedback controls for systems of more than one thousand coupled, non-linear ODEs that represent epidemic processes and coupled oscillators.
\end{abstract}

\section{Introduction}\label{sec:introduction}
Dynamical processes on complex networks are common tools to model a wide range of real-world phenomena including opinion dynamics~\cite{bottcher2018clout,hoferer2020impact}, epidemic spreading~\cite{brockmann2013hidden,antulov2015identification,bottcher2017targeted}, synchronization~\cite{rodrigues2016kuramoto, yeung1999time}, and financial distress propagation~\cite{Delpini2013}. 
Continuous-time dynamics on complex networks can be described by different frameworks including Chapman--Kolmogorov~\cite{van1992stochastic}, 
Fokker--Planck~\cite{StohProcStatPhy}, stochastic differential~\cite{andersson2012stochastic}, and ordinary differential~\cite{castellano2010thresholds, gleeson2013binary,barzel2013universality} equations.
The structure of many real-world systems is described by networks with certain common properties including small-world effects~\cite{watts1998collective}, heavy-tail degree distributions~\cite{price1965networks, barabasi1999emergence}, community structure~\cite{girvan2002community}, and other features~\cite{dorogovtsev2008critical,leskovec2010kronecker}. The control of dynamical processes on networks~\cite{liu2011controllability,liu2016control} is a challenging task with applications in engineering, biology, and the social sciences~\cite{barrat2008dynamical,RevModPhys.87.925}. 
Control signals can be calculated by solving boundary-value PMP problems~\cite{mangasarian1966sufficient,kamien1971sufficient,mcshane1989calculus}, or computing solutions of the Hamilton–Jacobi–Bellman equation (HJB).
Complementing the above approaches, we develop a Neural-ODE control (NODEC) framework that controls fully observable graph dynamical systems using neural ODEs~\cite{chen2018neural}. 
Within this framework, feedback control signals are calculated by minimizing a loss function describing differences between the current and target states. We perform extensive numerical experiments on coupled high-dimensional non-linear dynamical systems to showcase the ability of NODEC to calculate effective control signals.

Mathematically, systems are ``controllable'' if they can be steered from any initial state ${\tv{\statevar}}(t_0)$ to any desired state $\target{\tv{\statevar}}(T)$ in finite time $T$.
For linear systems, an analytical condition for controllability of linear time-invariant (LTI) systems was derived by Kalman in the 1960s~\cite{kalman1960contributions} and is known today as Kalman's rank criterion. 
In 1969, Popov, Belevitch, and Hautus~\cite{hautus1969controllability} introduced another controllability test for LTI systems that relies on solutions of an eigenvalue problem of the state matrix. 
In the 1970s, Lin introduced the framework of structural controllability~\cite{lin1974structural} as a generalization of prior definitions of controllability on graphs. 
More recently, different large-scale social, technical, and biological networks were analyzed from a network controllability perspective~\cite{liu2011controllability,ruths2014control} building on the framework introduced by Lin~\cite{lin1974structural}. 
Controlling a complex system becomes more challenging as the number of nodes that can receive a control signal (driver nodes) decreases.
Furthermore, Ref.~\cite{yan2012controlling} addresses the important issue of quantifying the (control) energy that is needed to control LTI systems.
Steering the dynamical system to the target state becomes even harder when energy minimization is also accounted for.

To solve general non-linear optimal control problems with energy and driver node constraints, two main approaches are used: (i) Pontryagin's maximum principle (PMP) ~\cite{mangasarian1966sufficient,kamien1971sufficient,mcshane1989calculus} and (ii) Bellman's (approximate) dynamic programming~\cite{zhou1990maximum,fleming2006controlled,bellman2015applied,sutton2018reinforcement,kiumarsi2017optimal}.
Pontryagin's maximum principle~\cite{mangasarian1966sufficient,kamien1971sufficient,mcshane1989calculus} is based on variational calculus and transforms the original infinite-dimensional control problem to a boundary-value problem in a Hamiltonian framework.
The downside of this approach is that the resulting boundary-value problems are often very difficult to solve.
An alternative to variational methods is provided by Bellman's dynamic programming, which relies on the HJB equation. 
Given a quadratic loss on the control input, the HJB equation can be transformed into a partial-differential equation (PDE)~\cite{fleming2006controlled}. 
Dynamic programming and PMP are connected through the viscosity solutions of the aforementioned PDEs~\cite{zhou1990maximum}.
However, in most cases, the HJB equation is hard to solve~\cite{bellman2015applied} and does not admit smooth solutions~\cite{Frankowska}.
Most reinforcement-learning-based controls~\cite{sutton2018reinforcement} rely on optimizing the HJB equation and can be viewed as an approximation of the dynamic programming~\cite{kiumarsi2017optimal} approach.

In this article we follow an alternative approach, where we extend the neural ordinary differential equation framework to solve feedback control problems.
We describe and evaluate the ability of  neural ordinary differential equation control (NODEC) to efficiently control non-linear continuous time dynamical systems by calculating feedback control signals.
In \Cref{sec:related:work}, we discuss related work. \Cref{sec:dynamical:systems} summarizes mathematical concepts that are relevant for controlling graph dynamical systems. In \Cref{sec:NODEC}, 
we provide an overview of the basic features of NODEC and formulate conditions for its successful application to solve control problems. 
In \Cref{sec:results}, we showcase the ability of NODEC to efficiently control different graph dynamical systems that are described by coupled ODEs. 
In particular, we use NODEC to calculate feedback controls that synchronize coupled oscillators and contain disease dynamics with limited number of driver nodes.
Interestingly, NODEC achieves low energy controls without sacrificing performance.
\Cref{sec:discussion} concludes our paper. 

\section{Related Work}
\label{sec:related:work}
Previous works used neural networks in control applications~\cite{lewis2020neural}, 
in particular for parameter estimation of model predictive control~\cite{yoo2005stable,amos2018differentiable}.
Extensive applications of neural networks are also found in the field of Proportional-Integral-Derivative (PID) controllers~\cite{lewis2020neural},
where the gain factors are calculated via neural networks.
Shallow neural networks have been trained to interact with and control smaller-scale ODE systems~\cite{lewis2020neural}, 
without using neural ODEs or deep architectures.
Recently, deep neural networks have demonstrated high performance in control tasks, 
and notably on related work on differentiable physics~\cite{holl2020learning} that often use PMP.
Deep reinforcement learning~\cite{pang2020reinforcement} models are also used and rely on approximation of the HJB approach.
Other gradient-based non-neural network approaches rely on the usage of adjoint methods~\cite{biccari2020stochastic}. 
Such model approaches follow the solutions of the PMP principle and calculus of variations solutions. 
One can also design generic approaches to control network dynamics~\cite{wijayanto2017flow,cornelius2013realistic}.
Optimal control with NODEC, where the the neural network is only a function of time $t$ is extensively studied in Ref.~\cite{bottcher2021implicit}, 
where it is compared with analytically derived methods.
The current work focuses on feedback control methods where the input of the neural network is the state vector $\tv{\statevar}(t)$.
We study non-linear dynamical systems, where minimum energy (optimal) controls are not always known.
In our work, we always choose state-of-the-art control solutions when available, 
such as feedback control~\cite{skardal2016controlling} and deep reinforcement learning methods~\cite{haarnoja2018soft,fujimoto2018addressing},
so that we can compare NODEC performance with corresponding baselines. 
The main contributions of this work are: 
(i) introduction of an adaptive efficient feedback control approximation methodology with implicit energy regularization properties that relies on neural ODEs, 
(ii) detailed numerical experiments involving high-dimensional non-linear dynamical systems with minimum driver node constraints,
and (iii) an extensively tested codebase that can be easily used and extended on other nonlinear control applications.

\section{Feedback Control of Graph Dynamical Systems}
\label{sec:dynamical:systems}

A graph $G(V,E)$ is an ordered pair, where $\ts{V}$ and $\ts{E} \subseteq \ts{V} \times \ts{V}$ are the corresponding sets of $|\ts{V}|=N$ nodes and edges. Although, in network science~\cite{newman2018networks}, it is more common to refer to graphs as networks, 
in this paper we will use the term ``graph'' instead of ``network'' to avoid confusion with neural networks. 
Throughout this paper, we study dynamical systems on graphs described by the adjacency-matrix $\matr{A}$, which has non-zero elements $\matr{A}_{i j}$ if and only if nodes $i$ and $j$ are connected. 
We describe controlled graph dynamical systems by ODEs of the form
\begin{equation}
     \dot{\tv{\statevar}} (t) =\tomcom{\tv{f}_\matr{A}\left(t, \tv{\statevar}(t), \tv{u}(\tv{\statevar}(t))\right),}
     \label{eq:dynamical_system}
\end{equation}
where $\tv{\statevar}(t) \in \mathbb{R}^N$ denotes the state vector and $\tv{u}(\tv{\statevar}(t))\in \mathbb{R}^{M}$ ($M \leq N$) an external control signal \tomcom{applied to $M \leq N$ (driver) nodes. 
The adjacency matrix in the subscript $\tv{f}_{\matr{A}}$ denotes the graph-coupled interactions in the ODE system. 
For the remainder of the article, we omit the subscript as all systems under evaluation are graph-coupled ODEs that have fixed adjacency matrices over time.}
We use Newton's dot notation for differentiation $\dot{\tv{\statevar}} (t)$. The function $\tv{f}$ in \Cref{eq:dynamical_system} accounts for both (time-dependent) interactions between nodes and the influence of external control signals on the evolution of $\tv{\statevar}(t)$. 
We assume that the system state $\tv{\statevar}$ is fully observable.
In control theory, the control signal $\tv{u}(\tv{\statevar}(t))$ is often calculated via two approaches:
either by using time as input (i.e., $\tv{u}=\tv{u}(t)$)~\cite{yan2012controlling} 
or by using the system's state at time $t$ as input (i.e., $\tv{u}=\tv{u}(\tv{\statevar}(t)))$~\cite{mayr1970origins}.
The latter calculation is often used in state-feedback control~\cite{mayr1970origins}, 
where the control signal is calculated as a function of the difference between the current system target state and the control target state $g(\tv{\statevar}(t) - \tv{\statevar}^*)$.
In the present article, we focus on state-feedback control and denote control signals by $\tv{u}(\tv{\statevar}(t))$.
The applicability of the current framework on time-dependent controls is evaluated in detail in Ref.~\cite{bottcher2021implicit}.

In principle, \Cref{eq:dynamical_system} can be solved numerically, for instance using an explicit Euler scheme: 
For some given state $\tv{\statevar}(t)$ at time $t$, the state of the system at time $t+\Delta t$ is
$\tv{\statevar}(t+\Delta t) = \tv{\statevar}(t) + \Delta t\,\tv{f}\left(t, \tv{\statevar}(t), \tv{u}(\tv{\statevar}(t))\right)$. 
Apart from an Euler forward integration scheme, 
there exist many more numerical methods~\cite{shampine2018numerical} to solve \Cref{eq:dynamical_system}. 
We use the expression $\text{ODESolve}(\tv{\statevar}(t),t_0,T, f, \tv{u}(\tv{\statevar}(t)))$ to indicate a generic ODE solver that uses the right-hand side of \Cref{eq:dynamical_system} as an input and computes \tomcom{the state trajectory, or set of state vectors, ${\ts{\statevar}}_{t_0}^T = \{\tv{\statevar}(t)\}_{t_0 \leq t \leq T}$}.
In \Cref{sec:results}, we employ Dormand--Prince and Runge--Kutta schemes as our $\text{ODESolve}$ methods.
Nevertheless, when numerically calculating analytical solutions may introduce numerical instabilities and can be computationally expensive for large systems. 
Numerical methods are also required for general non-linear ODE systems, which do not have analytic tractable solutions of optimal control signals. 
Additionally, as mentioned in \Cref{sec:introduction}, the control of a complex dynamical system becomes more challenging when considering minimum energy and driver node constraints.

\subsection{Driver Node Selection}\label{sec:driver:nodes}
Our aim is to showcase the ability of NODEC to produce efficient feedback controls for systems where the number driver nodes approaches the minimum number necessary to achieve control.
Thus, we need to identify set of driver nodes that are able to fully control the underlying dynamics. 
Usually, we are interested in finding the minimum set of driver nodes, 
which is equivalent to the graph-theoretical problems of maximum matching or minimum edge dominating sets~\cite{Commault2002, yamada1990graph}. 
However, for general graphs, finding the maximum matching set is NP-hard~\cite{garey1979computers,olshevsky2014minimal}. 
In our NODEC framework, we determine driver nodes according to two methods:  (i) the maximum matching method~\cite{liu2011controllability} for disease dynamics and (ii) from stability criteria in the case of Kuramoto oscilaltors~\cite{skardal2015control}.
We denote the set of driver nodes $\ts{B} \subseteq \ts{V}$ and its cardinality $M$. A driver matrix $\matr{B} \in \mathbb{R}^{N\times M}$, 
where we set $\matr{B}_{i,m}=1$ if $i$ is a driver node and $u_m$ is applied on $i$ and $\matr{B}_{i,m}=0$ otherwise. 
\tomcom{The driver matrix $\matr{B}$ connects a control input $u_{m}(\tv{\statevar}(t))$ for a driver node $m$ to the corresponding graph node $i$ for non-zero elements $\matr{B}_{i,m}$.
Although NODEC can be used to evaluate shared and/or interacting control signals\footnote{i.e. the same control signal $u_m(\tv{x}(t)$ is applied to more than one node or more than one control signals $u_m(\tv{x}(t)), u_{m'}(\tv{x}(t))$ are applied to the same node $i$.}, in the current article we evaluate dynamical systems where each control signal $u_m(\tv{\statevar}(t))$ is assigned to one and only one graph node $i$, thus only one matrix element $\matr{B}_{i,m}$ is non-zero per row $\matr{B}_i$.}
Literature is rich in studies on driver node placement on graphs, 
there is considerably fewer work that addresses ways of efficiently finding control inputs for high-dimensional dynamical systems with a limited number of driver nodes. 

\subsection{Control Energy Constraints}
In complex systems, it may not always be possible to apply any control signal to a driver node. 
Consider a disease that spreads between networked communities (nodes) and a control signal that denotes the intensity of quarantine. 
Applying a constant control signal with high values indicating blanket lockdown measures may not be acceptable by society.
In the given example, our goal would be to contain disease spreading as much as possible, while applying appropriate control signals to the driver nodes.
A widely use metric for the intensity of the control signal~\cite{liu2011controllability} is the control energy
\begin{equation}
E\left(\tv{u}(\tv{\statevar}(t))\right) = \int_{t_0}^{T} \norm{\tv{u}(\tv{\statevar}(t))}_2^2\, \mathrm{d}t,
\label{eq:energy}
\end{equation}
where $\norm{\cdot}_2$ denotes the L2 norm. 
In our numerical experiments \Cref{eq:energy}, we approximate the corresponding integral by
\begin{equation}\label{eq:energy:approx}
    E\left(T\right) \approx \sum_{k=1}^K \norm{\tv{u}\left( \tv{\statevar}(t_0+k \Delta t)\right)}^2_2 \, \Delta t,
\end{equation}

In Ref.~\cite{bottcher2021implicit}, we show that NODEC approximates optimal (or minimum energy) control signals without the necessity of explicitly accounting for an integrated energy cost in the underlying loss function. 
Instead, NODEC implicitly minimizes the control energy via the interplay of an induced gradient descent, neural-ODE solver dynamics, and neural-network initialization. 
Avoiding the control energy term in a constrained optimization also reduces computational cost of learning compared to solving boundary-value PMP problems~\cite{mangasarian1966sufficient,kamien1971sufficient,mcshane1989calculus}, 
or computing solutions of the Hamilton–Jacobi–Bellman (HJB) equation~\cite{zhou1990maximum,fleming2006controlled,bellman2015applied,sutton2018reinforcement,kiumarsi2017optimal}.
In the present article, we provide evidence that NODEC achieves lower energy and higher performance when compared to feedback controls for large complex systems.

\section{Neural ODE Control}
\label{sec:NODEC}
\begin{figure}
    \centering
    \resizebox{1\columnwidth}{!}{

\tikzset{every picture/.style={line width=0.75pt}} 
\begin{tikzpicture}[x=0.75pt,y=0.75pt,yscale=-1,xscale=1]

\draw [color={rgb, 255:red, 74; green, 144; blue, 226 }  ,draw opacity=1 ]   (89.56,62.5) -- (284.31,62.5) ;
\draw [color={rgb, 255:red, 155; green, 155; blue, 155 }  ,draw opacity=1 ] [dash pattern={on 0.84pt off 2.51pt}]  (89.62,73.37) -- (89.62,262.35) ;
\draw [color={rgb, 255:red, 155; green, 155; blue, 155 }  ,draw opacity=1 ] [dash pattern={on 0.84pt off 2.51pt}]  (154.18,73.37) -- (154.18,262.35) ;
\draw [color={rgb, 255:red, 155; green, 155; blue, 155 }  ,draw opacity=1 ] [dash pattern={on 0.84pt off 2.51pt}]  (177.19,73.37) -- (177.19,262.35) ;
\draw [color={rgb, 255:red, 155; green, 155; blue, 155 }  ,draw opacity=1 ] [dash pattern={on 0.84pt off 2.51pt}]  (232.36,73.37) -- (232.36,262.35) ;
\draw [color={rgb, 255:red, 155; green, 155; blue, 155 }  ,draw opacity=1 ] [dash pattern={on 0.84pt off 2.51pt}]  (266.72,73.37) -- (266.72,262.35) ;

\draw [line width=2.25]  [dash pattern={on 2.53pt off 3.02pt}]  (52.24,192.99) -- (39.3,173.13) ;
\draw    (54.65,194.53) -- (62.02,259.19) ;
\draw  [color={rgb, 255:red, 230; green, 97; blue, 1 }  ,draw opacity=1 ][fill={rgb, 255:red, 255; green, 255; blue, 255 }  ,fill opacity=1 ] (47.71,192.36) .. controls (47.71,188.46) and (50.87,185.29) .. (54.78,185.29) .. controls (58.68,185.29) and (61.85,188.46) .. (61.85,192.36) .. controls (61.85,196.27) and (58.68,199.43) .. (54.78,199.43) .. controls (50.87,199.43) and (47.71,196.27) .. (47.71,192.36) -- cycle ;
\draw  [color={rgb, 255:red, 94; green, 60; blue, 153 }  ,draw opacity=1 ][fill={rgb, 255:red, 255; green, 255; blue, 255 }  ,fill opacity=1 ] (52.95,259.69) .. controls (52.95,255.79) and (56.11,252.62) .. (60.02,252.62) .. controls (63.92,252.62) and (67.09,255.79) .. (67.09,259.69) .. controls (67.09,263.6) and (63.92,266.76) .. (60.02,266.76) .. controls (56.11,266.76) and (52.95,263.6) .. (52.95,259.69) -- cycle ;
\draw  [color={rgb, 255:red, 230; green, 97; blue, 1 }  ,draw opacity=1 ] (65.39,229.32) -- (59.6,236.33) -- (52.6,230.54) ;
\draw  [color={rgb, 255:red, 230; green, 97; blue, 1 }  ,draw opacity=1 ] (64.78,222.92) -- (58.99,229.93) -- (51.99,224.14) ;

\draw [color={rgb, 255:red, 94; green, 60; blue, 153 }  ,draw opacity=1 ]   (93.04,238.48) .. controls (112.96,217.21) and (172.32,226.8) .. (187.35,224.26) .. controls (202.38,221.73) and (221.95,201.76) .. (238.66,209.92) .. controls (255.37,218.08) and (252.52,221.54) .. (267.9,214.45) ;
\draw [color={rgb, 255:red, 178; green, 171; blue, 210 }  ,draw opacity=1 ] [dash pattern={on 4.5pt off 4.5pt}]  (93.04,238.48) .. controls (131.77,238.61) and (170.2,240.91) .. (186.64,247.37) .. controls (203.08,253.83) and (227.1,251.01) .. (234.76,251.14) .. controls (242.41,251.27) and (256.26,249.48) .. (269.09,250.82) ;
\draw [color={rgb, 255:red, 200; green, 200; blue, 200 }  ,draw opacity=1 ][line width=0.75]    (87.7,201.99) -- (272.54,201.99) ;
\draw [color={rgb, 255:red, 0; green, 0; blue, 0 }  ,draw opacity=1 ][line width=0.75]    (89.62,262.48) -- (279.71,262.48) ;
\draw [shift={(282.71,262.48)}, rotate = 180] [fill={rgb, 255:red, 0; green, 0; blue, 0 }  ,fill opacity=1 ][line width=0.08]  [draw opacity=0] (5.36,-2.57) -- (0,0) -- (5.36,2.57) -- cycle    ;
\draw [color={rgb, 255:red, 230; green, 97; blue, 1 }  ,draw opacity=1 ]   (89.24,126.24) .. controls (130.35,135.33) and (136.49,134.2) .. (176.38,123.99) .. controls (216.27,113.78) and (226.7,145.54) .. (267.2,146.05) ;
\draw [color={rgb, 255:red, 222; green, 180; blue, 149 }  ,draw opacity=1 ] [dash pattern={on 4.5pt off 4.5pt}]  (89.9,143.39) .. controls (123.7,154.63) and (136.97,139) .. (170.16,129.91) .. controls (203.35,120.82) and (219.65,152.05) .. (266.72,140.72) ;
\draw [color={rgb, 255:red, 253; green, 184; blue, 99 }  ,draw opacity=1 ]   (89.79,86.64) .. controls (126.13,82.33) and (140.29,104.09) .. (183.4,97.11) .. controls (226.51,90.13) and (221.58,95.13) .. (268.38,106.51) ;
\draw [color={rgb, 255:red, 74; green, 74; blue, 74 }  ,draw opacity=1 ][line width=0.75]    (45.83,117.25) -- (273.41,117.25) ;
\draw [color={rgb, 255:red, 74; green, 74; blue, 74 }  ,draw opacity=1 ][line width=0.75]    (46.21,77.58) -- (274.42,77.58) ;
\draw [color={rgb, 255:red, 230; green, 97; blue, 1 }  ,draw opacity=1 ]   (90.69,186.06) .. controls (133.39,169.94) and (156.58,189.66) .. (185.58,181.34) .. controls (214.57,173.03) and (231.97,160.97) .. (268.34,178.45) ;
\draw [color={rgb, 255:red, 230; green, 97; blue, 1 }  ,draw opacity=1 ] [dash pattern={on 4.5pt off 4.5pt}]  (90.69,186.06) .. controls (168.18,208.02) and (179.78,181.23) .. (216.15,182.7) .. controls (252.53,184.16) and (248.84,186.46) .. (268.34,186.87) ;
\draw  [color={rgb, 255:red, 255; green, 255; blue, 255 }  ,draw opacity=1 ][fill={rgb, 255:red, 230; green, 97; blue, 1 }  ,fill opacity=1 ] (267.18,171.21) -- (268.97,174.83) -- (272.96,175.41) -- (270.07,178.22) -- (270.75,182.19) -- (267.18,180.32) -- (263.61,182.19) -- (264.29,178.22) -- (261.41,175.41) -- (265.4,174.83) -- cycle ;
\draw  [color={rgb, 255:red, 255; green, 255; blue, 255 }  ,draw opacity=1 ][fill={rgb, 255:red, 94; green, 60; blue, 153 }  ,fill opacity=1 ] (266.74,208.38) -- (268.52,211.99) -- (272.51,212.57) -- (269.63,215.39) -- (270.31,219.36) -- (266.74,217.48) -- (263.17,219.36) -- (263.85,215.39) -- (260.96,212.57) -- (264.95,211.99) -- cycle ;
\draw  [color={rgb, 255:red, 255; green, 255; blue, 255 }  ,draw opacity=1 ][fill={rgb, 255:red, 230; green, 97; blue, 1 }  ,fill opacity=1 ] (263.92,187.65) .. controls (263.92,185.71) and (265.49,184.14) .. (267.43,184.14) .. controls (269.36,184.14) and (270.93,185.71) .. (270.93,187.65) .. controls (270.93,189.59) and (269.36,191.16) .. (267.43,191.16) .. controls (265.49,191.16) and (263.92,189.59) .. (263.92,187.65) -- cycle ;
\draw  [color={rgb, 255:red, 255; green, 255; blue, 255 }  ,draw opacity=1 ][fill={rgb, 255:red, 94; green, 60; blue, 153 }  ,fill opacity=1 ] (87.19,248.48) .. controls (87.19,246.55) and (88.76,244.98) .. (90.69,244.98) .. controls (92.63,244.98) and (94.2,246.55) .. (94.2,248.48) .. controls (94.2,250.42) and (92.63,251.99) .. (90.69,251.99) .. controls (88.76,251.99) and (87.19,250.42) .. (87.19,248.48) -- cycle ;
\draw  [color={rgb, 255:red, 255; green, 255; blue, 255 }  ,draw opacity=1 ][fill={rgb, 255:red, 94; green, 60; blue, 153 }  ,fill opacity=1 ] (151.05,241.33) .. controls (151.05,239.39) and (152.62,237.82) .. (154.55,237.82) .. controls (156.49,237.82) and (158.06,239.39) .. (158.06,241.33) .. controls (158.06,243.26) and (156.49,244.83) .. (154.55,244.83) .. controls (152.62,244.83) and (151.05,243.26) .. (151.05,241.33) -- cycle ;
\draw  [color={rgb, 255:red, 255; green, 255; blue, 255 }  ,draw opacity=1 ][fill={rgb, 255:red, 94; green, 60; blue, 153 }  ,fill opacity=1 ] (150.53,222.98) .. controls (150.53,221.04) and (152.1,219.47) .. (154.03,219.47) .. controls (155.97,219.47) and (157.54,221.04) .. (157.54,222.98) .. controls (157.54,224.91) and (155.97,226.48) .. (154.03,226.48) .. controls (152.1,226.48) and (150.53,224.91) .. (150.53,222.98) -- cycle ;
\draw  [color={rgb, 255:red, 255; green, 255; blue, 255 }  ,draw opacity=1 ][fill={rgb, 255:red, 94; green, 60; blue, 153 }  ,fill opacity=1 ] (174.57,244.82) .. controls (174.57,242.88) and (176.14,241.31) .. (178.07,241.31) .. controls (180.01,241.31) and (181.58,242.88) .. (181.58,244.82) .. controls (181.58,246.75) and (180.01,248.32) .. (178.07,248.32) .. controls (176.14,248.32) and (174.57,246.75) .. (174.57,244.82) -- cycle ;
\draw  [color={rgb, 255:red, 255; green, 255; blue, 255 }  ,draw opacity=1 ][fill={rgb, 255:red, 94; green, 60; blue, 153 }  ,fill opacity=1 ] (174.82,224.4) .. controls (174.82,222.46) and (176.39,220.89) .. (178.33,220.89) .. controls (180.27,220.89) and (181.84,222.46) .. (181.84,224.4) .. controls (181.84,226.33) and (180.27,227.9) .. (178.33,227.9) .. controls (176.39,227.9) and (174.82,226.33) .. (174.82,224.4) -- cycle ;
\draw  [color={rgb, 255:red, 255; green, 255; blue, 255 }  ,draw opacity=1 ][fill={rgb, 255:red, 94; green, 60; blue, 153 }  ,fill opacity=1 ] (228.84,207.86) .. controls (228.84,205.92) and (230.41,204.35) .. (232.35,204.35) .. controls (234.29,204.35) and (235.85,205.92) .. (235.85,207.86) .. controls (235.85,209.79) and (234.29,211.36) .. (232.35,211.36) .. controls (230.41,211.36) and (228.84,209.79) .. (228.84,207.86) -- cycle ;
\draw  [color={rgb, 255:red, 255; green, 255; blue, 255 }  ,draw opacity=1 ][fill={rgb, 255:red, 94; green, 60; blue, 153 }  ,fill opacity=1 ] (229.1,251.54) .. controls (229.1,249.6) and (230.67,248.03) .. (232.61,248.03) .. controls (234.54,248.03) and (236.11,249.6) .. (236.11,251.54) .. controls (236.11,253.47) and (234.54,255.04) .. (232.61,255.04) .. controls (230.67,255.04) and (229.1,253.47) .. (229.1,251.54) -- cycle ;
\draw  [color={rgb, 255:red, 255; green, 255; blue, 255 }  ,draw opacity=1 ][fill={rgb, 255:red, 94; green, 60; blue, 153 }  ,fill opacity=1 ] (263.48,250.5) .. controls (263.48,248.57) and (265.05,247) .. (266.98,247) .. controls (268.92,247) and (270.49,248.57) .. (270.49,250.5) .. controls (270.49,252.44) and (268.92,254.01) .. (266.98,254.01) .. controls (265.05,254.01) and (263.48,252.44) .. (263.48,250.5) -- cycle ;
\draw  [color={rgb, 255:red, 255; green, 255; blue, 255 }  ,draw opacity=1 ][fill={rgb, 255:red, 230; green, 97; blue, 1 }  ,fill opacity=1 ] (228.64,183.77) .. controls (228.64,181.84) and (230.21,180.27) .. (232.15,180.27) .. controls (234.08,180.27) and (235.65,181.84) .. (235.65,183.77) .. controls (235.65,185.71) and (234.08,187.28) .. (232.15,187.28) .. controls (230.21,187.28) and (228.64,185.71) .. (228.64,183.77) -- cycle ;
\draw  [color={rgb, 255:red, 255; green, 255; blue, 255 }  ,draw opacity=1 ][fill={rgb, 255:red, 230; green, 97; blue, 1 }  ,fill opacity=1 ] (228.64,173.43) .. controls (228.64,171.49) and (230.21,169.92) .. (232.15,169.92) .. controls (234.08,169.92) and (235.65,171.49) .. (235.65,173.43) .. controls (235.65,175.36) and (234.08,176.93) .. (232.15,176.93) .. controls (230.21,176.93) and (228.64,175.36) .. (228.64,173.43) -- cycle ;
\draw  [color={rgb, 255:red, 255; green, 255; blue, 255 }  ,draw opacity=1 ][fill={rgb, 255:red, 230; green, 97; blue, 1 }  ,fill opacity=1 ] (173.59,190.36) .. controls (173.59,188.43) and (175.16,186.86) .. (177.09,186.86) .. controls (179.03,186.86) and (180.6,188.43) .. (180.6,190.36) .. controls (180.6,192.3) and (179.03,193.87) .. (177.09,193.87) .. controls (175.16,193.87) and (173.59,192.3) .. (173.59,190.36) -- cycle ;
\draw  [color={rgb, 255:red, 255; green, 255; blue, 255 }  ,draw opacity=1 ][fill={rgb, 255:red, 230; green, 97; blue, 1 }  ,fill opacity=1 ] (173.59,183.35) .. controls (173.59,181.42) and (175.16,179.85) .. (177.09,179.85) .. controls (179.03,179.85) and (180.6,181.42) .. (180.6,183.35) .. controls (180.6,185.29) and (179.03,186.86) .. (177.09,186.86) .. controls (175.16,186.86) and (173.59,185.29) .. (173.59,183.35) -- cycle ;
\draw  [color={rgb, 255:red, 255; green, 255; blue, 255 }  ,draw opacity=1 ][fill={rgb, 255:red, 230; green, 97; blue, 1 }  ,fill opacity=1 ] (150.33,194.21) .. controls (150.33,192.27) and (151.9,190.7) .. (153.83,190.7) .. controls (155.77,190.7) and (157.34,192.27) .. (157.34,194.21) .. controls (157.34,196.14) and (155.77,197.71) .. (153.83,197.71) .. controls (151.9,197.71) and (150.33,196.14) .. (150.33,194.21) -- cycle ;
\draw  [color={rgb, 255:red, 255; green, 255; blue, 255 }  ,draw opacity=1 ][fill={rgb, 255:red, 230; green, 97; blue, 1 }  ,fill opacity=1 ] (149.94,182.58) .. controls (149.94,180.64) and (151.51,179.07) .. (153.44,179.07) .. controls (155.38,179.07) and (156.95,180.64) .. (156.95,182.58) .. controls (156.95,184.51) and (155.38,186.08) .. (153.44,186.08) .. controls (151.51,186.08) and (149.94,184.51) .. (149.94,182.58) -- cycle ;
\draw  [color={rgb, 255:red, 255; green, 255; blue, 255 }  ,draw opacity=1 ][fill={rgb, 255:red, 230; green, 97; blue, 1 }  ,fill opacity=1 ] (86.02,186.06) .. controls (86.02,184.13) and (87.59,182.56) .. (89.52,182.56) .. controls (91.46,182.56) and (93.03,184.13) .. (93.03,186.06) .. controls (93.03,188) and (91.46,189.57) .. (89.52,189.57) .. controls (87.59,189.57) and (86.02,188) .. (86.02,186.06) -- cycle ;
\draw  [color={rgb, 255:red, 0; green, 0; blue, 0 }  ,draw opacity=1 ][fill={rgb, 255:red, 155; green, 155; blue, 155 }  ,fill opacity=1 ] (87.17,262.54) .. controls (87.17,261.23) and (88.24,260.16) .. (89.56,260.16) .. controls (90.88,260.16) and (91.95,261.23) .. (91.95,262.54) .. controls (91.95,263.86) and (90.88,264.93) .. (89.56,264.93) .. controls (88.24,264.93) and (87.17,263.86) .. (87.17,262.54) -- cycle ;
\draw  [color={rgb, 255:red, 0; green, 0; blue, 0 }  ,draw opacity=1 ][fill={rgb, 255:red, 155; green, 155; blue, 155 }  ,fill opacity=1 ] (151.79,262.54) .. controls (151.79,261.23) and (152.86,260.16) .. (154.18,260.16) .. controls (155.5,260.16) and (156.57,261.23) .. (156.57,262.54) .. controls (156.57,263.86) and (155.5,264.93) .. (154.18,264.93) .. controls (152.86,264.93) and (151.79,263.86) .. (151.79,262.54) -- cycle ;
\draw  [color={rgb, 255:red, 0; green, 0; blue, 0 }  ,draw opacity=1 ][fill={rgb, 255:red, 155; green, 155; blue, 155 }  ,fill opacity=1 ] (175.49,262.54) .. controls (175.49,261.23) and (176.56,260.16) .. (177.87,260.16) .. controls (179.19,260.16) and (180.26,261.23) .. (180.26,262.54) .. controls (180.26,263.86) and (179.19,264.93) .. (177.87,264.93) .. controls (176.56,264.93) and (175.49,263.86) .. (175.49,262.54) -- cycle ;
\draw  [color={rgb, 255:red, 0; green, 0; blue, 0 }  ,draw opacity=1 ][fill={rgb, 255:red, 155; green, 155; blue, 155 }  ,fill opacity=1 ] (229.98,262.54) .. controls (229.98,261.23) and (231.05,260.16) .. (232.36,260.16) .. controls (233.68,260.16) and (234.75,261.23) .. (234.75,262.54) .. controls (234.75,263.86) and (233.68,264.93) .. (232.36,264.93) .. controls (231.05,264.93) and (229.98,263.86) .. (229.98,262.54) -- cycle ;
\draw  [color={rgb, 255:red, 0; green, 0; blue, 0 }  ,draw opacity=1 ][fill={rgb, 255:red, 155; green, 155; blue, 155 }  ,fill opacity=1 ] (264.33,262.54) .. controls (264.33,261.23) and (265.4,260.16) .. (266.72,260.16) .. controls (268.04,260.16) and (269.1,261.23) .. (269.1,262.54) .. controls (269.1,263.86) and (268.04,264.93) .. (266.72,264.93) .. controls (265.4,264.93) and (264.33,263.86) .. (264.33,262.54) -- cycle ;
\draw   (40.71,50.76) .. controls (40.71,35.78) and (52.86,23.63) .. (67.85,23.63) -- (314,23.63) .. controls (314,23.63) and (314,23.63) .. (314,23.63) -- (314,132.18) .. controls (314,147.17) and (301.85,159.32) .. (286.86,159.32) -- (40.71,159.32) .. controls (40.71,159.32) and (40.71,159.32) .. (40.71,159.32) -- cycle ;
\draw    (154.18,54.09) -- (154.18,62.5) ;
\draw    (150.75,47.03) -- (154.18,56.55) ;
\draw    (157.61,47.03) -- (154.18,56.55) ;
\draw    (157.61,47.03) -- (161.04,56.55) ;
\draw    (161.04,54.09) -- (154.18,62.5) ;
\draw    (147.32,54.09) -- (154.18,62.5) ;
\draw    (150.75,47.03) -- (147.32,56.55) ;
\draw  [fill={rgb, 255:red, 255; green, 255; blue, 255 }  ,fill opacity=1 ] (158.55,52.42) -- (163.53,52.42) -- (163.53,57.4) -- (158.55,57.4) -- cycle ;
\draw  [fill={rgb, 255:red, 255; green, 255; blue, 255 }  ,fill opacity=1 ] (144.83,52.42) -- (149.81,52.42) -- (149.81,57.4) -- (144.83,57.4) -- cycle ;
\draw  [fill={rgb, 255:red, 255; green, 255; blue, 255 }  ,fill opacity=1 ] (151.69,52.42) -- (156.67,52.42) -- (156.67,57.4) -- (151.69,57.4) -- cycle ;

\draw  [color={rgb, 255:red, 255; green, 255; blue, 255 }  ,draw opacity=1 ][fill={rgb, 255:red, 74; green, 74; blue, 74 }  ,fill opacity=1 ] (155.12,44.54) -- (160.1,44.54) -- (160.1,49.51) -- (155.12,49.51) -- cycle ;
\draw  [color={rgb, 255:red, 255; green, 255; blue, 255 }  ,draw opacity=1 ][fill={rgb, 255:red, 74; green, 74; blue, 74 }  ,fill opacity=1 ] (148.26,44.54) -- (153.24,44.54) -- (153.24,49.51) -- (148.26,49.51) -- cycle ;

\draw  [color={rgb, 255:red, 255; green, 255; blue, 255 }  ,draw opacity=1 ][fill={rgb, 255:red, 74; green, 74; blue, 74 }  ,fill opacity=1 ] (151.69,60.01) -- (156.67,60.01) -- (156.67,64.98) -- (151.69,64.98) -- cycle ;
\draw    (150.6,35.05) -- (150.6,41.54) ;
\draw [shift={(150.6,44.54)}, rotate = 270] [fill={rgb, 255:red, 0; green, 0; blue, 0 }  ][line width=0.08]  [draw opacity=0] (5.36,-2.57) -- (0,0) -- (5.36,2.57) -- (3.56,0) -- cycle    ;
\draw    (157.53,34.83) -- (157.53,41.62) ;
\draw [shift={(157.53,44.62)}, rotate = 270] [fill={rgb, 255:red, 0; green, 0; blue, 0 }  ][line width=0.08]  [draw opacity=0] (5.36,-2.57) -- (0,0) -- (5.36,2.57) -- (3.56,0) -- cycle    ;
\draw    (154.18,62.5) -- (154.18,69.05) ;
\draw [shift={(154.18,72.05)}, rotate = 270] [fill={rgb, 255:red, 0; green, 0; blue, 0 }  ][line width=0.08]  [draw opacity=0] (5.36,-2.57) -- (0,0) -- (5.36,2.57) -- (3.56,0) -- cycle    ;

\draw    (89.56,54.09) -- (89.56,62.5) ;
\draw    (86.13,47.03) -- (89.56,56.55) ;
\draw    (93,47.03) -- (89.56,56.55) ;
\draw    (93,47.03) -- (96.43,56.55) ;
\draw    (96.43,54.09) -- (89.56,62.5) ;
\draw    (82.7,54.09) -- (89.56,62.5) ;
\draw    (86.13,47.03) -- (82.7,56.55) ;
\draw  [fill={rgb, 255:red, 255; green, 255; blue, 255 }  ,fill opacity=1 ] (93.94,52.42) -- (98.91,52.42) -- (98.91,57.4) -- (93.94,57.4) -- cycle ;
\draw  [fill={rgb, 255:red, 255; green, 255; blue, 255 }  ,fill opacity=1 ] (80.21,52.42) -- (85.19,52.42) -- (85.19,57.4) -- (80.21,57.4) -- cycle ;
\draw  [fill={rgb, 255:red, 255; green, 255; blue, 255 }  ,fill opacity=1 ] (87.08,52.42) -- (92.05,52.42) -- (92.05,57.4) -- (87.08,57.4) -- cycle ;

\draw  [color={rgb, 255:red, 255; green, 255; blue, 255 }  ,draw opacity=1 ][fill={rgb, 255:red, 74; green, 74; blue, 74 }  ,fill opacity=1 ] (90.51,44.54) -- (95.48,44.54) -- (95.48,49.51) -- (90.51,49.51) -- cycle ;
\draw  [color={rgb, 255:red, 255; green, 255; blue, 255 }  ,draw opacity=1 ][fill={rgb, 255:red, 74; green, 74; blue, 74 }  ,fill opacity=1 ] (83.65,44.54) -- (88.62,44.54) -- (88.62,49.51) -- (83.65,49.51) -- cycle ;

\draw  [color={rgb, 255:red, 255; green, 255; blue, 255 }  ,draw opacity=1 ][fill={rgb, 255:red, 74; green, 74; blue, 74 }  ,fill opacity=1 ] (87.08,60.01) -- (92.05,60.01) -- (92.05,64.98) -- (87.08,64.98) -- cycle ;
\draw    (85.98,35.05) -- (85.98,41.54) ;
\draw [shift={(85.98,44.54)}, rotate = 270] [fill={rgb, 255:red, 0; green, 0; blue, 0 }  ][line width=0.08]  [draw opacity=0] (5.36,-2.57) -- (0,0) -- (5.36,2.57) -- (3.56,0) -- cycle    ;
\draw    (92.92,34.83) -- (92.92,41.62) ;
\draw [shift={(92.92,44.62)}, rotate = 270] [fill={rgb, 255:red, 0; green, 0; blue, 0 }  ][line width=0.08]  [draw opacity=0] (5.36,-2.57) -- (0,0) -- (5.36,2.57) -- (3.56,0) -- cycle    ;
\draw    (89.56,62.5) -- (89.56,69.05) ;
\draw [shift={(89.56,72.05)}, rotate = 270] [fill={rgb, 255:red, 0; green, 0; blue, 0 }  ][line width=0.08]  [draw opacity=0] (5.36,-2.57) -- (0,0) -- (5.36,2.57) -- (3.56,0) -- cycle    ;

\draw    (177.31,54.09) -- (177.31,62.5) ;
\draw    (173.88,47.03) -- (177.31,56.55) ;
\draw    (180.74,47.03) -- (177.31,56.55) ;
\draw    (180.74,47.03) -- (184.18,56.55) ;
\draw    (184.18,54.09) -- (177.31,62.5) ;
\draw    (170.45,54.09) -- (177.31,62.5) ;
\draw    (173.88,47.03) -- (170.45,56.55) ;
\draw  [fill={rgb, 255:red, 255; green, 255; blue, 255 }  ,fill opacity=1 ] (181.69,52.42) -- (186.66,52.42) -- (186.66,57.4) -- (181.69,57.4) -- cycle ;
\draw  [fill={rgb, 255:red, 255; green, 255; blue, 255 }  ,fill opacity=1 ] (167.96,52.42) -- (172.94,52.42) -- (172.94,57.4) -- (167.96,57.4) -- cycle ;
\draw  [fill={rgb, 255:red, 255; green, 255; blue, 255 }  ,fill opacity=1 ] (174.83,52.42) -- (179.8,52.42) -- (179.8,57.4) -- (174.83,57.4) -- cycle ;

\draw  [color={rgb, 255:red, 255; green, 255; blue, 255 }  ,draw opacity=1 ][fill={rgb, 255:red, 74; green, 74; blue, 74 }  ,fill opacity=1 ] (178.26,44.54) -- (183.23,44.54) -- (183.23,49.51) -- (178.26,49.51) -- cycle ;
\draw  [color={rgb, 255:red, 255; green, 255; blue, 255 }  ,draw opacity=1 ][fill={rgb, 255:red, 74; green, 74; blue, 74 }  ,fill opacity=1 ] (171.39,44.54) -- (176.37,44.54) -- (176.37,49.51) -- (171.39,49.51) -- cycle ;

\draw  [color={rgb, 255:red, 255; green, 255; blue, 255 }  ,draw opacity=1 ][fill={rgb, 255:red, 74; green, 74; blue, 74 }  ,fill opacity=1 ] (174.83,60.01) -- (179.8,60.01) -- (179.8,64.98) -- (174.83,64.98) -- cycle ;
\draw    (173.73,35.05) -- (173.73,41.54) ;
\draw [shift={(173.73,44.54)}, rotate = 270] [fill={rgb, 255:red, 0; green, 0; blue, 0 }  ][line width=0.08]  [draw opacity=0] (5.36,-2.57) -- (0,0) -- (5.36,2.57) -- (3.56,0) -- cycle    ;
\draw    (180.66,34.83) -- (180.66,41.62) ;
\draw [shift={(180.66,44.62)}, rotate = 270] [fill={rgb, 255:red, 0; green, 0; blue, 0 }  ][line width=0.08]  [draw opacity=0] (5.36,-2.57) -- (0,0) -- (5.36,2.57) -- (3.56,0) -- cycle    ;
\draw    (177.31,62.5) -- (177.31,69.05) ;
\draw [shift={(177.31,72.05)}, rotate = 270] [fill={rgb, 255:red, 0; green, 0; blue, 0 }  ][line width=0.08]  [draw opacity=0] (5.36,-2.57) -- (0,0) -- (5.36,2.57) -- (3.56,0) -- cycle    ;

\draw    (232.5,54.09) -- (232.5,62.5) ;
\draw    (229.06,47.03) -- (232.5,56.55) ;
\draw    (235.93,47.03) -- (232.5,56.55) ;
\draw    (235.93,47.03) -- (239.36,56.55) ;
\draw    (239.36,54.09) -- (232.5,62.5) ;
\draw    (225.63,54.09) -- (232.5,62.5) ;
\draw    (229.06,47.03) -- (225.63,56.55) ;
\draw  [fill={rgb, 255:red, 255; green, 255; blue, 255 }  ,fill opacity=1 ] (236.87,52.42) -- (241.85,52.42) -- (241.85,57.4) -- (236.87,57.4) -- cycle ;
\draw  [fill={rgb, 255:red, 255; green, 255; blue, 255 }  ,fill opacity=1 ] (223.15,52.42) -- (228.12,52.42) -- (228.12,57.4) -- (223.15,57.4) -- cycle ;
\draw  [fill={rgb, 255:red, 255; green, 255; blue, 255 }  ,fill opacity=1 ] (230.01,52.42) -- (234.98,52.42) -- (234.98,57.4) -- (230.01,57.4) -- cycle ;

\draw  [color={rgb, 255:red, 255; green, 255; blue, 255 }  ,draw opacity=1 ][fill={rgb, 255:red, 74; green, 74; blue, 74 }  ,fill opacity=1 ] (233.44,44.54) -- (238.41,44.54) -- (238.41,49.51) -- (233.44,49.51) -- cycle ;
\draw  [color={rgb, 255:red, 255; green, 255; blue, 255 }  ,draw opacity=1 ][fill={rgb, 255:red, 74; green, 74; blue, 74 }  ,fill opacity=1 ] (226.58,44.54) -- (231.55,44.54) -- (231.55,49.51) -- (226.58,49.51) -- cycle ;

\draw  [color={rgb, 255:red, 255; green, 255; blue, 255 }  ,draw opacity=1 ][fill={rgb, 255:red, 74; green, 74; blue, 74 }  ,fill opacity=1 ] (230.01,60.01) -- (234.98,60.01) -- (234.98,64.98) -- (230.01,64.98) -- cycle ;
\draw    (228.92,35.05) -- (228.92,41.54) ;
\draw [shift={(228.92,44.54)}, rotate = 270] [fill={rgb, 255:red, 0; green, 0; blue, 0 }  ][line width=0.08]  [draw opacity=0] (5.36,-2.57) -- (0,0) -- (5.36,2.57) -- (3.56,0) -- cycle    ;
\draw    (235.85,34.83) -- (235.85,41.62) ;
\draw [shift={(235.85,44.62)}, rotate = 270] [fill={rgb, 255:red, 0; green, 0; blue, 0 }  ][line width=0.08]  [draw opacity=0] (5.36,-2.57) -- (0,0) -- (5.36,2.57) -- (3.56,0) -- cycle    ;
\draw    (232.5,62.5) -- (232.5,69.05) ;
\draw [shift={(232.5,72.05)}, rotate = 270] [fill={rgb, 255:red, 0; green, 0; blue, 0 }  ][line width=0.08]  [draw opacity=0] (5.36,-2.57) -- (0,0) -- (5.36,2.57) -- (3.56,0) -- cycle    ;

\draw    (266.61,54.09) -- (266.61,62.5) ;
\draw    (263.18,47.03) -- (266.61,56.55) ;
\draw    (270.04,47.03) -- (266.61,56.55) ;
\draw    (270.04,47.03) -- (273.47,56.55) ;
\draw    (273.47,54.09) -- (266.61,62.5) ;
\draw    (259.75,54.09) -- (266.61,62.5) ;
\draw    (263.18,47.03) -- (259.75,56.55) ;
\draw  [fill={rgb, 255:red, 255; green, 255; blue, 255 }  ,fill opacity=1 ] (270.99,52.42) -- (275.96,52.42) -- (275.96,57.4) -- (270.99,57.4) -- cycle ;
\draw  [fill={rgb, 255:red, 255; green, 255; blue, 255 }  ,fill opacity=1 ] (257.26,52.42) -- (262.24,52.42) -- (262.24,57.4) -- (257.26,57.4) -- cycle ;
\draw  [fill={rgb, 255:red, 255; green, 255; blue, 255 }  ,fill opacity=1 ] (264.12,52.42) -- (269.1,52.42) -- (269.1,57.4) -- (264.12,57.4) -- cycle ;

\draw  [color={rgb, 255:red, 255; green, 255; blue, 255 }  ,draw opacity=1 ][fill={rgb, 255:red, 74; green, 74; blue, 74 }  ,fill opacity=1 ] (267.56,44.54) -- (272.53,44.54) -- (272.53,49.51) -- (267.56,49.51) -- cycle ;
\draw  [color={rgb, 255:red, 255; green, 255; blue, 255 }  ,draw opacity=1 ][fill={rgb, 255:red, 74; green, 74; blue, 74 }  ,fill opacity=1 ] (260.69,44.54) -- (265.67,44.54) -- (265.67,49.51) -- (260.69,49.51) -- cycle ;

\draw  [color={rgb, 255:red, 255; green, 255; blue, 255 }  ,draw opacity=1 ][fill={rgb, 255:red, 74; green, 74; blue, 74 }  ,fill opacity=1 ] (264.12,60.01) -- (269.1,60.01) -- (269.1,64.98) -- (264.12,64.98) -- cycle ;
\draw    (263.03,35.05) -- (263.03,41.54) ;
\draw [shift={(263.03,44.54)}, rotate = 270] [fill={rgb, 255:red, 0; green, 0; blue, 0 }  ][line width=0.08]  [draw opacity=0] (5.36,-2.57) -- (0,0) -- (5.36,2.57) -- (3.56,0) -- cycle    ;
\draw    (269.96,34.83) -- (269.96,41.62) ;
\draw [shift={(269.96,44.62)}, rotate = 270] [fill={rgb, 255:red, 0; green, 0; blue, 0 }  ][line width=0.08]  [draw opacity=0] (5.36,-2.57) -- (0,0) -- (5.36,2.57) -- (3.56,0) -- cycle    ;
\draw    (266.61,62.5) -- (266.61,69.05) ;
\draw [shift={(266.61,72.05)}, rotate = 270] [fill={rgb, 255:red, 0; green, 0; blue, 0 }  ][line width=0.08]  [draw opacity=0] (5.36,-2.57) -- (0,0) -- (5.36,2.57) -- (3.56,0) -- cycle    ;

\draw [color={rgb, 255:red, 74; green, 144; blue, 226 }  ,draw opacity=1 ]   (308.9,231.3) .. controls (297.86,222.91) and (310.16,212.9) .. (299.92,212.78) ;
\draw    [color={rgb, 255:red, 74; green, 144; blue, 226 }  ,draw opacity=1 ] (308.9,231.3) .. controls (297.68,239.43) and (309.76,249.73) .. (299.51,249.61) ;
\draw [color={rgb, 255:red, 74; green, 144; blue, 226 }  ,draw opacity=1 ][line width=1.5]    (313.67,257.75) -- (313.67,100.35) ;
\draw [shift={(313.67,97.35)}, rotate = 450] [color={rgb, 255:red, 74; green, 144; blue, 226 }  ,draw opacity=1 ][line width=1.5]    (9.95,-2.99) .. controls (6.32,-1.27) and (3.01,-0.27) .. (0,0) .. controls (3.01,0.27) and (6.32,1.27) .. (9.95,2.99)   ;
\draw [color={rgb, 255:red, 74; green, 144; blue, 226 }  ,draw opacity=1 ]   (104.37,42.66) -- (104.37,62.51) ;
\draw [shift={(104.37,40.66)}, rotate = 90] [color={rgb, 255:red, 74; green, 144; blue, 226 }  ,draw opacity=1 ][line width=0.75]    (6.56,-2.94) .. controls (4.17,-1.38) and (1.99,-0.4) .. (0,0) .. controls (1.99,0.4) and (4.17,1.38) .. (6.56,2.94)   ;
\draw [color={rgb, 255:red, 74; green, 144; blue, 226 }  ,draw opacity=1 ]   (165.35,43.16) -- (165.35,63.01) ;
\draw [shift={(165.35,41.16)}, rotate = 90] [color={rgb, 255:red, 74; green, 144; blue, 226 }  ,draw opacity=1 ][line width=0.75]    (6.56,-2.94) .. controls (4.17,-1.38) and (1.99,-0.4) .. (0,0) .. controls (1.99,0.4) and (4.17,1.38) .. (6.56,2.94)   ;
\draw [color={rgb, 255:red, 74; green, 144; blue, 226 }  ,draw opacity=1 ]   (190.28,42.16) -- (190.28,62.01) ;
\draw [shift={(190.28,40.16)}, rotate = 90] [color={rgb, 255:red, 74; green, 144; blue, 226 }  ,draw opacity=1 ][line width=0.75]    (6.56,-2.94) .. controls (4.17,-1.38) and (1.99,-0.4) .. (0,0) .. controls (1.99,0.4) and (4.17,1.38) .. (6.56,2.94)   ;
\draw [color={rgb, 255:red, 74; green, 144; blue, 226 }  ,draw opacity=1 ]   (244.44,42.16) -- (244.44,62.01) ;
\draw [shift={(244.44,40.16)}, rotate = 90] [color={rgb, 255:red, 74; green, 144; blue, 226 }  ,draw opacity=1 ][line width=0.75]    (6.56,-2.94) .. controls (4.17,-1.38) and (1.99,-0.4) .. (0,0) .. controls (1.99,0.4) and (4.17,1.38) .. (6.56,2.94)   ;
\draw [color={rgb, 255:red, 74; green, 144; blue, 226 }  ,draw opacity=1 ]   (278.67,42.49) -- (278.67,62.35) ;
\draw [shift={(278.67,40.49)}, rotate = 90] [color={rgb, 255:red, 74; green, 144; blue, 226 }  ,draw opacity=1 ][line width=0.75]    (6.56,-2.94) .. controls (4.17,-1.38) and (1.99,-0.4) .. (0,0) .. controls (1.99,0.4) and (4.17,1.38) .. (6.56,2.94)   ;
\draw  [color={rgb, 255:red, 255; green, 255; blue, 255 }  ,draw opacity=1 ][fill={rgb, 255:red, 74; green, 74; blue, 74 }  ,fill opacity=1 ] (107.45,298.55) .. controls (107.45,296.61) and (109.02,295.04) .. (110.96,295.04) .. controls (112.9,295.04) and (114.47,296.61) .. (114.47,298.55) .. controls (114.47,300.48) and (112.9,302.05) .. (110.96,302.05) .. controls (109.02,302.05) and (107.45,300.48) .. (107.45,298.55) -- cycle ;
\draw  [color={rgb, 255:red, 0; green, 0; blue, 0 }  ,draw opacity=1 ][fill={rgb, 255:red, 155; green, 155; blue, 155 }  ,fill opacity=1 ] (322.71,298.55) .. controls (322.71,297.23) and (323.78,296.16) .. (325.1,296.16) .. controls (326.41,296.16) and (327.48,297.23) .. (327.48,298.55) .. controls (327.48,299.86) and (326.41,300.93) .. (325.1,300.93) .. controls (323.78,300.93) and (322.71,299.86) .. (322.71,298.55) -- cycle ;
\draw [color={rgb, 255:red, 0; green, 0; blue, 0 }  ,draw opacity=1 ][line width=0.75]    (191.1,298.55) -- (179.77,298.55) ;
\draw   (37.29,287.88) -- (354.09,287.88) -- (354.09,309.21) -- (37.29,309.21) -- cycle ;
\draw  [color={rgb, 255:red, 255; green, 255; blue, 255 }  ,draw opacity=1 ][fill={rgb, 255:red, 74; green, 74; blue, 74 }  ,fill opacity=1 ] (46.85,292.8) -- (48.63,296.22) -- (52.62,296.77) -- (49.74,299.44) -- (50.42,303.2) -- (46.85,301.42) -- (43.28,303.2) -- (43.96,299.44) -- (41.07,296.77) -- (45.06,296.22) -- cycle ;
\draw [color={rgb, 255:red, 253; green, 184; blue, 99 }  ,draw opacity=1 ] [dash pattern={on 4.5pt off 4.5pt}]  (89.4,101.09) -- (269.3,101.09) ;
\draw  [fill={rgb, 255:red, 255; green, 255; blue, 255 }  ,fill opacity=1 ] (313.67,27.35) -- (348.67,62.35) -- (313.67,97.35) -- (278.67,62.35) -- cycle ;

\draw  [color={rgb, 255:red, 74; green, 144; blue, 226 }  ,draw opacity=1 ]   (308.9,180.3) .. controls (297.86,171.91) and (310.16,161.9) .. (299.92,161.78) ;
\draw   [color={rgb, 255:red, 74; green, 144; blue, 226 }  ,draw opacity=1 ]   (308.9,180.3) .. controls (297.68,188.43) and (309.76,198.73) .. (299.51,198.61) ;
\draw [color={rgb, 255:red, 0; green, 0; blue, 0 }  ,draw opacity=1 ][line width=0.75]    (258.08,298.55) -- (253.25,298.55) ;
\draw [color={rgb, 255:red, 0; green, 0; blue, 0 }  ,draw opacity=1 ][line width=0.75]    (264.52,298.55) -- (259.7,298.55) ;

\draw (73.26,178.18) node [anchor=north west][inner sep=0.75pt]  [font=\footnotesize]  {$\statevar_{i} \ $};
\draw (73.34,239.16) node [anchor=north west][inner sep=0.75pt]  [font=\footnotesize]  {$\statevar_{j} \ $};
\draw (56.22,131.18) node [anchor=north west][inner sep=0.75pt]  [font=\footnotesize]  {$\dot{\statevar}_{i} \ $};
\draw (55.14,88.16) node [anchor=north west][inner sep=0.75pt]  [font=\footnotesize]  {$\hat{u}_{i} \ $};
\draw (54.06,50.6) node [anchor=north west][inner sep=0.75pt]  [font=\scriptsize] [align=left] {NN};
\draw (92.42,270.61) node  [font=\tiny]  {$t_0$};
\draw (157.56,270.61) node  [font=\tiny]  {$\tau $};
\draw (182.96,270.61) node  [font=\tiny]  {$\tau \ +h$};
\draw (237.04,270.61) node  [font=\tiny]  {$\tau \ +2.7h$};
\draw (269.21,270.61) node  [font=\tiny]  {$T$};
\draw (60.02,259.69) node  [font=\scriptsize]  {$j$};
\draw (284,259.61) node [anchor=north west][inner sep=0.75pt]  [font=\tiny]  {$t$};
\draw (148.69,5.2) node [anchor=north west][inner sep=0.75pt]  [font=\footnotesize]  {ODESolve($\tv{\statevar}$, $t$, $T$, $f$, $\nnc{\tv{u}}(\tv{\statevar}(t); \tv{w})$)};
\draw (54.78,192.36) node  [font=\scriptsize]  {$i$};
\draw (274.72,162.47) node [anchor=north west][inner sep=0.75pt]  [font=\scriptsize]  {$\statevar^{*}_{i} \ $};
\draw (272.75,185.59) node [anchor=north west][inner sep=0.75pt]  [font=\scriptsize]  {$\statevar_{i}( T) \ $};
\draw (276.23,211.95) node [anchor=north west][inner sep=0.75pt]  [font=\scriptsize]  {$\statevar^{*}_{j} \ $};
\draw (270.88,234.3) node [anchor=north west][inner sep=0.75pt]  [font=\scriptsize]  {$\statevar_{j}( T) \ $};
\draw (40.29,163.79) node [anchor=north west][inner sep=0.75pt]  [font=\scriptsize] [align=left] {Graph};
\draw (316.53,62.06) node  [font=\scriptsize]  {$J\left(\ts{\statevar}_{t_0}^T, \tv{\statevar}^{*}\right)$};
\draw (316.84,235.58) node [anchor=north west][inner sep=0.75pt]  [font=\small,rotate=-269.45] [align=left] {backpropagation};
\draw (341.13,298.55) node  [font=\scriptsize] [align=left] {{\scriptsize time}};
\draw (77.69,298.55) node  [font=\scriptsize] [align=left] {{\scriptsize target state}};
\draw (149.14,298.55) node  [font=\scriptsize] [align=left] {{\scriptsize sampled state}};
\draw (223.34,298.55) node  [font=\scriptsize] [align=left] {{\scriptsize post-learning}};
\draw (291.3,298.55) node  [font=\scriptsize] [align=left] {{\scriptsize pre-learning}};

\end{tikzpicture}

             }
    \caption{A schematic that summarizes the training process of NODEC. 
    A NN learns the control within the ODESolve method.}
    \label{fig:nodec:framework}
\end{figure}
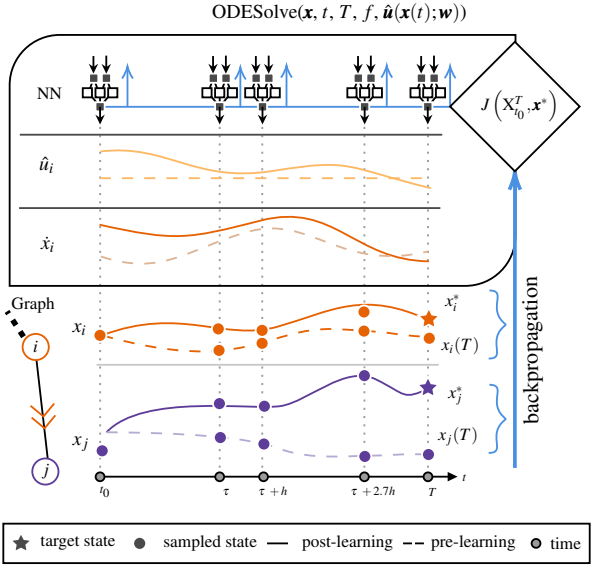

As in \Cref{sec:dynamical:systems}, we consider a dynamical system \eqref{eq:dynamical_system} with initial state $\tv{\statevar}(t_0)$, reached state $\tv{\statevar}(T)$, and target state $\tv{\statevar}^{*}$. 
The goal of NODEC is to minimize differences between $\tv{\statevar}(T)$ and $\tv{\statevar}^{*}$ using control inputs $\nnc{\tv{u}}(\tv{\statevar}(t), \tv{w})$, where the vector $\tv{w}$ represents the weights of an underlying neural network.
We quantify differences between reached and target states with the control loss function \tomcom{$J(\ts{\statevar}_{t_0}^T, \tv{\statevar}^*)$ over the control trajectory $\ts{\statevar}_{t_0}^T$.}
The general NODEC procedure is thus based on finding weights $\tv{w}$ that minimize a loss function $J(\ts{\statevar}_{t_0}^T, \tv{\statevar}^*)$ under the constraint \eqref{eq:dynamical_system}, using a gradient descent update rule over a certain number of epochs. That is
\begin{equation}
\label{eq:NODEC:problem}
\begin{aligned}
    \min_w & \,J(\ts{\statevar}_{t_0}^T, \tv{\statevar}^*;\tv {w}) & \\
    \text{s.t.}\
    \dot{\tv{\statevar}}(t)&= \tv{f}(t,\tv{\statevar}(t),\tv{u}(\tv{\statevar}(t))),
\end{aligned}
\end{equation}
where the control signal $\tv{u}(\tv{\statevar}(t)) = \nnc{\tv{u}}(\tv{\statevar}(t); \tv{w})$ is calculated as a neural network output and
\begin{equation}
\label{eq:NODEC:problem2}
\begin{aligned}
\tv{w} \leftarrow \tv{w}+\Delta \tv{w}\quad\text{with}\quad\Delta \tv{w} = -\eta \nabla_{\tv{w}} J(\ts{\statevar}_{t_0}^T, \tv{\statevar}^*;\tv {w}),
\end{aligned}
\end{equation}
where $\eta > 0$ denotes the learning rate parameter for trainning the NN.
Our proposed method relies on the usage of neural ODEs~\cite{chen2018neural}, which are a natural choice for the approximation of continuous-time control signals. Using neural ODEs instead of discrete-time controls allows us to approximate a continuous-time interaction and express the control function $\nnc{\tv{u}}(\tv{\statevar}(t); \tv{w})$ as a parameterized neural network within an ODE solver  (see \Cref{fig:nodec:framework}). 

The NODEC framework can be used to control both linear and non-linear graph dynamical systems with various loss functions.
Our approach is of particular relevance for continuous time control problems with unknown and intractable optimal control functions.
NODEC is based on universal approximation theorems for the approximation of continuous-time control functions with neural networks (NNs) and able to learn control inputs directly from the dynamics in an interactive manner akin to Reinforcement Learning (RL).
Contrary to other control approaches~\cite{yan2012controlling,fleming2006controlled,mangasarian1966sufficient,kamien1971sufficient,mcshane1989calculus}, we do not impose a control energy constraint directly on our optimization loss function, improving the learning efficiency considerably.\footnote{Imposing an energy constraint would require collecting and back-propagating the norm of all control inputs at each time step during training.
Using such a back-propagation scheme would increase training times considerably because of the potentially large number of control inputs in large-scale graph dynamical systems.}

In Algorithms 1 and 2, we show the two parts of a generic NODEC algorithm that approximates control signals. 
The main elements of NODEC are: 
(i) input and target states, 
(ii) graph coupled dynamics, 
(iii) neural network architecture and initialization, 
the parameters of the (iv) ODE solver (e.g., step-size) and of the 
(v) gradient descent algorithm and its hyper-parameters,  such as learning rate.
Note that Algorithm 2 relies on the automatic differentiation methods~\cite{baydin2017automatic,paszke2019pytorch}, where the gradients ``flow'' through the underlying neural network, that is time-unfolded by ODE solvers~\cite{shampine2018numerical}. 

\begin{figure}[!htb]
\removelatexerror
\begin{algorithm}[H]
\caption{A generic algorithm that describes the parameter learning of NODEC.}
 \label{algo:NODEC}
\footnotesize
\SetAlgoLined
\KwResult{$\tv{w}$}
\KwInit: $t_0, \tv{\statevar}_0, \tv{w}$, $\tv{f}(\cdot)$, $\text{ODESolve}(\cdot), J(\cdot), \target{\tv{\statevar}}$\;
\KwParams: $\eta$, epochs\;
epoch $\gets$ 0\;
\While{epoch $<$ epochs}{
    \tcp{Generate a trajectory based on NODEC.}
    $\ts{\statevar}_{t_0}^T \gets \text{ODESolve}(\tv{\statevar}_0, t_0, T, f, \nnc{\tv{u}}(\tv{\statevar}(t); \tv{w}))$\;
    \tcp{gradient descent update}
    $\tv{w}\gets \tv{w} - \eta \nabla_{\tv{w}} J(\ts{\statevar}_{t_0}^T, \target{\tv{\statevar}})$\;
    \tcp{ or Quasi-Newton with Hessian:}
    \tcp{$\tv{w}\gets \tv{w} - \eta H^{-1} \nabla_{\tv{w}} J(\ts{\statevar}_{t_0}^T, \target{\tv{\statevar}})$}
 }
 \end{algorithm}

\end{figure}
\begin{figure}[!htb]
\removelatexerror
\begin{algorithm}[H]
\caption{A simple ODESolve implementation.
 \label{algo:odesolve}
\vspace{0.1 cm}
}
\footnotesize
\SetAlgoLined
 \SetKwProg{Fn}{Function}{ }{end}
 \Fn{ODESolve($\tv{\statevar}(t_0)$, $t_0$, $T$, $f$, $\nnc{\tv{u}}(\tv{\statevar}(t); \tv{w})$): }{
  \tcp{Euler Method}
  $t \gets t_0$\;
  \tcp{State trajectory:a set of state vectors.}
  \KwSet $\ts{X} \gets \{\tv{\statevar}(t_0)\}$\; 
  \While{$t \leq T$}{
        \tcp{Computational graph is}
        \tcp{preserved through time}
        \tcp{gradients flow through $\tv{\statevar}$}
        $\nnc{\tv{u}} \gets \nnc{\tv{u}}(\tv{\statevar}(t); \tv{w})$\;
        $\tv{\statevar}\gets \tv{\statevar} + \tau \tv{f}(t, \tv{\statevar}, \nnc{\tv{u}})$\;
        $\ts{X}\gets\ts{X}\cup\{\tv{\statevar}\}$
        
        \tcp{Step $\tau$ could be adaptive}
        $t\gets t + \tau$\;
    }
        \KwRet $\ts{X}$\;}
 \vspace{0.29 cm}
\end{algorithm}

\end{figure}

\subsection{Neural ODE and NODEC Learning Settings}

Although NODEC utilizes neural ODEs~\cite{chen2018neural}, the learning tasks of both frameworks differ significantly. 
Neural ODEs~\cite{chen2018neural} model dynamics of the hidden state $h(t)$ of a neural network according to
\begin{equation}
\dot{\tv{h}}(t)=\tv{\mathcal{f}}(t,\tv{h}(t),\tv{w}),
\label{eq:NeuralODE}
\end{equation}
where $\tv{\mathcal{f}}(\tv{h}(t),t,\tv{w})$ and $\dot{\tv{h}}(t)$ represent the neural network and hidden-state derivative/dynamics, respectively. 
Previously, neural ODEs were mainly applied in supervised learning tasks~\cite{kidger2020neural} and in normalizing flows~\cite{chen2018neural}.  
For NODEC, we use a neural network as a parameterized function to approximate the control term $\tv{u}(\tv{\statevar}(t))$ in graph dynamical systems \eqref{eq:dynamical_system}. Contrary to supervised applications of neural ODEs~\cite{chen2018neural}, our proposed framework numerically solves control problems in an interactive manner, similar to reinforcement learning. 

\subsection{Learnability of Control with Neural Networks}
\label{sec:learnability}
As reachability of a target state $\target{\tv{\statevar}}$ from an initial state $\tv{\statevar}(t_0)$ implies the existence of a control function $\tv{u}(\tv{\statevar}(t))$, we now focus on the ability to approximate (i.e., learn) $\tv{u}(\tv{\statevar}(t))$ for reachable target states with a neural network. 

\begin{prop}
\label{cor:universal}
Given that (i) a target state $\target{\tv{\statevar}}$ is reachable with continuous time dynamics \eqref{eq:dynamical_system} and (ii) the control function $\tv{u}(\tv{\statevar}(t))$ that reaches the target state $\target{\tv{\statevar}}$ is continuous or Lebesque integrable in its domain, then a corresponding universal approximation (UA) theorem applies for a neural network that can approximate a control function $\nnc{\tv{u}}(\tv{\statevar}(t); \tv{w}) \to \tv{u}(\tv{\statevar}(t))$ by learning parameters $\tv{w}$.
\end{prop}
The above proposition holds when an appropriate UA theorem~\cite{stone1948generalized,zhou2020universality,schafer2006recurrent, sontag1995computational,hanin2017approximating} and reachability~\cite{lygeros2004reachability} requirements are satisfied by the underlying dynamics and the neural network controller.
\tomcom{Related work indicates UA properties for neural ODEs~\cite{teshima2020universal} that can be leveraged to calculate control trajectories that successfully optimize \Cref{eq:NODEC:problem}.}
The ability of a neural network to learn control signals has also been covered in the literature outside of the domain of neural ODEs~\cite{bucci2019control,hua2004adaptive,yoo2005stable,bhasin2013novel}. 
In the current work we choose to compare our proposed model to an analytical feedback control baseline~\cite{skardal2015control} and state-of-the-art reinforcement learning~\cite{fujimoto2018addressing} for non-linear dynamical systems describing for Kuramoto oscillators and disease spreading.
 
\subsection{Learning Loss and Control Goals}
\label{sec:losses:goals}
To apply NODEC to control tasks, we have to translate a control goal into an adequate learning loss.
The choice of the control goal depends on the underlying dynamics, graph structure, and objectives of the control designer. 
A very common goal in literature~\cite{liptak2018instrument} is ``microscopic'' control where each node $i$ has to reach a predetermined state value within time $T$, i.e.~$\statevar_i(T) = \statevar_i^{*}$. 
Such a control goal is often applied in industrial applications and may be used to steer electric and mechanical systems~\cite{liptak2018instrument}. 
This goal is achieved by minimizing a metric that quantifies the distance between the target and reached states $\tv{\statevar}(T)$ and $\tv{\statevar}^*$. 
One possible choice of such a metric is the mean squared error (MSE) $J(\tv{\statevar}(T), \tv{\statevar}^{*}) = \frac{1}{N}\sum_{i=1}^N (\statevar_i(T)-\target{\statevar_i})^2$.
When the MSE is used, corresponding optimal control problems may be expressed as convex optimization problems~\cite{bemporad2002explicit}. 
For more details on the application of NODEC to microscopic loss function, see Ref.~\cite{bottcher2021implicit}.

We focus on control goals that do not require a specific target state value for each node, but instead require that constraints over aggregate values or statistical properties of the system's states are satisfied.
For the control of certain complex systems, it is useful to consider such ``macroscopic'' constraints~\cite{sethna2006statistical,barrat2008dynamical}.
Often such goals lack exact optimal control solutions, thus offering many opportunities for novel control applications of NODEC.

A common macroscopic control goal is that nodes in the target state are required to be synchronized, i.e.~the nodes' states are required to have the same value or constant phase shifts. 
Such synchronization conditions are often considered in the context of controlling oscillator systems~\cite{biccari2020stochastic,brede2008locals}.
\tomcom{When synchronizing oscillators reaching the target state at time $T$ may not satisfy the control goal completely, as we may require the system to preserve the state properties that satisfy the goal for a longer time $\ts{\statevar}_{t}^T$, e.g. keeping the oscillators synchronized for a time period.
In that case without loss of generality, we assume that the system goal requires that the state trajectory $\ts{\statevar}_{t_0}^{T}$ reached within a time period $t_0 \leq t \leq T$ satisfies some target state properties measured by the control loss function $J(\ts{\statevar}_{t_0}^{T}, \target{\tv{\statevar}})$.}
In the current work we showcase that NODEC can be perfectly adapted to optimize such goals.

We also consider more complex control goals, when the system evolution includes coupled ODEs with more than one state variable, such as disease spreading dynamics.
\tomcom{In the context of epidemic models, the state $\tv{\statevar}_i$ of a node $i$ is represented by a vector, that consists of multiple state variables.}
For susceptible-infected-recovered (SIR) models, three state variables, $S_i(t)$, $I_i(t)$, and $R_i(t)$, 
are used to model the part of a population on node $i$ at time $t$ that is susceptible, infected, and recovered, respectively.
A relevant control goal for controlling epidemics is the ``flattening'' of the curve, or reducing the maximum infected population that occurs at time $\target{t} \in [t0, T]$.
In this case, the target time $\target{t} \in [t_0,T]$ at which the control target state that satisfies the control loss is unknown and becomes part of the control problem.

\section{Experimental Evaluation}\label{sec:results}
In this section, we evaluate the ability of NODEC to 
(i) reach target states efficiently with a limited number of driver nodes, 
(ii) control different dynamics and losses, and
(iii) calculate low energy control signals.
We first evaluate the performance of NODEC for two non-linear systems with very different control tasks to showcase its applicability and versatility in computationally challenging settings for which analytical solutions or approximate control schemes may not exist.
We describe the experimental setup by defining the dynamical systems, initial state, control goal, and neural-network hyper-parameters used for training. 
The choice of neural-network hyper-parameters focuses mainly on the network architecture, inputs, optimizers, and training procedures. 
For the sake of brevity, we omit technical details in the main text and provide further information in the Supplemental Material and in our code~\cite{nncgit,nodecocean} and data repositories~\cite{nodecdataport}.
\subsection{Coupled Oscillator Dynamics}
\label{sec:osc}
Here we study the ability of NODEC to control a network of coupled oscillators via feedback control.
Such systems are used to model power grids and brain networks~\cite{cumin2007generalising,dorfler2012synchronization}.
One common control goal for oscillator systems is to reach a fully synchronized target state and stabilize the system over time.
This introduces two main challenges: (i) a target state that satisfies this goal needs to be reached and preserved and (ii) the trained model needs to be able to achieve synchronization stability for initial states not seen in training.
For continuous time linear time invariant systems and systems that can be linearized, there exist optimal feedback control methods~\cite{schoenwald1992optimal}.
Continuous-time oscillatory dynamics may not always be linearizable~\cite{skardal2015control} and exhibit chaotic behavior~\cite{bick2018chaos,maistrenko2005chaotic}, which cannot be observed in (finite-dimensional) LTI systems. 
NODEC does not require linearization and could potentially control systems that are costly or intractable to linearize.

In a graph of $N$ coupled oscillators, a possible mathematical description of the evolution of phase $\statevar_i$ of oscillator $i$ with natural frequency $\omega_i$ is
\begin{equation}
    \begin{aligned}
    \dot{\statevar}_i = \omega_i + \sum_m \matr{B}_{i,m}u_{m}\left(\tv{\statevar}(t)\right) + K\sum_{j}\matr{A}_{i,j}\mathcal{h}(\statevar_j - \statevar_i)
    \end{aligned}
\label{eq:oscillator}
\end{equation}
where $\matr{A}$ is the interaction matrix, $K$ the coupling constant, and $\mathcal{h}$ a $2\pi$-periodic function~\cite{skardal2015control}.
For $2\pi$-periodic oscillator dynamics \eqref{eq:oscillator}, 
optimal feedback control can be achieved via linearization near the synchronized steady state\footnote{which exists in a rotating reference frame.}~\cite{skardal2015control} and are known to work only for low values of coupling frustration~\cite{skardal2015control}.
\begin{equation}
    \tv \statevar^{\diamond} = K^{-1} L^{\dagger} \tv{\omega},
\end{equation}
where $L^{\dagger}$ is the pseudo-inverse of the graph Laplacian and $\tv{\omega} = [\omega_1,\ldots,\omega_N]$ is the vector of natural frequencies.

To study the performance of NODEC, we consider the Kuramoto model~\cite{kuramoto1975self}
\begin{equation}
    \label{eq:kuramoto}
    \begin{aligned}
    \dot{\statevar}_i = \omega_i +  \sum_m \matr{B}_{i,m}u_{m}\left(\tv{\statevar}(t)\right) + K\sum_{j}\matr{A}_{i,j}\sin(\statevar_j - \statevar_i)
    \end{aligned}
\end{equation}
as a specific example of a model of coupled oscillators in a network.

\subsubsection{Control Baselines}\label{sec:osc:baseline}
A feedback control (FC) baseline for Kuramoto dynamics is presented in Ref.~\cite{skardal2015control}.
\tomcom{
First, the feedback control gain vector $\tv{b}^{(\textrm{FC})}$ is defined for the control baseline.
An element of the gain control vector $b^{(\textrm{FC})}_{i}$ is assigned to a graph node $i$ and needs to satisfy
\begin{equation}
    \label{eq:kuramoto:driver:selection}
    \footnotesize
    b^{(\textrm{FC})}_{i} \geq \sum_{j\neq i} \left[|K\matr{A}_{i,j}\cos(\statevar_i^{\diamond}-\statevar_j^{\diamond})-\epsilon| - (K\matr{A}_{i,j}\cos(\statevar_j^{\diamond}-\statevar_i^{\diamond}) -\epsilon)\right].
\end{equation}
We take the equality of the constraint in \Cref{eq:kuramoto:driver:selection} to calculate the control gain coefficients $b^{(\textrm{FC})}_{i}$ based on Ref. \cite{skardal2015control}.
Non-zero values $b^{(\textrm{FC})}_{i} \neq 0$ determine the driver nodes.
The baseline control signal $u_{i}$ for a node $i$ is calculated as
\begin{equation}
    u_{i}(\statevar_i(t)) = \zeta b^{(\textrm{FC})}_{i} \sin(\target{\statevar_{i}}-\statevar_i(t)).
\end{equation}
We note that here we follow the notation of Ref.~\cite{skardal2015control} and use a control gain vector $\tv{b}$ instead of a driver matrix. 
If we prefer to use the driver matrix notation, we iterate over all nodes and select a node $i$ as the $m\textrm{-th}$ driver node by setting the driver matrix element $\matr{B}_{i,m} = b^{\textrm{FC}}_{i}$ if $b^{\textrm{FC}}_{i} \neq 0$.
}

An error margin buffer is also implemented as suggested in related work~\cite{skardal2015control} by setting $\epsilon \geq 0$ when selecting driver nodes in \Cref{eq:kuramoto:driver:selection}.
For $\epsilon = 0$, the driver node selection might be insufficient and it may not be possible to drive the system to a desired target state~\cite{skardal2015control}.
Using an error margin buffer increases the driver node selection tolerance and, thus, selects more diver nodes, which can steer the system to a desired target state.
The non-zero values of the driver matrix can be chosen arbitrarily, as long as the constraint in \Cref{eq:kuramoto:driver:selection} is satisfied.

We require that feedback control reaches comparable performance to NODEC in terms of $r(t)$, thus we multiply the vector $\tv{b}^{(\textrm{FC})}$ with a positive scalar value $\zeta=10$\footnote{We tested several other values before selecting the specific value.}.
Higher absolute values of $\zeta|b^{(\textrm{FC})}_{i}|$ may create control signals that reach the target state in less time at the expense of a higher control energy.
As the driver matrix is calculated based on an approximation of the graph Laplacian pseudo-inverse $L^\dagger$ of a singular system, optimal control guarantees for minimum energy may not always hold.
The target state in \Cref{eq:kuramoto:driver:selection} is set to $\target{\statevar_{i}} = 0$. 

\subsubsection{Numerical Simulation Parameters}
\label{sec:osc:eval}
The control goal is to reach a synchronized state with zero phase difference $\target{\statevar_i}-\target{\statevar_j} = 0$.
To evaluate the system synchronicity, we calculate the order parameter (see \Cref{eq:kuramoto:order}), which reaches the maximum value $r(t) = 1$ if all oscillators are fully synchronized.

For our numerical experiments, we create an Erd\H{o}s--R\'enyi graph $G(N,p)$ with $N=1024$ nodes, mean degree $\overline{d} \approx 6$, and link probability $p=\overline{d}/(N-1)$. 
We generate the driver matrix as in \Cref{sec:osc:baseline}, and select the non-zero elements as driver nodes.
To reduce approximation errors due to the inversion of the Laplacian matrix, we set a buffer margin of $\epsilon = 0.1$ when selecting driver nodes.
Control signal energy is evaluated with \Cref{eq:energy:approx}.
Moreover, we set the coupling constant to $K=0.4$ and sample the natural frequencies $\omega_i$  from a uniform distribution $\mathcal{U}(-\sqrt{3},\sqrt{3})$~\cite{skardal2016controlling}.
This setting results in approximately $70\%$ of the nodes being assigned as driver nodes.

\subsubsection{NODEC Hyperparameters}
\label{sec:osc:training}
\begin{figure}
    \centering
        \subfloat[Kuramoto Controller network.]{
          \label{fig:osci:arch}
            \resizebox{0.27\columnwidth}{!}{

\tikzset{every picture/.style={line width=0.75pt}} 
\begin{tikzpicture}[x=0.75pt,y=0.75pt,yscale=-1,xscale=1]

\draw  [line width=0.75]  (160.11,182.05) -- (187.2,182.05) -- (187.2,209.18) -- (160.11,209.18) -- cycle ;
\draw [line width=0.75]    (161.98,202.63) .. controls (173.87,201.7) and (173.87,201.55) .. (177.46,197) .. controls (181.05,192.45) and (178.95,195.53) .. (184.4,187.36) ;

\draw  [color={rgb, 255:red, 255; green, 255; blue, 255 }  ,draw opacity=1 ][fill={rgb, 255:red, 128; green, 128; blue, 128 }  ,fill opacity=1 ] (106.11,237.23) -- (133.2,237.23) -- (133.2,264.36) -- (106.11,264.36) -- cycle ;
\draw [color={rgb, 255:red, 74; green, 74; blue, 74 }  ,draw opacity=1 ][line width=0.75]    (160.65,175.16) -- (173.66,158.04) ;
\draw    (186.66,175.16) -- (173.66,158.04) ;
\draw    (179.09,178.85) -- (173.66,158.04) ;
\draw    (168.22,178.85) -- (173.66,158.04) ;

\draw  [line width=0.75]  (160.11,75.89) -- (187.2,75.89) -- (187.2,103.02) -- (160.11,103.02) -- cycle ;
\draw [line width=0.75]    (181.91,81.2) -- (165.41,97.7) ;

\draw [color={rgb, 255:red, 74; green, 74; blue, 74 }  ,draw opacity=1 ][line width=0.75]    (160.65,228.5) -- (173.66,211.38) ;
\draw    (186.66,228.5) -- (173.66,211.38) ;
\draw    (179.09,232.19) -- (173.66,211.38) ;
\draw    (168.22,232.19) -- (173.66,211.38) ;

\draw  [color={rgb, 255:red, 255; green, 255; blue, 255 }  ,draw opacity=1 ][fill={rgb, 255:red, 74; green, 74; blue, 74 }  ,fill opacity=1 ] (213.11,73.73) -- (240.2,73.73) -- (240.2,100.86) -- (213.11,100.86) -- cycle ;

\draw    (136.03,251.63) -- (150.28,251.63) ;
\draw [shift={(153.28,251.63)}, rotate = 180] [fill={rgb, 255:red, 0; green, 0; blue, 0 }  ][line width=0.08]  [draw opacity=0] (8.93,-4.29) -- (0,0) -- (8.93,4.29) -- cycle    ;
\draw  [line width=0.75]  (160.11,237.39) -- (187.2,237.39) -- (187.2,264.52) -- (160.11,264.52) -- cycle ;
\draw    (164.35,250.77) .. controls (167.03,248.83) and (167.33,248.01) .. (169.26,248.01) .. controls (171.2,248.01) and (174.92,253.89) .. (177.9,253.89) .. controls (180.88,253.89) and (181.32,252.33) .. (182.96,250.84) ;

\draw  [line width=0.75]  (160.11,128.55) -- (187.2,128.55) -- (187.2,155.68) -- (160.11,155.68) -- cycle ;
\draw [line width=0.75]    (161.98,149.13) .. controls (173.87,148.2) and (173.87,148.05) .. (177.46,143.5) .. controls (181.05,138.95) and (178.95,142.03) .. (184.4,133.86) ;

\draw [color={rgb, 255:red, 74; green, 74; blue, 74 }  ,draw opacity=1 ][line width=0.75]    (160.65,121.16) -- (173.66,104.04) ;
\draw    (186.66,121.16) -- (173.66,104.04) ;
\draw    (179.09,124.85) -- (173.66,104.04) ;
\draw    (168.22,124.85) -- (173.66,104.04) ;

\draw    (192.03,87.63) -- (206.28,87.63) ;
\draw [shift={(209.28,87.63)}, rotate = 180] [fill={rgb, 255:red, 0; green, 0; blue, 0 }  ][line width=0.08]  [draw opacity=0] (8.93,-4.29) -- (0,0) -- (8.93,4.29) -- cycle    ;
\draw    (40.55,158.78) -- (40.55,174.78) ;
\draw    (34.02,140.65) -- (40.55,158.78) ;
\draw    (47.08,140.65) -- (40.55,158.78) ;
\draw    (47.08,140.65) -- (53.62,158.78) ;
\draw    (53.62,158.78) -- (40.55,174.78) ;
\draw    (27.49,158.78) -- (40.55,174.78) ;
\draw    (34.02,140.65) -- (27.49,158.78) ;
\draw  [fill={rgb, 255:red, 255; green, 255; blue, 255 }  ,fill opacity=1 ] (48.88,154.04) -- (58.35,154.04) -- (58.35,163.52) -- (48.88,163.52) -- cycle ;
\draw  [fill={rgb, 255:red, 255; green, 255; blue, 255 }  ,fill opacity=1 ] (22.75,154.04) -- (32.22,154.04) -- (32.22,163.52) -- (22.75,163.52) -- cycle ;
\draw  [fill={rgb, 255:red, 255; green, 255; blue, 255 }  ,fill opacity=1 ] (35.82,154.04) -- (45.29,154.04) -- (45.29,163.52) -- (35.82,163.52) -- cycle ;

\draw  [color={rgb, 255:red, 255; green, 255; blue, 255 }  ,draw opacity=1 ][fill={rgb, 255:red, 74; green, 74; blue, 74 }  ,fill opacity=1 ] (42.35,135.91) -- (51.82,135.91) -- (51.82,145.39) -- (42.35,145.39) -- cycle ;
\draw  [color={rgb, 255:red, 255; green, 255; blue, 255 }  ,draw opacity=1 ][fill={rgb, 255:red, 74; green, 74; blue, 74 }  ,fill opacity=1 ] (29.28,135.91) -- (38.76,135.91) -- (38.76,145.39) -- (29.28,145.39) -- cycle ;

\draw  [color={rgb, 255:red, 255; green, 255; blue, 255 }  ,draw opacity=1 ][fill={rgb, 255:red, 74; green, 74; blue, 74 }  ,fill opacity=1 ] (35.82,170.04) -- (45.29,170.04) -- (45.29,179.51) -- (35.82,179.51) -- cycle ;
\draw    (33.74,126.65) -- (33.74,132.91) ;
\draw [shift={(33.74,135.91)}, rotate = 270] [fill={rgb, 255:red, 0; green, 0; blue, 0 }  ][line width=0.08]  [draw opacity=0] (5.36,-2.57) -- (0,0) -- (5.36,2.57) -- (3.56,0) -- cycle    ;
\draw    (46.93,126.8) -- (46.93,133.06) ;
\draw [shift={(46.93,136.06)}, rotate = 270] [fill={rgb, 255:red, 0; green, 0; blue, 0 }  ][line width=0.08]  [draw opacity=0] (5.36,-2.57) -- (0,0) -- (5.36,2.57) -- (3.56,0) -- cycle    ;
\draw    (40.78,179.74) -- (40.78,186) ;
\draw [shift={(40.78,189)}, rotate = 270] [fill={rgb, 255:red, 0; green, 0; blue, 0 }  ][line width=0.08]  [draw opacity=0] (5.36,-2.57) -- (0,0) -- (5.36,2.57) -- (3.56,0) -- cycle    ;

\draw (131.82,225.5) node [anchor=east] [inner sep=0.75pt]    {$( N)$};
\draw (155.82,141) node [anchor=east] [inner sep=0.75pt]    {$( 3)$};
\draw (155.82,194) node [anchor=east] [inner sep=0.75pt]    {$( N)$};
\draw (119.66,250.79) node  [color={rgb, 255:red, 255; green, 255; blue, 255 }  ,opacity=1 ]  {$\statevar$};
\draw (156.82,87) node [anchor=east] [inner sep=0.75pt]    {$( 3)$};
\draw (241.82,112) node [anchor=east] [inner sep=0.75pt]    {$( M)$};
\draw (226.66,87.29) node  [font=\footnotesize,color={rgb, 255:red, 255; green, 255; blue, 255 }  ,opacity=1 ]  {$\hat{\mathbf{u}}$};
\draw (189.2,240.79) node [anchor=north west][inner sep=0.75pt]    {$sin( \cdot )$};
\draw (86.64,157.83) node  [font=\LARGE]  {$=$};

\draw (10,0.73) node [anchor=north west][inner sep=0.75pt]    {\LARGE $J(\cdot)\ =\ -\dfrac{1}{\Xi}\sum ^{\Xi}_{\xi=1} r(\xi\tau)$ 
};

\end{tikzpicture}

            }
        }
        \subfloat[Legend of neural networks operators.]{
          \label{fig:nn:legend}
            \resizebox{0.7\columnwidth}{!}{

\tikzset{every picture/.style={line width=0.75pt}} 
\begin{tikzpicture}[x=0.75pt,y=0.75pt,yscale=-1,xscale=1]

\draw  [line width=0.75]  (134.77,72.52) -- (155.61,72.52) -- (155.61,93.4) -- (134.77,93.4) -- cycle ;
\draw [line width=0.75]    (136.2,88.36) .. controls (145.35,87.64) and (145.35,87.53) .. (148.12,84.02) .. controls (150.88,80.52) and (149.26,82.9) .. (153.46,76.61) ;

\draw  [line width=0.75]  (134.77,22.06) -- (155.61,22.06) -- (155.61,42.94) -- (134.77,42.94) -- cycle ;
\draw [line width=0.75]    (151.54,26.15) -- (138.84,38.85) ;

\draw [line width=0.75]    (152.59,52.25) -- (146.8,61.1) -- (137.79,61.33) ;
\draw  [line width=0.75]  (134.77,46.35) -- (155.61,46.35) -- (155.61,67.23) -- (134.77,67.23) -- cycle ;

\draw  [line width=0.75]  (134.77,99.66) -- (155.61,99.66) -- (155.61,120.53) -- (134.77,120.53) -- cycle ;
\draw    (137.22,115.39) .. controls (140.86,115.91) and (142.25,106.43) .. (145.55,106.28) .. controls (148.85,106.13) and (151.02,116.7) .. (153.88,115.39) ;

\draw    (15.13,82.99) .. controls (7.69,77.33) and (15.99,70.58) .. (9.07,70.49) ;
\draw    (15.13,82.99) .. controls (7.56,88.48) and (15.71,95.43) .. (8.8,95.35) ;

\draw  [line width=1.5]  (24.6,79.73) -- (21.24,83.23) -- (17.9,79.71) ;
\draw  [line width=1.5]  (17.87,86.72) -- (21.24,83.23) -- (24.58,86.75) ;

\draw [color={rgb, 255:red, 74; green, 74; blue, 74 }  ,draw opacity=1 ][line width=0.75]    (4.59,37.66) -- (14.6,24.49) ;
\draw    (24.6,37.66) -- (14.6,24.49) ;
\draw    (18.78,40.5) -- (14.6,24.49) ;
\draw    (10.42,40.5) -- (14.6,24.49) ;

\draw  [fill={rgb, 255:red, 155; green, 155; blue, 155 }  ,fill opacity=1 ] (3.8,61.17) -- (10.73,61.17) -- (10.73,68.11) -- (3.8,68.11) -- cycle(9.69,62.21) -- (4.84,62.21) -- (4.84,67.07) -- (9.69,67.07) -- cycle ;
\draw  [fill={rgb, 255:red, 155; green, 155; blue, 155 }  ,fill opacity=1 ] (10.73,61.17) -- (17.67,61.17) -- (17.67,68.11) -- (10.73,68.11) -- cycle(16.63,62.21) -- (11.77,62.21) -- (11.77,67.07) -- (16.63,67.07) -- cycle ;
\draw  [fill={rgb, 255:red, 155; green, 155; blue, 155 }  ,fill opacity=1 ] (17.67,61.17) -- (24.6,61.17) -- (24.6,68.11) -- (17.67,68.11) -- cycle(23.56,62.21) -- (18.71,62.21) -- (18.71,67.07) -- (23.56,67.07) -- cycle ;

\draw [line width=0.75]  [dash pattern={on 0.84pt off 2.51pt}]  (9.84,53.67) -- (7.47,60.75) ;
\draw  [dash pattern={on 0.84pt off 2.51pt}]  (18.56,53.67) -- (20.93,60.75) ;

\draw [line width=0.75]  [dash pattern={on 0.84pt off 2.51pt}]  (12.24,52.87) -- (9.87,59.95) ;
\draw  [dash pattern={on 0.84pt off 2.51pt}]  (20.96,52.87) -- (23.33,59.95) ;

\draw [line width=0.75]  [dash pattern={on 0.84pt off 2.51pt}]  (7.44,52.87) -- (5.07,59.95) ;
\draw  [dash pattern={on 0.84pt off 2.51pt}]  (16.16,52.87) -- (18.53,59.95) ;

\draw  [fill={rgb, 255:red, 155; green, 155; blue, 155 }  ,fill opacity=1 ] (7.27,45.46) -- (14.2,45.46) -- (14.2,52.41) -- (7.27,52.41) -- cycle(13.16,46.51) -- (8.31,46.51) -- (8.31,51.37) -- (13.16,51.37) -- cycle ;
\draw  [fill={rgb, 255:red, 155; green, 155; blue, 155 }  ,fill opacity=1 ] (14.2,45.46) -- (21.14,45.46) -- (21.14,52.41) -- (14.2,52.41) -- cycle(20.1,46.51) -- (15.24,46.51) -- (15.24,51.37) -- (20.1,51.37) -- cycle ;

\draw  [line width=0.75]  (3.76,99.11) -- (24.6,99.11) -- (24.6,119.99) -- (3.76,119.99) -- cycle ;
\draw   (20.38,108.27) .. controls (20.5,107.38) and (20.11,106.51) .. (19.38,106.01) .. controls (18.65,105.51) and (17.71,105.49) .. (16.95,105.94) .. controls (16.69,105.42) and (16.2,105.07) .. (15.63,104.98) .. controls (15.07,104.89) and (14.49,105.08) .. (14.09,105.49) .. controls (13.86,105.02) and (13.41,104.71) .. (12.9,104.66) .. controls (12.39,104.61) and (11.9,104.83) .. (11.59,105.25) .. controls (11.17,104.75) and (10.52,104.54) .. (9.9,104.71) .. controls (9.29,104.88) and (8.82,105.39) .. (8.71,106.03) .. controls (8.2,106.17) and (7.78,106.53) .. (7.56,107.01) .. controls (7.33,107.5) and (7.32,108.06) .. (7.52,108.55) .. controls (7.03,109.21) and (6.91,110.09) .. (7.22,110.86) .. controls (7.52,111.64) and (8.21,112.18) .. (9.02,112.3) .. controls (9.02,113.03) and (9.41,113.69) .. (10.04,114.04) .. controls (10.66,114.39) and (11.42,114.37) .. (12.02,113.98) .. controls (12.28,114.85) and (13,115.49) .. (13.87,115.62) .. controls (14.75,115.76) and (15.62,115.36) .. (16.11,114.61) .. controls (16.72,114.98) and (17.44,115.09) .. (18.12,114.9) .. controls (18.8,114.72) and (19.38,114.27) .. (19.73,113.64) .. controls (20.35,113.72) and (20.95,113.39) .. (21.23,112.83) .. controls (21.5,112.26) and (21.41,111.58) .. (20.98,111.12) .. controls (21.53,110.79) and (21.82,110.13) .. (21.68,109.49) .. controls (21.55,108.85) and (21.03,108.37) .. (20.39,108.31) ; \draw   (20.98,111.12) .. controls (20.73,111.27) and (20.43,111.34) .. (20.13,111.32)(19.73,113.64) .. controls (19.6,113.63) and (19.48,113.59) .. (19.36,113.55)(16.11,114.61) .. controls (16.2,114.47) and (16.28,114.32) .. (16.34,114.16)(12.02,113.98) .. controls (11.97,113.83) and (11.94,113.66) .. (11.93,113.5)(9.02,112.3) .. controls (9.01,111.53) and (9.44,110.82) .. (10.12,110.49)(7.52,108.55) .. controls (7.63,108.81) and (7.8,109.04) .. (8.01,109.23)(8.71,106.03) .. controls (8.69,106.14) and (8.68,106.25) .. (8.68,106.36)(11.59,105.25) .. controls (11.69,105.37) and (11.77,105.51) .. (11.84,105.66)(14.09,105.49) .. controls (14.14,105.6) and (14.18,105.72) .. (14.21,105.84)(16.95,105.94) .. controls (16.79,106.03) and (16.64,106.15) .. (16.51,106.28)(20.38,108.27) .. controls (20.36,108.39) and (20.34,108.51) .. (20.3,108.63) ;
\draw  [fill={rgb, 255:red, 0; green, 0; blue, 0 }  ,fill opacity=1 ] (18.48,111.15) .. controls (18.48,111.76) and (17.99,112.25) .. (17.38,112.25) .. controls (16.77,112.25) and (16.28,111.76) .. (16.28,111.15) .. controls (16.28,110.54) and (16.77,110.05) .. (17.38,110.05) .. controls (17.99,110.05) and (18.48,110.54) .. (18.48,111.15) -- cycle ;
\draw  [fill={rgb, 255:red, 0; green, 0; blue, 0 }  ,fill opacity=1 ] (12.88,108.15) .. controls (12.88,108.76) and (12.39,109.25) .. (11.78,109.25) .. controls (11.17,109.25) and (10.68,108.76) .. (10.68,108.15) .. controls (10.68,107.54) and (11.17,107.05) .. (11.78,107.05) .. controls (12.39,107.05) and (12.88,107.54) .. (12.88,108.15) -- cycle ;
\draw    (11.78,108.15) -- (17.38,111.15) ;

\draw  [color={rgb, 255:red, 255; green, 255; blue, 255 }  ,draw opacity=1 ][fill={rgb, 255:red, 128; green, 128; blue, 128 }  ,fill opacity=1 ] (243.16,72.52) -- (264.01,72.52) -- (264.01,93.4) -- (243.16,93.4) -- cycle ;
\draw  [line width=0.75]  (243.63,22.21) -- (264.48,22.21) -- (264.48,43.08) -- (243.63,43.08) -- cycle ;

\draw  [line width=0.75]  (243.16,46.52) -- (264.01,46.52) -- (264.01,67.39) -- (243.16,67.39) -- cycle ;

\draw  [color={rgb, 255:red, 255; green, 255; blue, 255 }  ,draw opacity=1 ][fill={rgb, 255:red, 74; green, 74; blue, 74 }  ,fill opacity=1 ] (243.16,97.79) -- (264.01,97.79) -- (264.01,118.67) -- (243.16,118.67) -- cycle ;

\draw (243.08,26.82) node [anchor=north west][inner sep=0.75pt]  [font=\scriptsize,color={rgb, 255:red, 0; green, 0; blue, 0 }  ,opacity=1 ]  {$\langle x\rangle _{i}$};
\draw (245.16,49.92) node [anchor=north west][inner sep=0.75pt]  [font=\small]  {$x\rangle $};
\draw (29.28,104.05) node [anchor=north west][inner sep=0.75pt]  [font=\footnotesize] [align=left] {Neigh. Embedding};
\draw (64.28,81.8) node  [font=\footnotesize] [align=left] {Layer stacks};
\draw (76.28,56.79) node  [font=\footnotesize] [align=left] {Conv. connection};
\draw (78.28,32.5) node  [font=\footnotesize] [align=left] {Dense connection};
\draw (157.59,104.59) node [anchor=north west][inner sep=0.75pt]  [font=\footnotesize] [align=left] {Softmax Layer};
\draw (187.09,82.96) node  [font=\footnotesize] [align=left] {ELU Layer};
\draw (190.59,56.79) node  [font=\footnotesize] [align=left] {ReLU Layer};
\draw (192.09,32.5) node  [font=\footnotesize] [align=left] {Linear Layer};
\draw (304.38,110.73) node  [font=\footnotesize] [align=left] {Output Layer};
\draw (299.38,83.47) node  [font=\footnotesize] [align=left] {Input Layer};
\draw (268.38,49.52) node [anchor=north west][inner sep=0.75pt]  [font=\footnotesize] [align=left] {Inner Product with};
\draw (268.38,27.21) node [anchor=north west][inner sep=0.75pt]  [font=\footnotesize] [align=left] {Axis Average};

\end{tikzpicture}

            }
        }\\
    \caption{Neural network architecture for controlling Kuramoto oscillators and symbol legend.}
\end{figure}
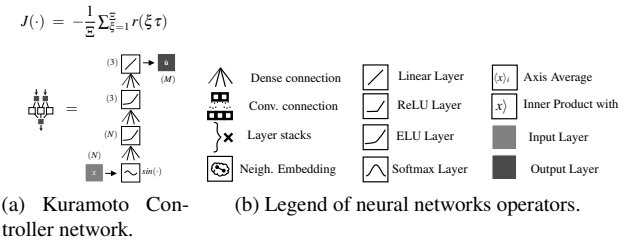

Only the current system state $\tv{\statevar}(t)$ is provided as an input for the neural network, similar to the baseline described in \Cref{sec:osc:baseline}.
We use a fully connected architecture as illustrated in \Cref{fig:osci:arch}.
Finally, to calculate the binary driver matrix for the neural network $B$ for in \Cref{eq:kuramoto} we assign $\matr{B}_{i,m}=1$ for the $m\textrm{-th}$ driver node if $b^{(\textrm{FC})}_{i} \neq 0$, as we require the network to learn the control signals per driver node without prior knowledge of the exact control gains, but only the available driver nodes.

Since one of our control goals is to stabilize Kuramoto oscillators in a synchronized state,
we need to adapt the training scheme presented in \Cref{algo:NODEC}.
The loss of synchronization may occur at any point of the trajectory, we train NODEC (see Supplemental Material \Cref{algo:training:curiculum}) in a curriculum learning procedure~\cite{bengio2009curriculum}, 
where NODEC is initially trained on trajectories sampled for low values of $T$.
The value of $T$ increases gradually as training proceeds.
The learning process in the beginning of the curriculum, when $T$ is very low, 
allows NODEC to learn controls that steer the oscillators through the transient state between synchronicity and no-synchronicity.
As $T$ increases the network also learns controls that preserve the network in the synchronized state.

In feedback control, the target is often to synchronize the system for different initial states~\cite{aastrom2010feedback}.
To train the system for more than one initial state, we use a mini-batch-training procedure that samples $8$ random initial states per epoch for training.
We observed that randomly sampling an initial state from a uniform distribution in $[0, 2\pi]$ does not improve training performance and fails to learn synchronization. 
It has been reported in the literature~\cite{ioffe2015batch} that normally-distributed layer inputs (with zero mean  and unit variance) can help neural networks converge faster. Therefore, we decided to sample initial states from a normal distribution with zero mean and unit variance.
Our results confirm that learning and convergence improve.
Sampling initial states enables us to use mini-batches to speed up and stabilize training as well.
In the Kuramoto example we use the Adam optimizer~\cite{kingma2014adam} for parameter optimization.
The complete training scheme is also illustrated in the Supplemental Material~\Cref{algo:training:curiculum}.

\subsubsection{Learning Loss Function}
For synchronization of Kuramoto oscillators, we consider the order parameter~\cite{brede2008locals}:
\begin{equation}
\label{eq:kuramoto:order}
    r(t) = \dfrac{1}{N}\sqrt{\sum_{i,j}\cos \left[\statevar_i(t)-\statevar_j(t)\right]} = \dfrac{1}{N}\sqrt{\sum_{i,j} e^{\iu (\statevar_i-\statevar_j)}}
\end{equation}
to determine the degree of synchronicity. 
The control loss may also aggregate the order parameter over time, when the control goals take stability into account.
In such a case, one might consider the mean order parameter over time
\begin{equation}
\label{eq:mean:parameter:cont}
 \overline{r}(t) = \frac{1}{T} \int_0^T r(t) \, \mathrm{d}t,
\end{equation}
which approaches zero if the oscillators are incoherent. 
By discretizing $T$ into $\Xi$ intervals, we can also discretize \Cref{eq:mean:parameter:cont} using
\begin{equation}
    \label{eq:mean:parameter:discrete}
    \overline{r}(t) = \dfrac{1}{\Xi}\sum ^{\Xi}_{\xi=0} r(\xi\tau)\tau, \quad \Xi \tau = T.
\end{equation}
and for the numerical calculations we omit $\tau$.
Equation \eqref{eq:mean:parameter:discrete} can be used as a loss function
\begin{equation}
    \label{eq:kuramoto:mean:loss}
    J(\ts{\statevar}_\tau^T) = - \overline{r}(t) = - \dfrac{1}{\Xi}\sum ^{\Xi}_{\xi=1} r(\xi\tau)
\end{equation}
to achieve stable synchronization of coupled oscillators.
Such a loss introduces two challenges with respect to the classical MSE loss~\cite{bemporad2002explicit}: 
(i) it is a macroscopic loss as we do not require to reach a specific state vector $\tv{\statevar}^*$ to minimize\footnote{
\tomcom{The target states that satisfy this control goal are not unique and not necessarily known, but satisfy $\tv{\statevar}^* = \argmax_{\tv{\statevar}}r(\tv{\statevar})$.
Since there is no specific dependence to a target state vector, we omit the term $\target{\tv{\statevar}}$ from the loss function.}
} \Cref{eq:kuramoto:mean:loss} and 
(ii) the loss is calculated over a time interval $[\tau, T]$.\footnote{
The initial time is omitted ($\xi=\{1,\dots,Xi\}$ in \Cref{eq:kuramoto:mean:loss}), since we assume that no control is applied prior to reaching the initial state.
}
In our numerical experiments we observed that using such a loss affects numerical stability, 
especially for long time intervals, e.g.~when $\Xi=100$ timesteps.
Averaging over $r(\xi t)$ in \Cref{fig:osci:loss:all} may smooth out temporal drops of $r(t)$, especially for very high values of $\Xi$.
When such drops occur in sampled training trajectories, NODEC learns to achieve high synchronicity only temporarily.
NODEC learns controls that yield highly synchronized stable trajectories similar to FC,
when we extend \Cref{eq:kuramoto:mean:loss} by subtracting the minimum order parameter value $\min_{t\in[\tau,T]} {r(t)}$ over time:
\begin{equation}
\label{eq:kuramoto:loss}
    J\left(\ts{\statevar}_{\tau}^T\right)= -\left[\overline{r}(t) + \min_{t\in[\tau,T]} {r(t)}\right].
\end{equation}
Introducing the minimum order parameter term increases the stability of the learned control,
as the loss creates higher gradients for controls that cause loss of synchronization.
NODEC is trained on trajectories that may at maximum reach total time of $T=40$, 
but is evaluated on trajectories of $T=150$.

\subsubsection{Results}\label{sec:osc:results}
\begin{figure*}
    \centering
    \subfloat[Control energy over time.]{
    \includegraphics[width=0.54\columnwidth,valign=b]{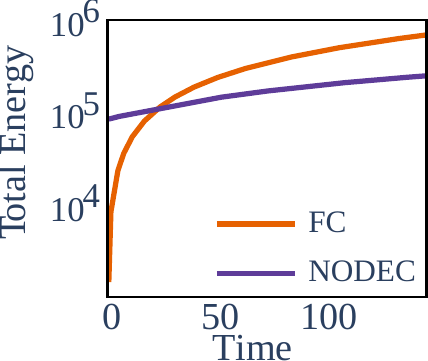}
    \label{fig:osci:energy}
    }
    \hfill
    \subfloat[Order parameter over time.]{
    \includegraphics[width=0.50\columnwidth, valign=b]{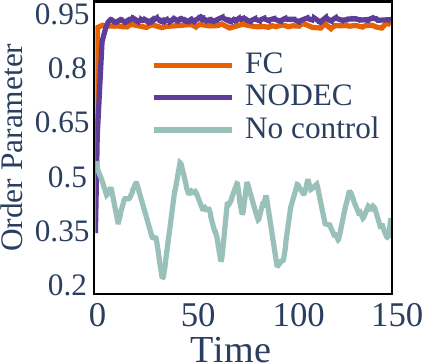}
    \label{fig:osci:loss:all}
    }\hfill
    \subfloat[Relative NODEC performance against FC.]{
    \includegraphics[width=0.64\columnwidth,valign=b]{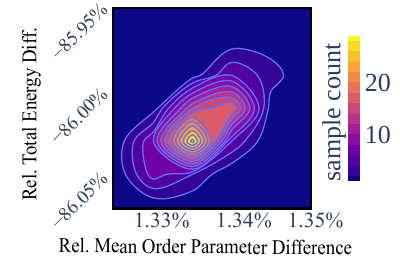}
    \label{fig:kuramoto:resutls:many}
    }
    \caption{Comparison of NODEC and feedback control in terms of energy and synchronization stability.}
\end{figure*}
To test the control performance of NODEC, we first sample an unobserved initial state close to the synchronized steady state in accordance with~\cite{skardal2015control}.
The initial state values for single sample evaluation (see \Cref{fig:osci:energy,fig:osci:loss:all}) are uniformly sampled within $-10\%$ of the synchronized steady state values, 
i.e.~$\statevar_i \in [0.9 \statevar_i^{\diamond}, \statevar_i^{\diamond}]$, in order to be close to the synchronized steady state as proposed in Ref.~\cite{skardal2015control}.
We observe that the neural network achieves a target state with larger order parameter values (see \Cref{fig:osci:loss:all}) and requires lower energy (see \Cref{fig:osci:energy}) than the FC baseline.
We also observe that NODEC requires higher energy and slightly more time to synchronize the system but less to preserve it, compared to the FC baseline (see \Cref{fig:osci:loss:all} and Supplemental Material~\Cref{appdx:fig:osci:loss:zoom}).

To determine whether NODEC can achieve synchronization stability regardless of the initial state choice (see \Cref{fig:kuramoto:resutls:many}) and its proximity to the synchronized steady state, we test the trained model on $100$ initial states, with values uniformly sampled in $[0, 1]$.
In \Cref{fig:kuramoto:resutls:many} the vertical axis represents the relative total energy difference between NODEC and FC for the same initialization $\left( E_{\rm NODEC}(T) - E_{\rm FC}(T) \right)/E_{\rm FC}(T)$.
The horizontal axis represents the mean relative order parameter difference calculated as $\left( r_{\rm NODEC}(T) - r_{\rm FC}(T) \right)/r_{\rm FC}(T)$.
NODEC achieves around $1\%$ higher order parameter values and almost $86\%$ less total control energy for all samples.
More sophisticated strategies of adapting the constant term $\zeta$ in \Cref{eq:kuramoto:driver:selection} could be applied to adapt the driver matrix values in feedback control. 
This is, however, out of scope of this paper.
Our results show that NODEC can be adapted to achieve highly synchronized states in Kuramoto dynamics on an Erd\H{o}s--R\'enyi graph via feedback control.

\subsection{Epidemic Spreading and Targeted Interventions}\label{sec:spread}
Designing targeted intervention and immunization strategies~\cite{bottcher2017targeted,salathe2010dynamics} is important to contain the spread of epidemics.
\tomcom{
To study the performance of NODEC in such containment tasks, we will use the SIR-type model~\cite{maier2020effective} that extends the SIR model by accounting for quarantine interventions and other preventive or reactive measures for disease containment. 
In our formulation of SIR-type dynamics, we also account for control inputs and network structure. The ``R'' compartment in our model is used to describe (i) recovered individuals that were infected and acquired immunity and (ii) removed individuals (i.e., susceptible individuals under quarantine who do not interact with anyone else).
In this case, the complete state of the dynamics is now a matrix $\matr{x}(t) \in \mathbb{R}^{4\times N}$, where each row represents a state vector of the SIR-type dynamics\footnote{We note that here we use capital letters for the SIR-type variables, to follow the common notation in related literature.}. 
The states of node represent the fraction of $\matr{\statevar}_{1,i} = S_i$, infected $\matr{\statevar}_{2,i} = I_i$, recovered $\matr{\statevar}_{3,i} = R_i$, and quarantined $\matr{\statevar}_{4,i} = Y_i$ individuals in the node.
}
The corresponding generalized SIR-type dynamics of node $i$ is described by a set of rate equations:

\tomcom{
\begin{subequations}
\footnotesize
\begin{align}
\begin{split}
    \label{eq:spread:susc}
    \dot{S}_i(t) &= -\beta S_i(t) \sum_j \matr{A}_{i,j}I_j(t) - \sum_m \matr{B}_{i,m} u_{m}(\tv{\statevar}(t)) S_i(t)
\end{split}\\
\begin{split}
    \label{eq:spread:inf}
    \dot{I}_i(t) &= \beta S_i(t) \sum_j \matr{A}_{i,j}I_j(t) - \gamma I_i(t) - \sum_m \matr{B}_{i,m} u_{m}(\tv{\statevar}(t)) I_i(t)
 \end{split}\\
 \begin{split}
    \label{eq:spread:rem}
    \dot{R}_i(t) &= \gamma I_i(t) + \sum_m \matr{B}_{i,m} u_{m}(\tv{\statevar}(t)) S_i(t)
\end{split}\\
\begin{split}
    \label{eq:spread:cont}
    \dot{Y}_i(t) &= \sum_m \matr{B}_{i,m} u_{m}(\tv{\statevar}(t)) I_i(t)
\end{split}
\end{align}
\label{eq:SIRX}
\end{subequations}
}

\noindent subject to the conditions that (i) the the total population is conserved and (ii)
the control budget is set as $\mathcal{b}$: 
\begin{subequations}
\begin{align}
\begin{split}
\label{eq:sirx:conservation}
\sum_i \left[S_i + I_i + R_i + Y_i \right] &= N,
\end{split}\\
\begin{split}
\label{eq:sirx:budget}
\sum_{m,i} \matr{B}_{i,m} u_{m}(\tv{\statevar}(t)) &\leq \mathcal{b}.
\end{split}
\end{align}
\end{subequations}
\tomcom{
The driver nodes $B_{i,m} = 1$ can be selected via different methods, e.g. the nodes/communities that are willing to apply proactive and reactive measures.
For the specific example, driver nodes are selected with the maximum matching method~\cite{Commault2002, yamada1990graph}.
}
In our simulations, we assume that the epidemic originates from a localized part in the graph and we minimize the proposed epidemic loss in \Cref{eq:spread:loss} for a different part of the graph.
The parameters $\beta$ and $\gamma$ are the infection and recovery rates, and $u_{m}(\tv{\statevar}(t))$ describes the effect of containment interventions (e.g., quarantine, mask-usage and distancing). 
When a neural network controller (NODEC or RL) is used, we set $u_{m}(\tv{\statevar}(t)) = \nnc{u}_m(\tv{\statevar}(t))$,
We observe that control terms $\sum_m \matr{B}_{i,m} u_{m}(\tv{\statevar}(t))$ cancel out when summing over the pairs of \Cref{eq:spread:susc,eq:spread:inf} and \Cref{eq:spread:cont,eq:spread:rem}.
These terms are used to model preventive and reactive measures, respectively. 
For example, susceptible individuals may isolate themselves and completely avoid infection ($S \to R$) until the pandemic passes (preventive) or infected individuals are quarantined and put to intensive care to avoid spreading and recover ($I\to Y$) (reactive measure).

\subsubsection{Control Baselines}\label{:baselines}
A baseline that takes structural node properties (e.g., node degree or centrality) into the account,
may be a good baseline for structural-heterogeneous graphs, but not for regular structures like lattices.
Clearly, a weak baseline (RND) would be assigning random control inputs to driver nodes with $u_{m}(t) = \mathcal{b}c_m/\sum_{m'=0}^{M}c_{m'}, c_m\sim \mathcal{U}(0,1)$.
However, a targeted constant control baseline (TCC), 
which in the presence of an ``oracle'' assigns constant control inputs $u_{m}(t) = \mathcal{b}/M$ to every driver node in $\target{G}$, 
is a strong baseline for constant control.
The budget constraint [see \Cref{eq:sirx:budget}], 
the high number of nodes connecting $\target{G}$ to the rest of lattice graph and constraint to only control driver nodes does not allow to create dense "impenetrable walls" of containment,
as an infection can still pass through contained nodes at a lower rate.
As TCC is a static control, it already protects the driver nodes from $t=0$ on, so TCC-controlled nodes will be infected very slowly. 
Assigning all budget to all driver nodes of interest also minimizes wasted ``containment'' budget.
Still, distributing more budget to a smaller number of nodes increases the L2 norm of the control, making controls very expensive when considering quadratic energy costs.
To have a control with less energy, it is important to distribute the budget to more nodes, therefore enabling more global containment and less constant containment on the target sub-graph.

We also study the performance of neural dynamic control baselines, such as continuous-action RL, with fully-connected neural networks or our variant (see \Cref{fig:nn:gnn}) as policy architectures, which we discuss further in the \Cref{sec:spread:nodec,appdx:sirx:nn}.
Only one of the three evaluated training routines of RL provided high-performance results.
We tested: SAC~\cite{haarnoja2018soft}, TD3~\cite{fujimoto2018addressing}, and A2C~\cite{mnih2016asynchronous}, but we report only the results of TD3 which were more competitive with respect to NODEC.
To allow RL to tackle the SIR-type control problem, we first implement SIR-type dynamics as an RL environment.
The input of the RL is the tensor of all SIR-type states at time $t$. 
We consider an observation space, which includes continuous values in $[0,1]$ and has dimension $4 \times N$.
RL actions $a_m(t) \in \mathbb{R}$ are continuous values for each driver node and correspond to control signals.
Once the actions are passed to the environment, a pre-processing operation takes place to convert the RL action into valid control signals (see decision network of \Cref{fig:nn:gnn}).
Reinforcement learning is allowed to provide change the control signals to (interact with) the environment in a fixed discrete time interaction interval $\Delta t = 10^{-2}$ during training. 
Lower interaction intervals were also considered, but required longer training and did not seem to improve performance.
For reinforcement learning, we need to express the control goal as a reward function which is used for the approximation of action value function within the RL framework.
We tested several reward designs and we describe this process in the Supplemental Material \Cref{sec:appendix:sirx:rl}, but we observed best performance with the following reward function:
\begin{equation}\label{eq:reward:sum:diff}
    \rho(t)= 
\begin{cases}
    \quad 0 , \quad \text{if } \bar{I}_{\target{G}}(t) \leq \max_{\uptau < t} (\bar{I}_{\target{G}}(\tau) )\\
    -\bar{I}_{\target{G}}^{2}(t) +(\max_{\tau < t} \bar{I}_{\target{G}}(\tau))^{2} , \quad \text{otherwise}
\end{cases}.
\end{equation}
\subsubsection{Numerical Simulation}
\label{sec:spread:eval}
To determine the target time $T$, we observe the SIR-type dynamics ($\beta=6$ and $\gamma=1.8$) on a $32 \times 32$ lattice without control and set its value to the time at which the mean infection over all nodes is approximately zero. Initially, the epidemic starts from a deterministic selection of nodes in the upper-right quadrant.
For all control strategies, the budget (maximal number of control interventions) is $\mathcal{b}=600$.
Given that Reinforcement Learning takes considerably longer to converge and that we were required to perform a much more extensive hyper parameter search, we showcase our experiments only on the lattice graph and a single initial state.
Our control goal is to contain epidemic outbreaks (i.e., ``flattening'' the infection curve) in the sub-graph $\target{G}$,
which is located in the bottom-left quadrant (see \Cref{fig:sirx}). 
All baselines are compared under interaction interval of $\Delta t = 10^{-3}$.
\subsubsection{NODEC Hyperparameters}
\label{sec:spread:nodec}
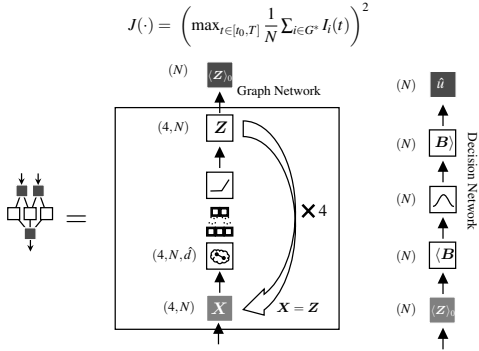
\begin{figure}
    \centering
   \resizebox{0.8\columnwidth}{!}{

\tikzset{every picture/.style={line width=0.75pt}} 
\begin{tikzpicture}[x=0.75pt,y=0.75pt,yscale=-1,xscale=1]

\draw    (412.83,222.75) -- (412.83,208.5) ;
\draw [shift={(412.83,205.5)}, rotate = 450] [fill={rgb, 255:red, 0; green, 0; blue, 0 }  ][line width=0.08]  [draw opacity=0] (8.93,-4.29) -- (0,0) -- (8.93,4.29) -- cycle    ;
\draw  [line width=0.75]  (402.41,137.13) -- (423.26,137.13) -- (423.26,158) -- (402.41,158) -- cycle ;
\draw    (404.86,152.86) .. controls (408.51,153.39) and (409.9,143.9) .. (413.19,143.75) .. controls (416.49,143.6) and (418.66,154.17) .. (421.53,152.86) ;

\draw  [color={rgb, 255:red, 255; green, 255; blue, 255 }  ,draw opacity=1 ][fill={rgb, 255:red, 74; green, 74; blue, 74 }  ,fill opacity=1 ] (402.41,45.56) -- (423.26,45.56) -- (423.26,66.44) -- (402.41,66.44) -- cycle ;

\draw    (412.83,133.81) -- (412.83,119.56) ;
\draw [shift={(412.83,116.56)}, rotate = 450] [fill={rgb, 255:red, 0; green, 0; blue, 0 }  ][line width=0.08]  [draw opacity=0] (8.93,-4.29) -- (0,0) -- (8.93,4.29) -- cycle    ;
\draw    (412.83,89.08) -- (412.83,74.83) ;
\draw [shift={(412.83,71.83)}, rotate = 450] [fill={rgb, 255:red, 0; green, 0; blue, 0 }  ][line width=0.08]  [draw opacity=0] (8.93,-4.29) -- (0,0) -- (8.93,4.29) -- cycle    ;
\draw  [line width=0.75]  (402.05,181.46) -- (422.89,181.46) -- (422.89,202.33) -- (402.05,202.33) -- cycle ;

\draw    (412.83,178.75) -- (412.83,164.5) ;
\draw [shift={(412.83,161.5)}, rotate = 450] [fill={rgb, 255:red, 0; green, 0; blue, 0 }  ][line width=0.08]  [draw opacity=0] (8.93,-4.29) -- (0,0) -- (8.93,4.29) -- cycle    ;
\draw  [line width=0.75]  (402.05,92.13) -- (422.89,92.13) -- (422.89,113) -- (402.05,113) -- cycle ;

\draw  [color={rgb, 255:red, 255; green, 255; blue, 255 }  ,draw opacity=1 ][fill={rgb, 255:red, 128; green, 128; blue, 128 }  ,fill opacity=1 ] (401.78,225.06) -- (422.62,225.06) -- (422.62,245.94) -- (401.78,245.94) -- cycle ;

\draw    (412.83,266.75) -- (412.83,252.5) ;
\draw [shift={(412.83,249.5)}, rotate = 450] [fill={rgb, 255:red, 0; green, 0; blue, 0 }  ][line width=0.08]  [draw opacity=0] (8.93,-4.29) -- (0,0) -- (8.93,4.29) -- cycle    ;
\draw    (84.55,158.78) -- (84.55,174.78) ;
\draw    (78.02,140.65) -- (84.55,158.78) ;
\draw    (91.08,140.65) -- (84.55,158.78) ;
\draw    (91.08,140.65) -- (97.62,158.78) ;
\draw    (97.62,158.78) -- (84.55,174.78) ;
\draw    (71.49,158.78) -- (84.55,174.78) ;
\draw    (78.02,140.65) -- (71.49,158.78) ;
\draw  [fill={rgb, 255:red, 255; green, 255; blue, 255 }  ,fill opacity=1 ] (92.88,154.04) -- (102.35,154.04) -- (102.35,163.52) -- (92.88,163.52) -- cycle ;
\draw  [fill={rgb, 255:red, 255; green, 255; blue, 255 }  ,fill opacity=1 ] (66.75,154.04) -- (76.22,154.04) -- (76.22,163.52) -- (66.75,163.52) -- cycle ;
\draw  [fill={rgb, 255:red, 255; green, 255; blue, 255 }  ,fill opacity=1 ] (79.82,154.04) -- (89.29,154.04) -- (89.29,163.52) -- (79.82,163.52) -- cycle ;

\draw  [color={rgb, 255:red, 255; green, 255; blue, 255 }  ,draw opacity=1 ][fill={rgb, 255:red, 74; green, 74; blue, 74 }  ,fill opacity=1 ] (86.35,135.91) -- (95.82,135.91) -- (95.82,145.39) -- (86.35,145.39) -- cycle ;
\draw  [color={rgb, 255:red, 255; green, 255; blue, 255 }  ,draw opacity=1 ][fill={rgb, 255:red, 74; green, 74; blue, 74 }  ,fill opacity=1 ] (73.28,135.91) -- (82.76,135.91) -- (82.76,145.39) -- (73.28,145.39) -- cycle ;

\draw  [color={rgb, 255:red, 255; green, 255; blue, 255 }  ,draw opacity=1 ][fill={rgb, 255:red, 74; green, 74; blue, 74 }  ,fill opacity=1 ] (79.82,170.04) -- (89.29,170.04) -- (89.29,179.51) -- (79.82,179.51) -- cycle ;
\draw    (77.74,126.65) -- (77.74,132.91) ;
\draw [shift={(77.74,135.91)}, rotate = 270] [fill={rgb, 255:red, 0; green, 0; blue, 0 }  ][line width=0.08]  [draw opacity=0] (5.36,-2.57) -- (0,0) -- (5.36,2.57) -- (3.56,0) -- cycle    ;
\draw    (90.93,126.8) -- (90.93,133.06) ;
\draw [shift={(90.93,136.06)}, rotate = 270] [fill={rgb, 255:red, 0; green, 0; blue, 0 }  ][line width=0.08]  [draw opacity=0] (5.36,-2.57) -- (0,0) -- (5.36,2.57) -- (3.56,0) -- cycle    ;
\draw    (84.78,179.74) -- (84.78,186) ;
\draw [shift={(84.78,189)}, rotate = 270] [fill={rgb, 255:red, 0; green, 0; blue, 0 }  ][line width=0.08]  [draw opacity=0] (5.36,-2.57) -- (0,0) -- (5.36,2.57) -- (3.56,0) -- cycle    ;

\draw  [color={rgb, 255:red, 255; green, 255; blue, 255 }  ,draw opacity=1 ][fill={rgb, 255:red, 128; green, 128; blue, 128 }  ,fill opacity=1 ] (224.88,223.65) -- (245.73,223.65) -- (245.73,244.53) -- (224.88,244.53) -- cycle ;

\draw    (235.31,221.75) -- (235.31,207.5) ;
\draw [shift={(235.31,204.5)}, rotate = 450] [fill={rgb, 255:red, 0; green, 0; blue, 0 }  ][line width=0.08]  [draw opacity=0] (8.93,-4.29) -- (0,0) -- (8.93,4.29) -- cycle    ;
\draw  [line width=0.75]  (224.88,182.06) -- (245.73,182.06) -- (245.73,202.94) -- (224.88,202.94) -- cycle ;
\draw   (241.5,191.22) .. controls (241.62,190.33) and (241.23,189.46) .. (240.51,188.96) .. controls (239.78,188.46) and (238.83,188.44) .. (238.08,188.89) .. controls (237.81,188.37) and (237.32,188.02) .. (236.76,187.93) .. controls (236.19,187.84) and (235.62,188.03) .. (235.21,188.44) .. controls (234.98,187.97) and (234.54,187.66) .. (234.03,187.61) .. controls (233.52,187.56) and (233.02,187.78) .. (232.71,188.2) .. controls (232.3,187.7) and (231.64,187.49) .. (231.03,187.66) .. controls (230.41,187.83) and (229.95,188.34) .. (229.83,188.98) .. controls (229.33,189.12) and (228.91,189.48) .. (228.68,189.96) .. controls (228.45,190.45) and (228.44,191.01) .. (228.65,191.5) .. controls (228.15,192.16) and (228.03,193.04) .. (228.34,193.81) .. controls (228.65,194.59) and (229.34,195.13) .. (230.14,195.25) .. controls (230.15,195.98) and (230.54,196.64) .. (231.16,196.99) .. controls (231.78,197.34) and (232.54,197.32) .. (233.14,196.93) .. controls (233.4,197.8) and (234.12,198.44) .. (235,198.57) .. controls (235.87,198.71) and (236.75,198.31) .. (237.24,197.56) .. controls (237.84,197.93) and (238.57,198.04) .. (239.25,197.85) .. controls (239.93,197.67) and (240.51,197.22) .. (240.86,196.59) .. controls (241.47,196.67) and (242.07,196.34) .. (242.35,195.78) .. controls (242.63,195.21) and (242.53,194.53) .. (242.11,194.07) .. controls (242.66,193.74) and (242.94,193.08) .. (242.81,192.44) .. controls (242.67,191.8) and (242.15,191.32) .. (241.52,191.26) ; \draw   (242.11,194.07) .. controls (241.85,194.22) and (241.55,194.29) .. (241.25,194.27)(240.86,196.59) .. controls (240.73,196.58) and (240.6,196.54) .. (240.48,196.5)(237.24,197.56) .. controls (237.33,197.42) and (237.41,197.27) .. (237.46,197.11)(233.14,196.93) .. controls (233.1,196.78) and (233.07,196.61) .. (233.05,196.45)(230.14,195.25) .. controls (230.14,194.48) and (230.57,193.77) .. (231.25,193.44)(228.65,191.5) .. controls (228.76,191.76) and (228.93,191.99) .. (229.14,192.18)(229.83,188.98) .. controls (229.82,189.09) and (229.81,189.2) .. (229.81,189.31)(232.71,188.2) .. controls (232.81,188.32) and (232.9,188.46) .. (232.96,188.61)(235.21,188.44) .. controls (235.27,188.55) and (235.31,188.67) .. (235.33,188.79)(238.08,188.89) .. controls (237.92,188.98) and (237.77,189.1) .. (237.64,189.23)(241.5,191.22) .. controls (241.49,191.34) and (241.46,191.46) .. (241.43,191.58) ;
\draw  [fill={rgb, 255:red, 0; green, 0; blue, 0 }  ,fill opacity=1 ] (239.61,194.1) .. controls (239.61,194.71) and (239.11,195.2) .. (238.51,195.2) .. controls (237.9,195.2) and (237.41,194.71) .. (237.41,194.1) .. controls (237.41,193.49) and (237.9,193) .. (238.51,193) .. controls (239.11,193) and (239.61,193.49) .. (239.61,194.1) -- cycle ;
\draw  [fill={rgb, 255:red, 0; green, 0; blue, 0 }  ,fill opacity=1 ] (234.01,191.1) .. controls (234.01,191.71) and (233.51,192.2) .. (232.91,192.2) .. controls (232.3,192.2) and (231.81,191.71) .. (231.81,191.1) .. controls (231.81,190.49) and (232.3,190) .. (232.91,190) .. controls (233.51,190) and (234.01,190.49) .. (234.01,191.1) -- cycle ;
\draw    (232.91,191.1) -- (238.51,194.1) ;

\draw  [fill={rgb, 255:red, 155; green, 155; blue, 155 }  ,fill opacity=1 ] (224.9,169.61) -- (231.84,169.61) -- (231.84,176.56) -- (224.9,176.56) -- cycle(230.8,170.65) -- (225.94,170.65) -- (225.94,175.52) -- (230.8,175.52) -- cycle ;
\draw  [fill={rgb, 255:red, 155; green, 155; blue, 155 }  ,fill opacity=1 ] (231.84,169.61) -- (238.77,169.61) -- (238.77,176.56) -- (231.84,176.56) -- cycle(237.73,170.65) -- (232.88,170.65) -- (232.88,175.52) -- (237.73,175.52) -- cycle ;
\draw  [fill={rgb, 255:red, 155; green, 155; blue, 155 }  ,fill opacity=1 ] (238.77,169.61) -- (245.71,169.61) -- (245.71,176.56) -- (238.77,176.56) -- cycle(244.67,170.65) -- (239.81,170.65) -- (239.81,175.52) -- (244.67,175.52) -- cycle ;

\draw [line width=0.75]  [dash pattern={on 0.84pt off 2.51pt}]  (230.95,162.11) -- (228.58,169.19) ;
\draw  [dash pattern={on 0.84pt off 2.51pt}]  (239.66,162.11) -- (242.03,169.19) ;

\draw [line width=0.75]  [dash pattern={on 0.84pt off 2.51pt}]  (233.35,161.31) -- (230.98,168.39) ;
\draw  [dash pattern={on 0.84pt off 2.51pt}]  (242.06,161.31) -- (244.43,168.39) ;

\draw [line width=0.75]  [dash pattern={on 0.84pt off 2.51pt}]  (228.55,161.31) -- (226.18,168.39) ;
\draw  [dash pattern={on 0.84pt off 2.51pt}]  (237.26,161.31) -- (239.63,168.39) ;

\draw  [fill={rgb, 255:red, 155; green, 155; blue, 155 }  ,fill opacity=1 ] (228.37,153.91) -- (235.31,153.91) -- (235.31,160.85) -- (228.37,160.85) -- cycle(234.27,154.95) -- (229.41,154.95) -- (229.41,159.81) -- (234.27,159.81) -- cycle ;
\draw  [fill={rgb, 255:red, 155; green, 155; blue, 155 }  ,fill opacity=1 ] (235.31,153.91) -- (242.24,153.91) -- (242.24,160.85) -- (235.31,160.85) -- cycle(241.2,154.95) -- (236.35,154.95) -- (236.35,159.81) -- (241.2,159.81) -- cycle ;

\draw [line width=0.75]    (242.71,131.69) -- (236.92,140.54) -- (227.9,140.77) ;
\draw  [line width=0.75]  (224.88,125.79) -- (245.73,125.79) -- (245.73,146.67) -- (224.88,146.67) -- cycle ;

\draw    (235.31,122.48) -- (235.31,108.23) ;
\draw [shift={(235.31,105.23)}, rotate = 450] [fill={rgb, 255:red, 0; green, 0; blue, 0 }  ][line width=0.08]  [draw opacity=0] (8.93,-4.29) -- (0,0) -- (8.93,4.29) -- cycle    ;
\draw  [fill={rgb, 255:red, 255; green, 255; blue, 255 }  ,fill opacity=1 ] (295.5,169.4) .. controls (295.5,129.66) and (276.92,97.45) .. (254,97.45) -- (254,85) .. controls (276.92,85) and (295.5,117.21) .. (295.5,156.95) ;\draw  [fill={rgb, 255:red, 255; green, 255; blue, 255 }  ,fill opacity=1 ] (295.5,156.95) .. controls (295.5,186.45) and (285.26,211.81) .. (270.6,222.91) -- (270.6,218.76) -- (254,235.13) -- (270.6,239.51) -- (270.6,235.36) .. controls (285.26,224.26) and (295.5,198.9) .. (295.5,169.4)(295.5,156.95) -- (295.5,169.4) ;
\draw  [line width=1.5]  (310.49,151.69) -- (305.76,156.61) -- (301.05,151.65) ;
\draw  [line width=1.5]  (301.02,161.53) -- (305.75,156.61) -- (310.46,161.57) ;

\draw   (152.2,74.4) -- (330.2,74.4) -- (330.2,250.81) -- (152.2,250.81) -- cycle ;
\draw    (234.31,262.75) -- (234.31,248.5) ;
\draw [shift={(234.31,245.5)}, rotate = 450] [fill={rgb, 255:red, 0; green, 0; blue, 0 }  ][line width=0.08]  [draw opacity=0] (8.93,-4.29) -- (0,0) -- (8.93,4.29) -- cycle    ;
\draw    (235.31,78.75) -- (235.31,64.5) ;
\draw [shift={(235.31,61.5)}, rotate = 450] [fill={rgb, 255:red, 0; green, 0; blue, 0 }  ][line width=0.08]  [draw opacity=0] (8.93,-4.29) -- (0,0) -- (8.93,4.29) -- cycle    ;
\draw  [color={rgb, 255:red, 255; green, 255; blue, 255 }  ,draw opacity=1 ][fill={rgb, 255:red, 74; green, 74; blue, 74 }  ,fill opacity=1 ] (224.97,37.99) -- (245.81,37.99) -- (245.81,58.87) -- (224.97,58.87) -- cycle ;

\draw  [line width=0.75]  (224.88,80.25) -- (245.73,80.25) -- (245.73,101.12) -- (224.88,101.12) -- cycle ;

\draw (383.51,235.56) node  [font=\footnotesize] [align=left] {$ $$\displaystyle ( N)$};
\draw (383.65,57.16) node  [font=\footnotesize] [align=left] {$ $$\displaystyle ( N)$};
\draw (383.65,103.83) node  [font=\footnotesize] [align=left] {$ $$\displaystyle ( N)$};
\draw (383.65,147.83) node  [font=\footnotesize] [align=left] {$ $$\displaystyle ( N)$};
\draw (383.65,193.16) node  [font=\footnotesize] [align=left] {$ $$\displaystyle ( N)$};
\draw (441.69,92.53) node [anchor=north west][inner sep=0.75pt]  [font=\footnotesize,rotate=-90.26] [align=left] {Decision Network};
\draw (401.59,230.46) node [anchor=north west][inner sep=0.75pt]  [font=\scriptsize,color={rgb, 255:red, 255; green, 255; blue, 255 }  ,opacity=1 ]  {$\langle \matr{z}\rangle _{0}$};
\draw (401.83,95.23) node [anchor=north west][inner sep=0.75pt]  [font=\small]  {$\ \matr{B} \rangle $};
\draw (401.83,184.57) node [anchor=north west][inner sep=0.75pt]  [font=\small]  {$\ \langle \matr{B}$};
\draw (406.33,48.9) node [anchor=north west][inner sep=0.75pt]  [font=\small,color={rgb, 255:red, 255; green, 255; blue, 255 }  ,opacity=1 ]  {$\hat{u}$};
\draw (111,158.4) node [anchor=north west][inner sep=0.75pt]  [font=\LARGE]  {$=$};
\draw (202.31,45.3) node  [font=\footnotesize] [align=left] {$ $$\displaystyle ( N)$};
\draw (202.98,233.5) node  [font=\footnotesize] [align=left] {$ $$\displaystyle ( 4,N)$};
\draw (199.38,191.5) node  [font=\footnotesize] [align=left] {$ $$\displaystyle ( 4,N,\hat{d})$};
\draw (198.48,90.5) node  [font=\footnotesize] [align=left] {$ $$\displaystyle ( 4,N)$};
\draw (279,228.4) node [anchor=north west][inner sep=0.75pt]  [font=\footnotesize]  {$\matr{\statevar}=\matr{z}$};
\draw (317.38,156) node  [font=\normalsize] [align=left] {$ $$\displaystyle 4$};
\draw (247.8,55.6) node [anchor=north west][inner sep=0.75pt]  [font=\footnotesize] [align=left] {Graph Network};
\draw (236.31,89.68) node    {$\matr{Z}$};
\draw (224.39,43.33) node [anchor=north west][inner sep=0.75pt]  [font=\scriptsize,color={rgb, 255:red, 255; green, 255; blue, 255 }  ,opacity=1 ]  {$\langle \matr{Z}\rangle _{0}$};
\draw (235.03,234.09) node  [color={rgb, 255:red, 255; green, 255; blue, 255 }  ,opacity=1 ]  {$\matr{\statevar}$};

\draw (161.32,-10.8) node [anchor=north west][inner sep=0.75pt]    {$J(\cdot)=\ \left(\max_{t\in [ t_0,T]}\dfrac{1}{N}\sum _{i\in G^{*}} I_{i}( t)\right)^{2}$};

\end{tikzpicture}
            }
    \caption{NODEC architecture for controlling SIR-type dynamics.}
    \label{fig:nn:gnn}
\end{figure}
From a technical perspective, the SIR-type dynamics introduce extra state variables.
Therefore, fully-connected layers will require one to estimate considerably more parameters.
We observe that neither NODEC nor RL converged to a high-performance solution when using fully-connected layers, and we thus omit these results.
Furthermore, the control task requires the network to optimize a loss that is not calculated over whole graph, but rather on a specific sub-graph.
NODEC has no direct information on which nodes are part of sub-graph $\target{G}$.
The information is provided via the minimization of the learning loss-function in \Cref{eq:spread:loss}.
Back-propagation happens at time $\target{t} = \argmax_{t \leq T} J(I_{\target{G}}(t))$.
This time is approximated by preserving a sample of states when using the ODESolve, and picking the maximum observed peak infection from that sample. 

\tomcom{
As the existing neural architectures discussed in \Cref{sec:osc:training} did not perform well, we switch to an architecture that includes the graph structure.
To leverage the information of the graph-structure and generate efficient control signals that ``flatten'' the curve we decide to design a more specialized neural network architecture that includes the information of the graph-structure within its layers.
For that reason we use a Graph Neural Network (GNN) architecture (see \Cref{fig:nn:gnn,appdx:sirx:nn}).
We use a learning rate $\eta = 0.07$ and the Adam optimizer.
The same GNN architecture is implemented in the RL baselines as the policy network. 
GNN encountered fewer numerical instabilities during training and allowed for efficient learning without curriculum procedures.
We use a training procedure for SIR-type control as shown in Supplemental Material~\Cref{algo:trainning} that preserves the best performing model in terms of loss.
}
\tomcom{
The hidden state matrix $\matr{z}$ is calculated from the GNN and then provided as an input to the decision neural network (see \Cref{fig:nn:gnn} right side).
The decision network contains operations that enforce the budget and driver constraints by applying a softmax activation function and calculating control signal outputs for the driver nodes.
The decision network contains no learned parameters and is included inside the NODEC architecture and RL environment.
Transfer learning~\cite{pan2009survey} between NODEC and RL can be achieved by pre-training the GNN network with NODEC and then using it as an RL policy. 
RL achieves the same performance as NODEC when transfer learning is tested.
Further fine tuning of the pre-trained policy with RL does not improve performance of NODEC in this setting, 
but transfer learning indicates a possible future extension of combining model-based training with real-world model-free fine tuning.
}
\begin{figure}
\centering
\subfloat[Infected fraction.]{
\includegraphics[width=0.45\linewidth]{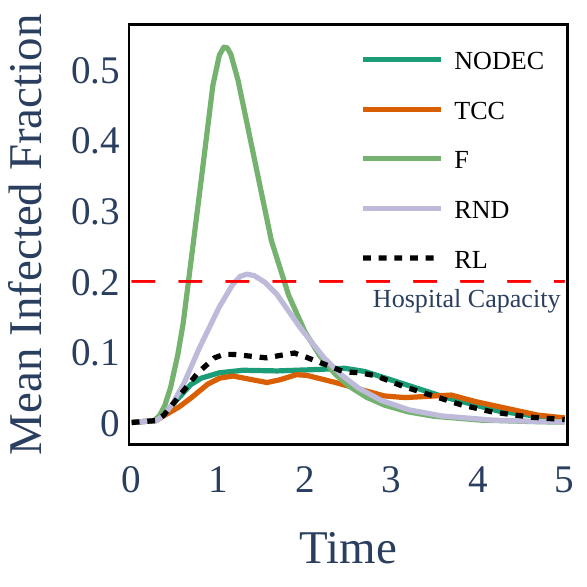}
\label{fig:sirx:inf:fraction}
}\hfill
\subfloat[Total energy.]{
\includegraphics[width=0.45\linewidth]{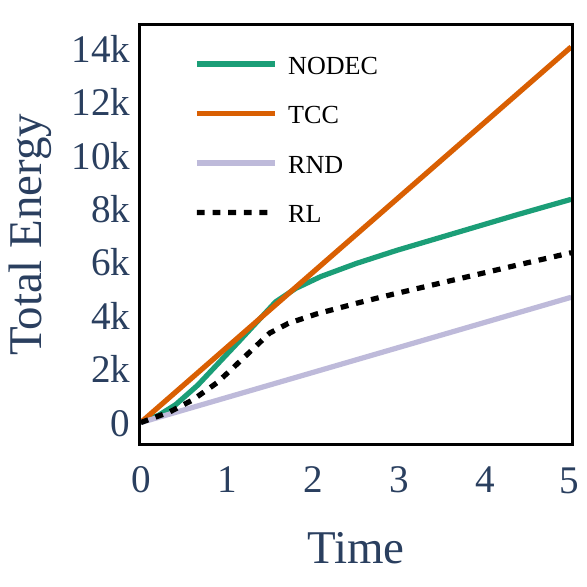}
\label{fig:sirx:energy}
}
\caption{SIR-type control evaluation. NODEC versus baselines: reinforcement learning (RL), targeted constant control (TCC), random constant control (RND), and free dynamics with no control (F).} \label{fig:sirx:eval}
\end{figure}

\subsubsection{Learning Loss}
The control goal is to ``flatten'' the curve, i.e.~to delay and minimize the mean infected-fraction over nodes in the sub-graph $\target{G}$, which has no overlap with the part of the graph containing the initial spreading seed. 
Based on these control goals, we formulate the following loss function:
\begin{equation}\label{eq:spread:loss}
    J(\ts{\statevar}_{t_0}^T, \target{\matr{\statevar}}) = [\max_{t_0 \leq t\leq T}\bar{I}_{\target{G}}(t)]^2,
\end{equation}
where $\bar{I}_{\target{G}}$ denotes the mean fraction of infected individuals in $\target{G}$. 
This goal is macroscopic, as we do not know the exact feasible state \tomcom{$\target{\matr{\statevar}}$ for which $ \target{I}(t^*) = \argmin_{\tv{I}_t}J(I(t))$ that minimizes such loss.
Furthermore the exact time $t^*$ that the minimum loss is achieved is not known, and therefore we need to evaluate samples from the state trajectory $\ts{\statevar}_{t_0}^T$ to determine $t^*$.}
Similar to \Cref{eq:kuramoto:loss}, the current control goal requires loss calculations over a time interval.
Moreover, this loss is not calculated over the whole state matrix $\matr{\statevar}$ but only on the infected state $I_{\target{G}}$ of the target sub-graph.
Intuitively, one would trivially achieve the proposed goal if there are no further constraints. If nodes that connect the sub-graph $\target{G}$ to the rest of the graph cannot be controlled efficiently, then achieving the control goal becomes non-trivial.
Tackling the outlined epidemic control problem allows us to evaluate NODEC on a complex control task (see \Cref{sec:spread:results}) with applications in disease control.

\subsubsection{Results}
\label{sec:spread:results}
\begin{figure*}[!htb]
\centering
\hfill
\begin{minipage}{0.12\textwidth}
    \centering
    \includegraphics[width=1\textwidth]{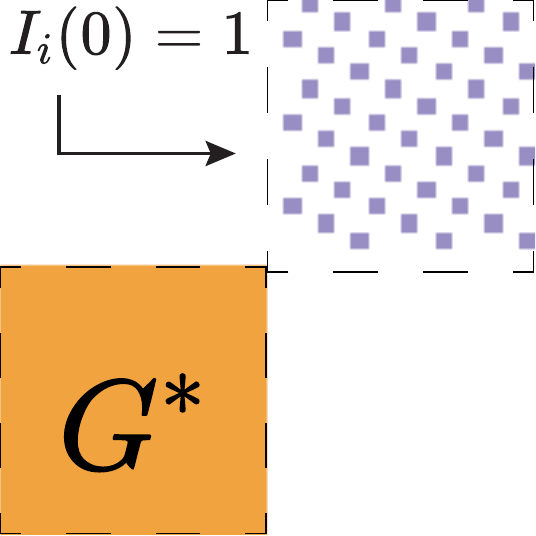}
\end{minipage}
\hfill
\begin{minipage}{0.77\textwidth}
    \includegraphics[width=1\textwidth]{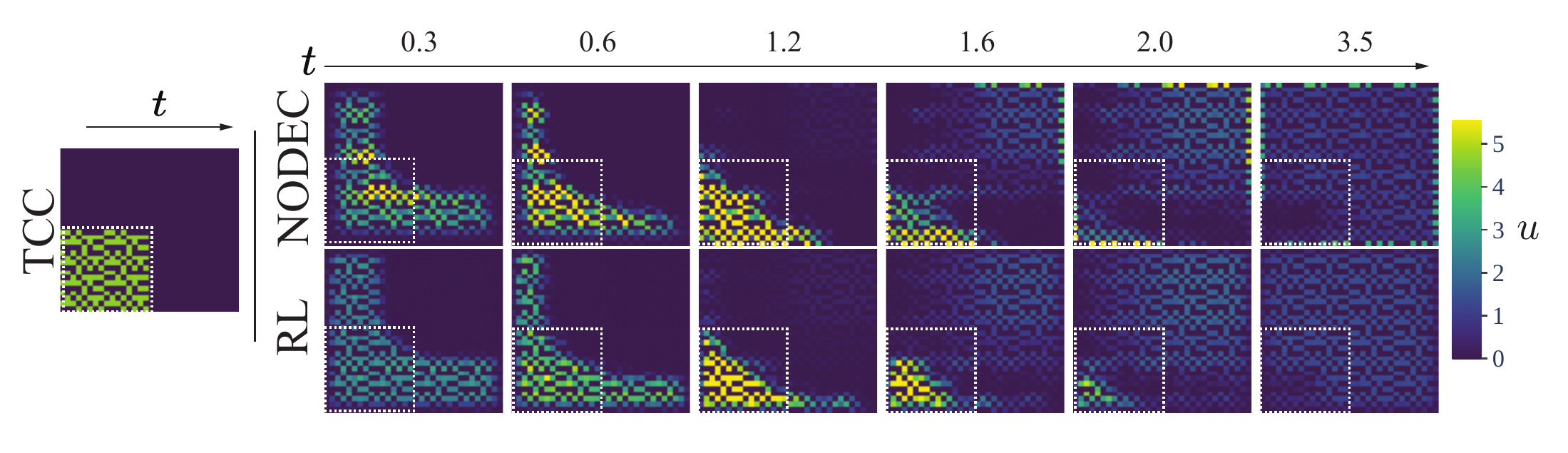}
\end{minipage}
\vspace{-0.3cm}
\caption{Initial infection, target sub-graph, and control trajectories for SIR-type dynamics.
    Colorscale plots represent 99.5\% of the presented values for dynamics with NODEC controls.
}\label{fig:sirx}
\end{figure*}
Our main results are summarized in \Cref{fig:sirx:eval} and \Cref{tab:sir:eval} and indicate similar superior performance of TCC and NODEC compared to the other control strategies, but with lower energy costs for NODEC.
In \Cref{fig:sirx:eval}, we observe that NODEC is providing strong protection with total energy costs that are not as high as TCC (see \Cref{tab:sir:eval}). 
If we assume that the proposed system will reach maximum hospital capacity at $20\%$ of the infected fraction in the target sub-graph, we observe that TCC, RL, and NODEC are sustainable control strategies.
In \Cref{fig:sirx:inf:fraction} and \Cref{tab:sir:eval}, NODEC underperforms TCC with approximately $1\%$ higher maximum infection fraction, but requires almost $41\%$ less control energy.
The effectiveness of the control can be attributed to the adaptive nature of NODEC.
The other adaptive baseline, RL requires around $54\%$ less energy than TCC but allows for $2.1\%$ higher peak infection compared to NODEC.
The effectiveness of targeted adaptive controls in time can be used to model and examine the effectiveness of proposed real-world long-term pandemic control strategies, such as rolling lockdowns~\cite{acemoglu2020optimal} and/or vaccine allocation~\cite{preciado2013optimal}.

NODEC achieves better performance at the cost of higher energy compared to RL.
Reinforcement learning is often described as ``model-free'' and addresses the (i) prediction problem and (ii) control problem~\cite{sutton2018reinforcement}. 
We note that RL approaches may suffer from credit assignment challenges, 
where a reward signal is uninformative regarding the specific actions (especially in terms of time) that help reach the goal~\cite{sutton1985temporal}.
However, even after testing different reward designs and parameters settings, no RL framework managed to perform better than our baselines.
It may be possible that extensive reward engineering, and other model upgrades may lead to a better performance. 
In contrast to RL, the proposed NODEC is not model-free and the underlying gradient descent is directly calculated from the loss function. 
Therefore, we do not need to consider value prediction and credit assignment.
It is possible to design a model-free NODEC by learning the underlying system dynamics simultaneously with control similar to Ref.~\cite{holl2020learning},
which could be an interesting future extension of our work.
\begin{table}[!htb]
\centering
\centering
\footnotesize
\caption{Total energy $E(T)$ and peak infection $\max_t(\bar{I}(t))$ achieved by different epidemic spreading control methods.}
\label{tab:sir:eval}
\begin{tabularx}{\linewidth}{R{0.35}L{0.5}L{0.5}}
\toprule
Control         &   Peak Infection         & Total Energy   \\
\midrule
    TCC &                            0.068 &        14062.6 \\
  NODEC &                            0.078 &         8356.6 \\
     RL &                            0.099 &         6358.0 \\
    RND &                            0.210 &         4688.9 \\
      F &                            0.532 &            0.0 \\
\bottomrule
\end{tabularx}
\end{table}

The spread of the epidemic, target sub-graph, and controls of the main baselines are illustrated in \Cref{fig:sirx}.
\tomcom{
RL and NODEC calculate dynamics controls that change over time and slowly fade out as $t\to T$.
We also observe that controls persist in some driver nodes even then the infection wave is over (see also Supplentary Material \Cref{fig:sirx:rl:states}).
This behavior is also observed in other baselines that satisfy the equality of the constraint~\Cref{eq:sirx:budget} (RND and TCC).
The budget constraint \Cref{eq:sirx:budget} allows control signals sum up to the budget value $\mathcal{b}$.
The implemented NN architecture calculates controls by multiplying the budget with a softmax activation function output over a hidden state output from the learned GNN architecture (see \Cref{fig:nn:gnn} right side). 
The output of the softmax activation function is non-zero by definition\footnote{In practice $0$ values can be generated due to floating point errors.} and thus the NN always calculates non-zero control signals over the driver nodes.
Once the infection wave has traversed the graph, both RL and NODEC controllers spread the control over several nodes, thus decreasing required control energy\footnote{Looking at the control energy \Cref{eq:energy}, we observe that low absolute value control signals assigned over many driver nodes may produce lower energy values compared to very high absolute value control signals applied to fewer driver node.}.
This outcome is an artifact of the softmax activation function, but may also indicate the implicit energy regularization properties of NODEC.
On the contrary, the higher energy costs of TCC keep increasing, as high control signals remain in place after the infection wave has passed.
}

\section{Discussion and Conclusion}
\label{sec:discussion}
Neural ODE control approximates dynamical systems based on observations of the system-state evolution and determines control inputs according to pre-defined target states. Contrary to Ref.~\cite{chen2018neural} that parameterizes the derivative of hidden states using neural networks, our neural-ODE systems describe controlled dynamical systems on graphs. 
In general, neural networks are able to approximate any control input as long as they satisfy corresponding universal approximation theorems. 
However, in practice, NODEC needs to deal with different numerical hurdles such as large losses and stiffness problems of the underlying ODE systems. By testing NODEC on various graph structures and dynamical systems, we provide evidence that these hurdles can be overcome with appropriate choices of both hyperparameters and numerical ODE solvers. 

Future studies may study the effectiveness of NODEC under additional constraints such as partial observability and delayed and noisy controls.

\section*{Acknowledgements}
L.B.~acknowledges financial support from the SNF (P2EZP2\_191888). N.A.-F.~has been funded by the European Program scheme ’INFRAIA-01- 2018-2019: Research and Innovation action’, grant agreement \#871042 ’SoBigData++: European Integrated Infrastructure for Social Mining and Big Data Analytics’. T.A.~received financial support from the LCM – K2 Center within the framework of the Austrian COMET-K2 program.

\section*{Data Availability Statement}
Figures and tables are available within the article and also the Supplemental Material Document.
The experiment data that support the findings of this study are openly available in the NODEC IEEE Dataport repository\cite{nodecdataport} at \url{https://dx.doi.org/10.21227/gdqj-am79}.
The code that fully reproduces the above experiments is found as a  code ocean capsule~\cite{nodecocean} at \url{https://codeocean.com/capsule/1934600/tree}.
A fully functioning code library~\cite{nncgit} written in python for neural network control with NODEC is found at \url{https://github.com/asikist/nnc} with coding examples and more applications.

\bibliographystyle{abbrv}
\bibliography{019_refs}

\clearpage
\newpage
\appendix

\noindent\textbf{\LARGE{Supplemental Material}}
\section{Kuramoto Oscillators}

\subsection{Curriculum Learning}
A curriculum learning procedure is used to train Kuramoto models.
The algorithm is illustrated below in \Cref{algo:training:curiculum}.
\removelatexerror
\small
\begin{algorithm}[!htb]
\small
\SetAlgoLined
\KwResult{$\tv{w}$}
\KwInit: $\tv{\statevar}_0$, $\tv{w}$, $\tv{f}(\cdot)$, $\text{ODESolve}(\cdot)$,  $\text{Optimizer}(\cdot)$, $J(\cdot)$, $\target{\tv{\statevar}}$\;
\KwParams: $\eta$, epochs, stepSize \;
epoch $\gets$ 0\;
$T \gets 0$
\While{epoch $<$ epochs}{
    $t \gets 0$ \;
    $c \sim \mathcal{U}(0,1)$ \;
    $T \gets T + 2 \cdot c$ \;
    $\tv{\statevar} \sim \mathcal{N}_{N}({0}_1^N,{1}_1^N)$\;
    
    meanLoss $\gets$ \KwList\;
    minLoss $\gets \infty$ \;

    \While{$ t < T$}{
        $\tv{\statevar}(t), \text{hasNumInstability} \gets \text{ODESolve}(\tv{\statevar}, t, t + \text{stepSize} , f, \nnc{\tv{u}}(\tv{\statevar}; \tv{w}))$\;
        \If{\KwNot hasNumInstability}{
               meanLoss $\gets (\text{stepSize}/T) \cdot J(\tv{\statevar}(t),\target{\tv{\statevar}})$)\;
               \If{minLoss >  $J(\tv{x_t},\target{\tv{\statevar}})$}{
                    minLoss $\gets$  $J(\tv{\statevar}_t,\target{\tv{\statevar}})$ \;
                    }
        }
        $t \gets t$ + stepSize \;
    }
   Optimizer.update($\tv{w}$,  meanLoss + minLoss )\;
   
 }
\caption{Curriculum training process of NODEC. A procedure that gradually increases total time is introduced in this algorithm. Here we present a stochastic procedure, but a deterministic procedure is also possible.}
 \label{algo:training:curiculum}
\end{algorithm}

\subsection{Synchronization Loss Before Convergence}
\begin{figure}[!htb]
    \centering
    \includegraphics[width=0.6\columnwidth]{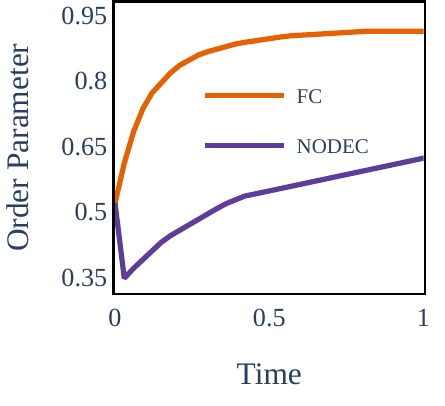}
    \caption{Early order parameter values based on \Cref{fig:osci:loss:all}.}
    \label{appdx:fig:osci:loss:zoom}
\end{figure}

In this section, we describe one of the results presented in \Cref{fig:osci:loss:all} in more detail.
We observe that NODEC takes more time to converge to a synchronized state in the example illustrated in \Cref{appdx:fig:osci:loss:zoom}.
We also observe that NODEC requires a higher amount of control energy before reaching the synchronized state \Cref{fig:osci:energy}.
Once synchronicity is reached, the neural network can adapt and produce lower energy controls.
This might not be the case for feedback control, which has a constant term $\zeta$ multiplied by the driver matrix values.

\section{SIR-type}
\subsection{Neural Network Architecture}\label{appdx:sirx:nn}
\tomcom{
Here, we provide some technical details and an overview of the GNN architecture presented in \Cref{fig:nn:gnn} and complements the code.
The final output of a neural network is a control vector $\nnc{\tv{u}}(\matr{\statevar}(t))$.
The input of the GNN is a tensor $\Psi \in \mathbb{R}^{4 \times N \times \hat{d}}$, where $\hat{d}$ is the maximum degree of the graph.
An element $\Psi_{k,j,i}$ of the tensor represents the $k\text{-th}$ state of the $j\text{-th}$ neighbor of node $i$.
The $j\text{-th}$ neighbor of node $i$ is fixed via any permutation of neighbors prior to training.
The operation that constructs a tensor $\Psi$ from the input state matrix $\matr{\statevar}(t)$ is referred to as ``neighborhood embedding''.
GNN applies an operation for each node that aggregates the state values over all neighboring nodes and produces a hidden state tensor $\matr{h(\Psi)}$.
This hidden state is provided to the consecutive layers, and a hidden state matrix (or embedding) $\matr{Z} \in \mathbb{R}^{4 \times N}$ is calculated, with same dimensions as the input state matrix $\matr{\statevar}$.
This matrix $\matr{Z}$ is provided again as input to the GNN structure described above (see left side  of \Cref{fig:nn:gnn})  and a new tensor $\Psi$ is calculated based on the neighborhood embedding procedure.
Providing the calculated hidden state matrix $\matr{z}$ as an input to the GNN is termed ``message passing''~\cite{xu2018powerful}, and is a typical procedure when training GNNs.
Message passing essentially allows the neural network to calculate a hidden state representation for each node $i$ but also leverage information of non-adjacent neighbors for the calculation after the first repetition.
We observe that allowing the message passing process to repeat 4 times maximizes the performance of the network for the current control task.
For example, in the second iteration of the above procedure, the input tensor $\matr{\Psi}$ of the GNN contains a representation calculated by a functional on an aggregation over all the state $i$ of all adjacent nodes $j'$ of each neighbor $j$ of node $i$, based on the neural network parameters.
In conclusion, the GNN architecture aims to learn a state representation $\matr{Z}$ that can be used to produce efficient control signals that take into account the states of non-neighboring nodes of each driver.
After the last message propagation is finished, the mean over the channels is calculated over the hidden state matrix $\langle \matr{z} \rangle_0$ generating a hidden state vector $\tv{z} \in \mathbb{R}^{N}$. 
}

\subsection{Reinforcement Learning}
\label{sec:appendix:sirx:rl}In this section, we focus on the technical details of the RL baseline we used in the main paper.
Reinforcement learning is often described as ``model-free'' and addresses the (i) prediction problem and (ii) control problem~\cite{sutton2018reinforcement}. We note that RL approaches may suffer from credit assignment challenges, where a reward signal is uninformative regarding the specific actions (especially in terms of time) that help reach the goal~\cite{sutton1985temporal}.
In contrast to RL, the proposed NODEC is not model-free and the underlying gradient descent is directly calculated from the loss function. Therefore, we do not need to consider value prediction and credit assignment.
It is possible to design a model-free NODEC by learning the underlying system dynamics simultaneously with control, which could be an interesting future extension of our work.
Note that a direct performance comparison between RL and NODEC in terms of target loss may be considered unfair especially towards RL methods, unless extensive hyper-parameter optimization is performed beforehand.

We first implement SIR-type dynamics as an RL environment.
The softmax activation function and budget assignment discussed in \Cref{sec:spread:nodec} take place in the environment and RL computes the softmax logit values over all nodes. 
Reinforcement learning is allowed to interact with the environment in a fixed interaction interval $\Delta t = 10^{-2}$, similar to NODEC. A2C and SAC implementations are taken from StableBaselines3\footnote{
\url{https://github.com/DLR-RM/stable-baselines3}
}.
Both implementations were tested for different parameter sets and trained for at least $50000$ steps.
Unfortunately, no implementation was able to ``flatten the curve'' considerably better than random control.
Next, we use the TD3 implementation from Tianshu\footnote{
\url{https://github.com/thu-ml/tianshou}
}, which currently showcases high-speed benchmarks and allows more customization of policy/critic architectures.
The corresponding RL training takes around $17$ seconds per epoch, whereas NODEC takes approximately $5.5$ seconds per epoch.
Neither TD3 or NODEC fully utilized the GPU in terms of computing and memory resources, often staying below $50\%$ of usage, while memory utilization usually was below $10$GB per method.

We show an overview of the hyperparameters that we use to train TD3 in \Cref{tab:rl:params}. For more detailed explanations of these hyperparameters, see Ref.~\cite{fujimoto2018addressing} and the Tianshu documentation\footnote{\url{https://tianshou.readthedocs.io/en/latest/api/tianshou.policy.html?highlight=td3\#tianshou.policy.TD3Policy}}.
Several baseline architectures in RL frameworks are often fully-connected multilayer perceptrons.
Still, we observe that the graph neural network presented in \Cref{fig:nn:gnn} was more efficient in converging rewards in less computation time.
We trained all models for 100 epochs and stored and evaluated the best model.
In SAC and A2C, one training environment was used, whereas TD3 was sampling from two independent environments simultaneously due to its computational speed.

In terms of parameters both the TD3 policy network and NODEC GNN have exactly the same learning parameters (weights), but training is very different, as the gradient flows presented in \Cref{fig:nodec:framework} and \Cref{algo:NODEC,algo:odesolve} cannot happen.
The value function is now used for the calculation of similar gradients by predicting the cumulative reward signal.
We studied several possible reward designs, and in the end we rigorously tested the following rewards:

The first reward signal we tested is calculated based on the mean number of infected nodes belonging to the target sub-graph $\bar{I}_{\target{G}}(t)$ at time $t$:
\begin{equation}\label{eq:reward:norm}
    \rho_{1}(t) = -(\bar{I}_{\target{G}}(t))^2{\Delta t}.
\end{equation}
Although this reward seemingly provides direct feedback for an action, it also leads to several challenges. 
First, it does not necessarily flatten the curve, but it minimizes the overall infection through time. Such a reward could, for instance, potentially reinforce actions that lead to ``steep'' peaks instead of a flattened infection curve, as in practice it minimizes the area under the $I(t)$ curve.
Furthermore, as current containment controls may have effect if applied consistently and in the long term, such reward design suffers from temporal credit assignment, since the reward value depends on a long and varying sequence of actions.
Finally, any actions that happen after the peak infection occurrence will still be rewarded negatively, although such actions do not contribute to the goal minimization.

The next reward 
\begin{equation}\label{eq:reward:sparse}
    \rho_{2}(t)= 
\begin{cases}
    \quad 0                                          &, \text{if } t<T\\
    -(\max_{t\leq T}\bar{I}_{\target{G}}(t))^2   &, \text{otherwise}
\end{cases}
\end{equation}
is designed to overcome the aforementioned shortcomings.
This reward signal is sparse through time, as it is non-zero only at the last step of the control when the infection peak is known. The main property of interest of \Cref{eq:reward:sparse} is that it has the same value as the loss that we used to train NODEC (see \Cref{eq:spread:loss}.
This reward signal also suffers from credit assignment problems.
As the reward is assigned at a fixed time and not as a direct result of the actions that caused it, the corresponding reward dynamics is non-Markovian~\cite{thiebaux2006decision}.
To address challenges caused by rewards with non-Markovian properties, reward shaping\cite{camacho2017non} and recurrent value estimators~\cite{mizutani2004two} can be used.
Furthermore, $n\text{--step}$ methods or eligibility traces can be evaluated if we expect the reward signal to be Markovian but with long and/or varying time dependencies.

\begin{figure*}[!htb]
    \centering
    \subfloat[SIR-type curves for no control (F) baseline.]{
    \includegraphics[width=0.3\linewidth ]{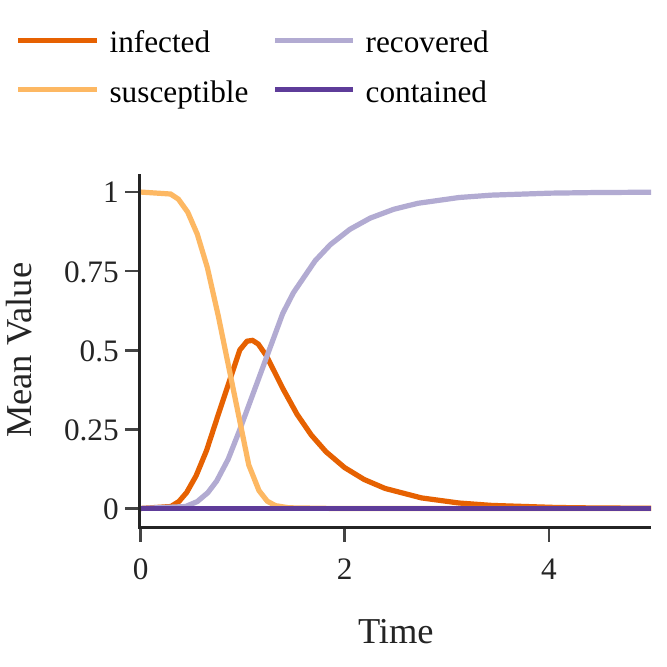}
    \label{fig:sirx:curves:f}
    }
    \hfill
    \subfloat[SIR-type curves for random control (RND) baseline.]{
    \includegraphics[width=0.3\linewidth ]{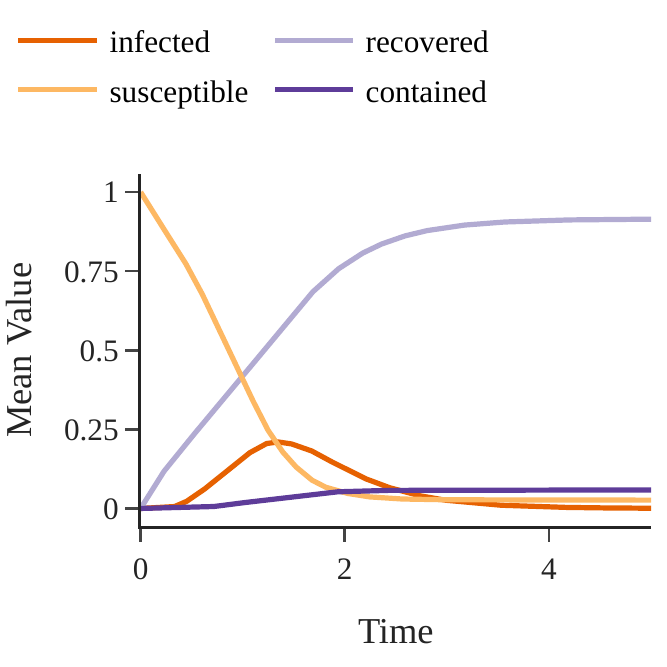}
    \label{fig:sirx:rl:curves:rnd}
    }
    \\
    \subfloat[SIR-type curves for reinforcement learning control (RL) baseline.]{
    \includegraphics[width=0.3\linewidth ]{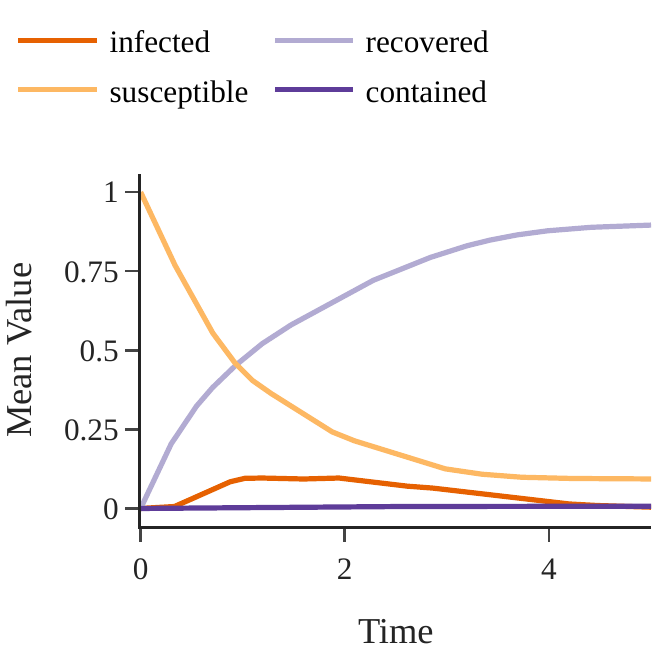}
    \label{fig:sirx:rl:curves:rl}
    }
    \hfill
    \subfloat[SIR-type curves for neural network control (NODEC) baseline.]{
    \includegraphics[width=0.3\linewidth ]{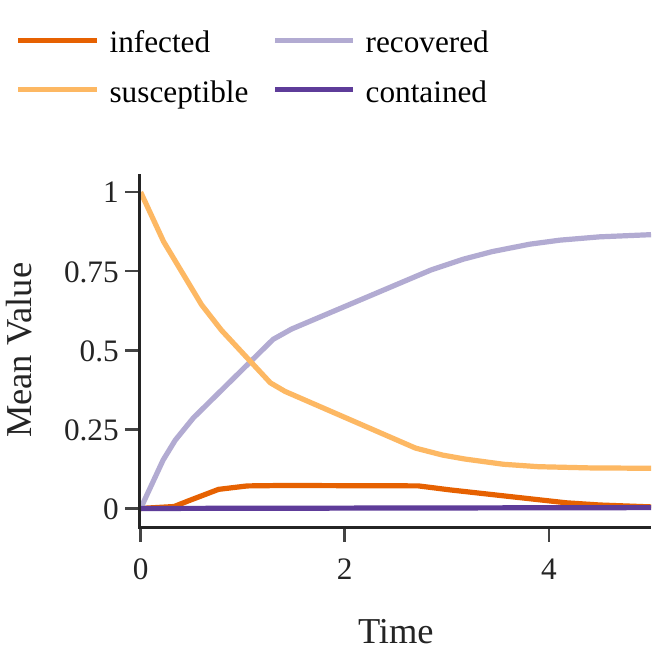}
    \label{fig:sirx:rl:curves:nnc}
    }
    \hfill
    \subfloat[SIR-type curves for targeted constant control (TCC) baseline.]{
    \includegraphics[width=0.3\linewidth]{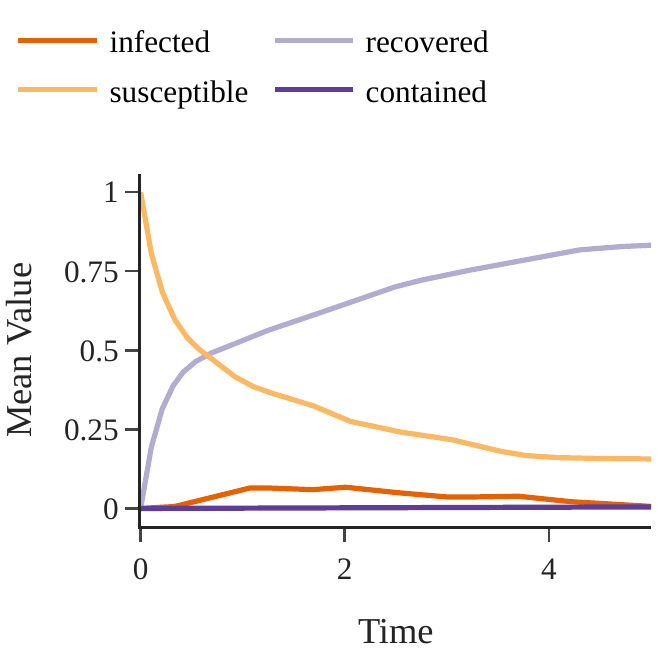}
    \label{fig:sirx:rl:curves:tcc}
    }
    \caption{SIR-type curves for all baselines in the target sub-graph $\target{G}$.}
    \label{fig:sirx:curves}
\end{figure*}

The final reward $\rho_3(t)$ that we evaluated and used in the presented results is designed with two principles in mind:
\begin{subequations}
\begin{align}
    \sum_t \rho_3(t) &\appropto{} \max_{t \leq T} (\bar{I}_{\target{G}}(t))^2 \label{eq:reward3}\\
    \argmin_{t \leq T} \sum_t r_3(t) &= \argmax_{t \leq T}(\bar{I}_{\target{G}}(t))\label{eq:reward3_2}.
\end{align}
\end{subequations}
Following those principles, the reward signal is approximately proportional to and provides information about the value of the infection peak used in the NODEC loss calculation. The reward sum minimizes exactly at the time when peak infection occurs.
This property is expected to reduce effects of temporal credit assignment.
When aiming to replace the proportionality in \Cref{eq:reward3} with an equality, we reach the following reward signal design presented in the main paper \Cref{eq:reward:sum:diff}:
\begin{equation}\label{appdx:eq:reward:sum:diff}
    \rho_{3}(t)= 
\begin{cases}
    0 , \quad\quad \text{if } \bar{I}_{\target{G}}(t) \leq \max_{\uptau < t} (\bar{I}_{\target{G}}(\tau) )\\
    -\bar{I}_{\target{G}}^{2}(t) +(\max_{\tau < t} \bar{I}_{\target{G}}(\tau))^{2}              ,\quad\quad \text{otherwise}
\end{cases}
\end{equation}
It is straightforward to show that \Cref{eq:reward:sum:diff} indeed satisfies $ \sum_t \rho_3(t) = \max_{t \leq T} (\bar{I}_{\target{G}}(t))^2$ and \Cref{eq:reward3_2}.
This reward greatly improved performance without resorting to recurrent value estimators or further reward shaping.
Still, after all proposed reward design and hyper-parameter optimization, NODEC has a higher performance (see \Cref{fig:sirx:curves}), although TD3 performs better than random control.

\begin{figure}[!htb]
    \centering
    \subfloat[Infection spread on lattice for all baselines.]{
    \includegraphics[width=0.9\linewidth ]{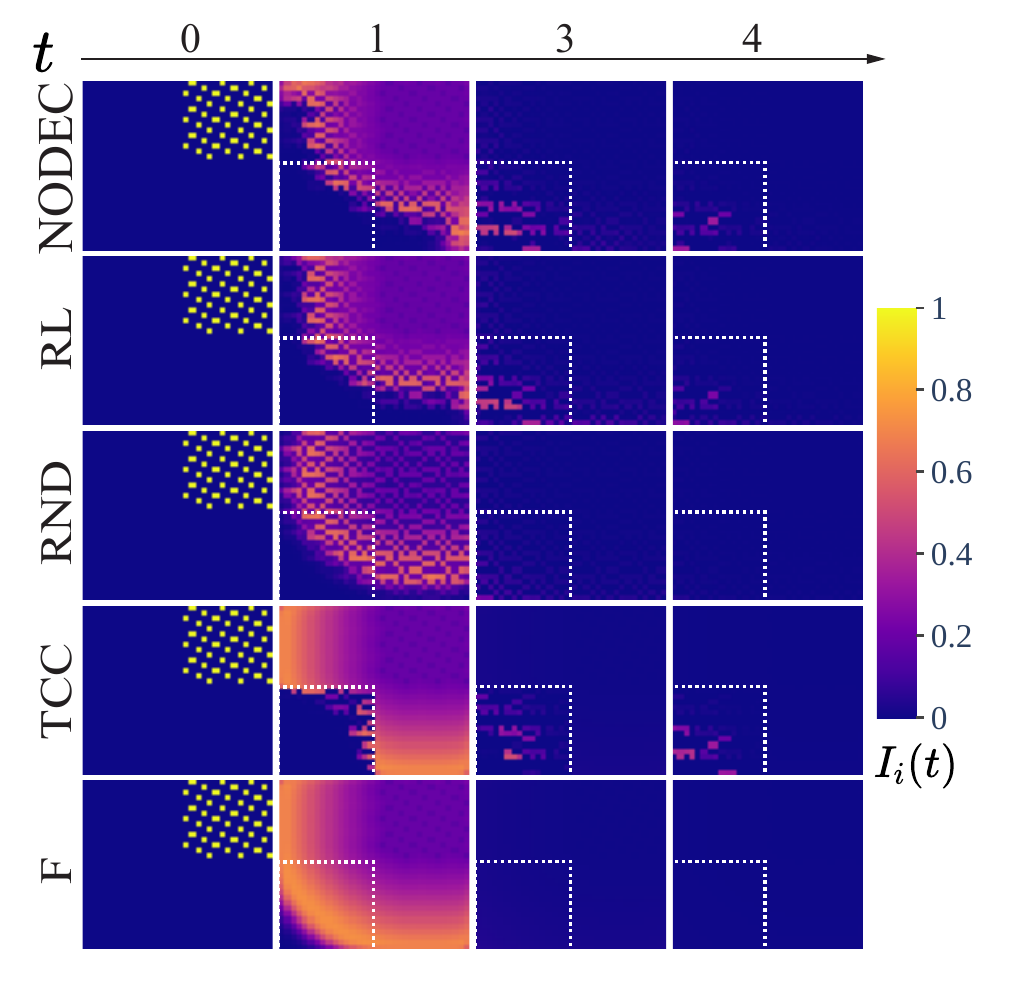}
    \label{fig:sirx:rl:inf}
    }
    \caption{Infection spread from baselines.}
    \label{fig:sirx:rl:states}
\end{figure}

In \Cref{fig:sirx:rl:inf,fig:sirx} the dynamic controls of both RL and NODEC seem to focus on protecting the target sub-graph by containing the infection as it spreads.
In contrast to targeted constant control, they succeed in doing so by protecting driver nodes outside the target sub-graph.
When comparing the dynamic control patterns, the budget allocation of NODEC seems to be much more concentrated on specific nodes, and it creates more often contiguous areas of containment.

In \Cref{fig:sirx:curves}, we also show the evolution of $S(t)$, $R(t)$, and $X(t)$.
We observe that TCC and NODEC show clear signs of flattening the curve by preserving the highest susceptibility fraction and lowest recovery fraction at time $T$, which can be interpreted as less susceptible nodes becoming infected and needing to recover.
The random method outperforms the other frameworks in terms of effective containment fractions, as random control assignments at each time step let the disease spread such that higher infection fractions $I(t)$ are reached in the target sub-graph and therefore drivers with high infection fractions are effectively contained when controlled.
Although low energy effective containment might seem favorable at first sight, it is not optimal in terms of flattening the curve with restricted budget, as it allows high infection fractions to occur within an area of interest.
Budget restrictions often do not allow to fully constrain the spread in all infected nodes.

\begin{figure*}[!htb]
    \centering
    \subfloat[Total episode reward as TD3 trains.]{
    \includegraphics[width=0.47\linewidth ]{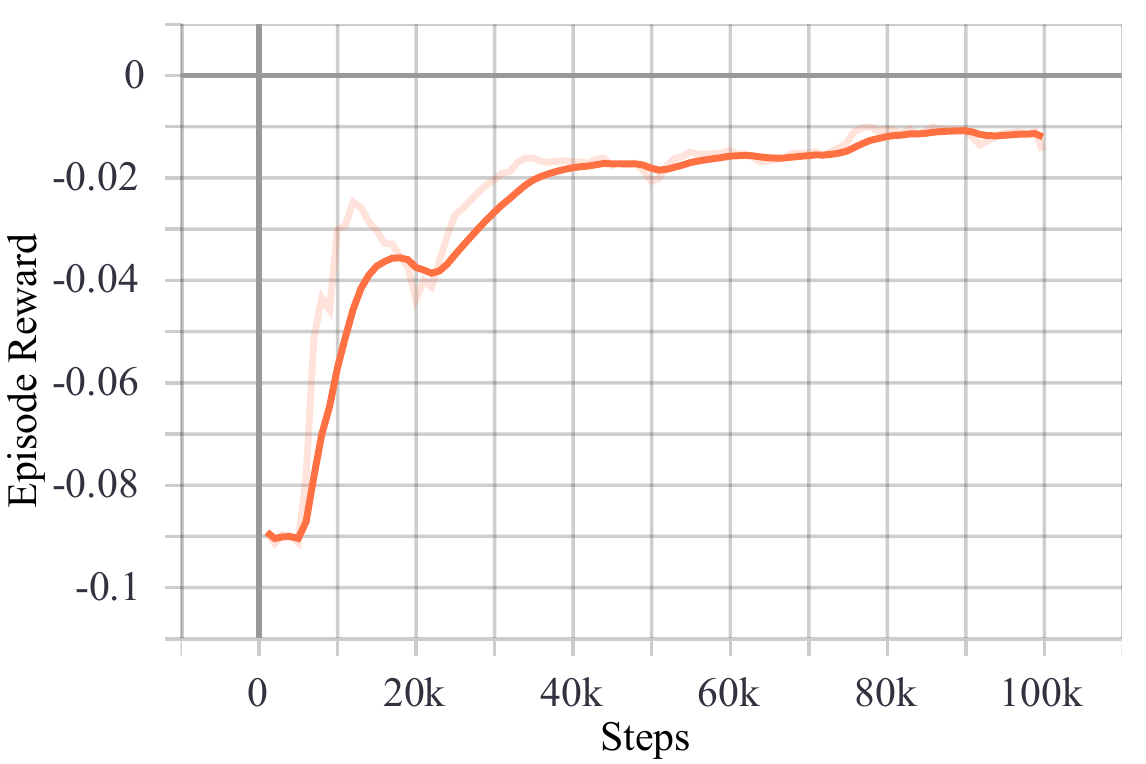}
    \label{fig:sirx:rl:rew}
    }
    \hfill
    \subfloat[Actor loss as TD3 trains.]{
    \includegraphics[width=0.47\linewidth ]{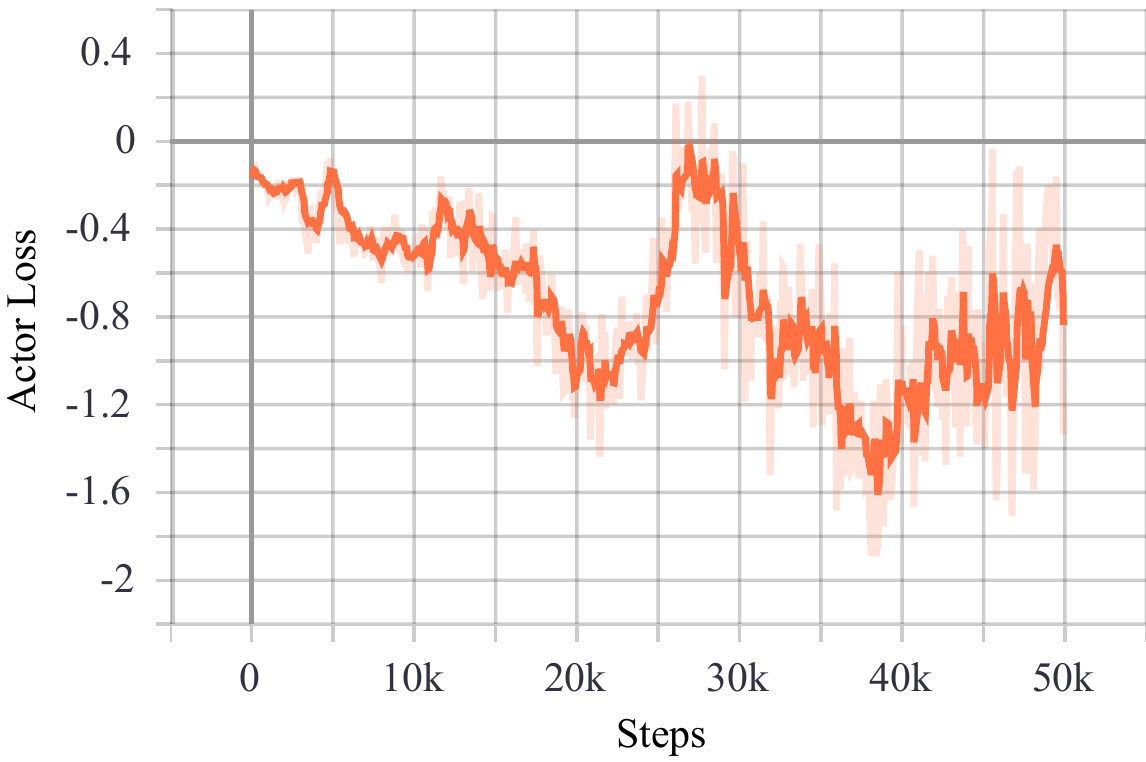}
    \label{fig:sirx:rl:actor}
    }
    \\
    \subfloat[First critic loss.]{
    \includegraphics[width=0.47\linewidth ]{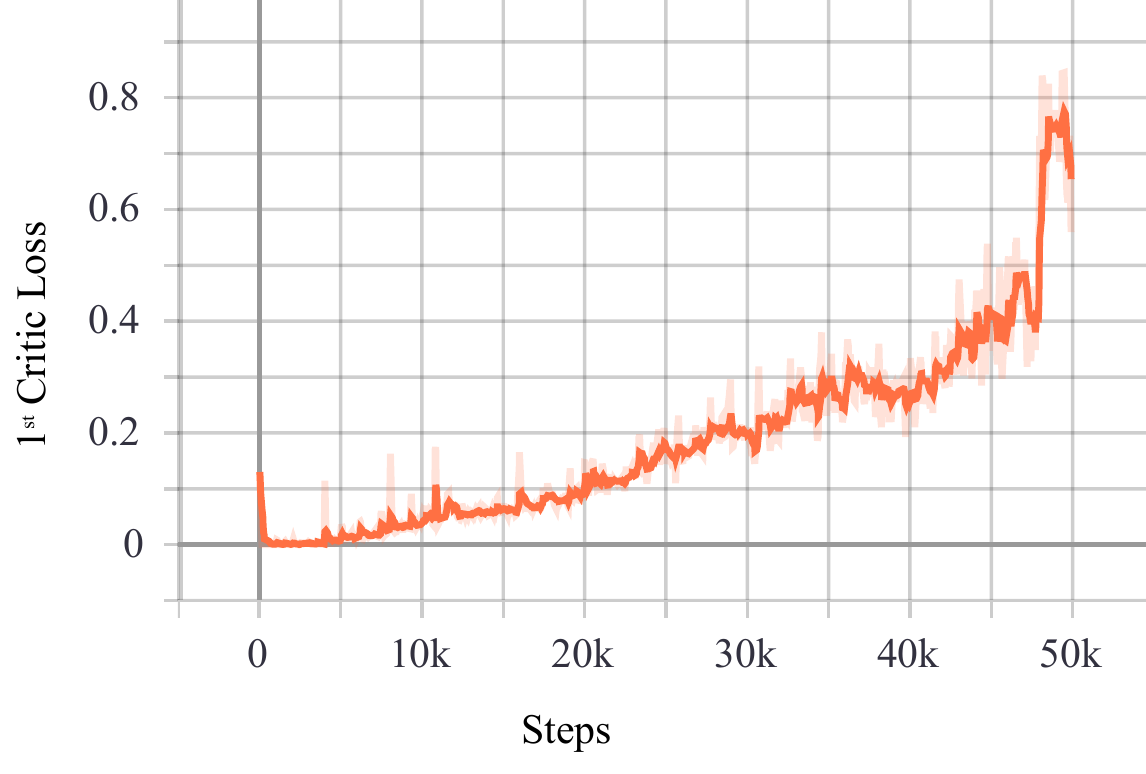}
    \label{fig:sirx:rl:crit:1}
    }
    \hfill
    \subfloat[Second critic loss.]{
    \includegraphics[width=0.47\linewidth ]{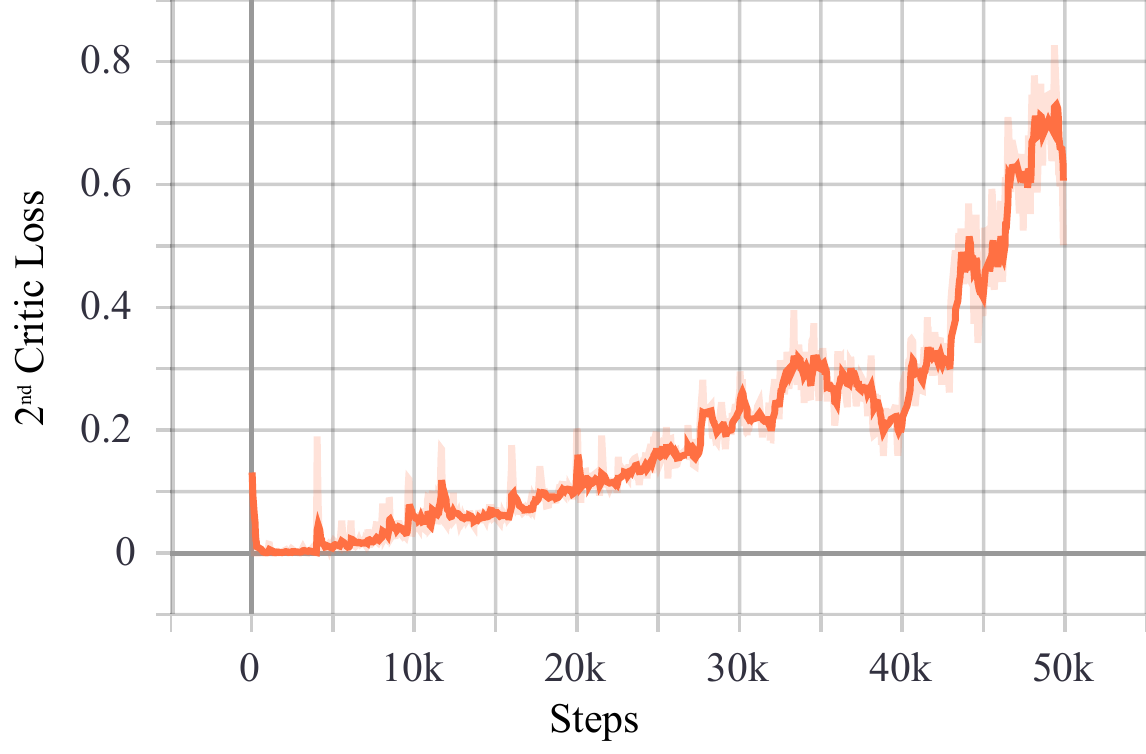}
    \label{fig:sirx:rl:crit:2}
    }
    \caption{RL learning performance evaluation plots using Tensorboard using $0.8$ smoothing.}
    \label{fig:sirx:rl:train}
\end{figure*}

In \Cref{fig:sirx:rl:train}, we observe that although RL does not converge in terms of critic and actor loss, it still converges to a higher reward.
This confirms that RL is capable of controlling continuous dynamics with arbitrary targets, but it requires significant parameterization and training effort to have good stable value estimates.

Finally, we tried to examine transfer learning capabilities from NODEC to RL. A closer look at \Cref{fig:nn:gnn} reveals that the parameterized graph neural architecture used for NODEC and RL can be the same, i.e.~there are no weights in the decision network layers of \Cref{fig:nn:gnn}.
This means that the architectures trained with NODEC can be used as the ``logit'' action policy in RL, showcasing an effective use of transfer learning.
In the given example, the RL policy network starting with trained NODEC parameters, is further trained for $100$ episodes.
After training, RL had a similar performance as NODEC since both methods flatten the curve at approximately $\bar{I}_{\target{G}}=0.0788$.
This means that RL did not improve the solution generated by NODEC.
This example can be used to illustrate the interplay between NODEC and RL and how they can be used in synergy, e.g.~when back-propagating through continuous dynamics is too expensive for high number of epochs.
Reinforcement learning can be used as a meta-heuristic on top of NODEC, and the latter can be treated as an alternative to imitation learning.
\begin{table}[!htb]
\centering
\scriptsize
\caption{Tested and evaluated hyperparameters for the TD3 reinforcement learning baseline.}
\label{tab:rl:params}
\begin{tabularx}{\linewidth}{R{1}L{0.5}L{0.5}}
\toprule
Hyper-Parameter    &  Value  & Tested Values\\
\midrule

Actor learning rate  & 0.0003  & 0.0003, 0.003, 0.03   \\
Actor architecture   &  GNN & GNN, FC \\
Critics learning rate & 0.0001   &  0.0001, 0.001, 0.01   \\
Critics architecture  & FC  & FC  \\
$\uptau$ (Polyak update parameter)   &  0.005 &    0.005,  0.05 \\
$\upgamma$ (discount factor) & 0.99 & 0.5, 0.8, 0.99, 1 \\
 exploration gaussian noise mean & 0.01 & 0, 0.01. 0.1\\
 update frequency of actor parameters & 4 epochs & 1--4 epochs\\
 policy noise & 0.001 & 0.001. 0.01, 0.1\\
 noise clip & 0.5 & 0.5, 0.2 \\
 reward normalization & True & True, False \\
\bottomrule
\end{tabularx}
\end{table}

\section{Other Notes}
\subsection{Hardware and code}\label{sec:hardware}

Our experiments were mainly conducted on a dedicated server that was equipped with an NVIDIA TITAN RTX GPU, 64GB of RAM, and an Intel I9 9900KF 8-core processor.
Partial code tests with assertions were conducted to examine (i) stiffness, (ii) numerical errors or bugs, and (iii) validity and similarity of the same dynamics controlled by different models.
For the majority of the experiments seeds are fixed and initial states parameters are persisted in data files to enable reproducibility.
ODEsolve and sample experiments may be affected by stochasticity on different machines.
Based on statistical testing, we observe that with a good initialization and NN hyperparameter optimization, NODEC performs close to the reported values.
Future works under provided repository, may perform extensive hyperparameter studies dedicated to specific dynamics, graphs.
The average training time of NODEC per task is between 5-10 minutes depending on the complexity of the task.
Baseline methods calculations and parameterizations would also take minutes, making time performance comparable.

The project code can be found on GitHub \url{https://github.com/asikist/nnc} under MIT license. 
Numerical experiments are stored in the experiment folder (please check github {\sc{readme}} for more details).

\subsection{ODE Solvers and Stiffness}\label{sec:solvers}
We prefer to use the Dormand--Prince solver~\cite{shampine2018numerical} for the majority of our numerical experiments (in particular for training). 
For evaluating our results, we use a specific method, which allows the controller to change the control signal at constant time intervals.
This choice allows us to compare control errors and energy costs without considering interaction frequency bias that occurs when one method outperforms another method because the solver allowed it to interact more often with the system and produce more tailored control signals.
Adaptive step length allows the network to learn controls for variant interaction intervals and approximate continuous control better.
We performed small-scale unit tests with VODE~\cite{shampine2018numerical} against Dormand--Prince, Runge--Kutta, and implicit Adams implementations, and we noticed that for most systems numerical errors were negligible.

The goal of this paper is to evaluate the ability of NODEC to learn controls within a solver.
In future works that aim at controlling large-scale systems, different ODE solvers may be chosen according to the system's stiffness and performance requirements of the application.
Whenever dynamics and training had high VRAM requirements, the adjoint method was used, mainly the implementations from Refs.~\cite{chen2018neural,kidger2020hey}.

\subsection{Adaptive Learning Rate Training}\label{sec:appendix:adaptive:train}
Learning rate plays an important role on reaching a low energy control.
In order to determine the optimal learning rate values we propose the adaptive learning rate scheme found in the \Cref{algo:trainning}.
\removelatexerror
\begin{algorithm}[!htb]
\footnotesize
\SetAlgoLined
\KwResult{$\tv{w}$}
\KwInit: $\tv{\statevar}_0$, $\tv{w}$, $\tv(f)(\cdot)$, $\text{ODESolve}(\cdot)$,  $\text{Optimizer}(\cdot)$, $J(\cdot)$, $\target{\tv{\statevar}}$\;
\KwParams: $\eta$, epochs, $\zeta$, tolRatio\;
epoch $\gets$ 0\;
bestLoss $\gets \infty$\;
bestParams $\gets \text{copy}(\tv{w})$\;
previousLoss\;
\While{epoch $<$ epochs}{
    $t \gets 0$ \;
    $\tv{\statevar} \gets \tv{x_0}$\;
    $\ts{\statevar}_{t_0}^T, \text{\scriptsize{hasNumInstability}} \gets \text{ODESolve}(\tv{\statevar}, 0, T, f, \nnc{\tv{u}}(\tv{\statevar}(t); \tv{w}))$\;
    \If{$J(\ts{\statevar}_{t_0}^T,\target{\tv{\statevar}}) > \text{tolRatio} \cdot \text{previousLoss} \lor$  hasNumInstability}{
        \tv{w \gets} bestParams\;
        $\eta \gets \eta\zeta$\;
        Optimizer.reset()\;
        Optimizer.learningRate $\gets \eta$\;
    } 
    \Else{
        \If{$J(\tv{\statevar},\target{\tv{\statevar}}) < $ bestLoss}{
            bestParams $\gets \text{copy}(\tv{w})$\;
            bestLoss $\gets J(\tv{\statevar},\target{\tv{\statevar}})$\;
        }
       previousLoss $\gets J(\tv{\statevar},\target{\tv{\statevar}})$\;
       Optimizer.update($\tv{w}$,$J(\tv{\statevar},\target{\tv{\statevar}})$)\;
   }
 }
\caption{Adaptive Learning rate training process of NODEC.}
\label{algo:trainning}
\end{algorithm}

\section{Nomenclature}
The notation used in this article is summarized in \Cref{tab:nodec:nomenclature:one,tab:nodec:nomenclature:two,tab:nodec:nomenclature:three}.

\begin{table*}[htb!]
    \caption{Nomenclature Part I for \Cref{sec:NODEC}.}
    \label{tab:nodec:nomenclature:one}
    \centering
    \scriptsize
    \begin{tabularx}{\textwidth}{e X}
        $t_0$ & The initial time for control of a dynamical process. Often we may also use $t=0$ without loss of generality.\\
        $T$   & The terminal time for control of a dynamical system.\\
        $\Delta t$ & A finite time difference between an initial and terminal time $\Delta t = t_2-t_1, t_2 > t_1$.\\
        $G(\ts{V},\ts{E})$ & A graph represented as an ordered pair of a set of nodes $\ts{V}$ and a set of edges $\ts{E}$.\\
        $N$ & The number of nodes in a graph $N = |\ts{V}|$.\\
        $\matr{A}$ & The adjacency matrix that represents a graph $G$. It has non zero elements $\matr{A}_{i,j}\neq 0$ if and only if nodes $i,j$ are connected.\\
        $\tv{\statevar}(t)$ & A vector $\tv{\statevar}(t)\in \mathbb{R}^N$, which denotes the state of a dynamical system at time $t$.\\
        $\target{\tv{\statevar}}$ & A vector that denotes the target state of a dynamical system.\\
        $\dot{\tv{\statevar}}(t)$ & Newton's dot notation for differentiation of the system state.\\
        $M$ & The number o`f drivers nodes, i.e. nodes that can be controlled in a graph. As the driver nodes is a subset of all the nodes we have $M\leq N$.\\
        $\tv{f}(t, \tv{\statevar}(t), \tv{u}(\tv{\statevar}(t)))$ & The system evolution function that denotes the dynamic interactions between nodes and drivers when calculating the state derivative.\\
        $\tv{u}(\tv{\statevar}(t))$ & A feedback control signal function $\tv{u}(\tv{\statevar}(t)): \mathbb{R}^N\to\mathbb{R}^M$ calculated based on the system state at time $t$.\\
        $\matr{B}$ & A driver matrix $\matr{B}\in \mathbb{R}^{N\times M}$, where $\matr{B}_{i,m}=1$ if node $i$ is the $m\textrm{-th}$ Driver node and receives a control signal $u_{m}(t)$.\\
        $E\left(\tv{u}(\tv{\statevar}(t))\right)$ & The total energy value of a control signal calculated from $t_0$ until time $t$.\\
        $\hat{\tv{u}}(\tv{\statevar}(t))$ & A control signal value calculated from NODEC.\\
        $\tv{w}$ & Vector with neural network parameters for NODEC.\\
        $\ts{\statevar}_{t_0}^T$ & The state trajectory between $t_0$ and $T$. An ordered set of state vectors $\tv{\statevar}(t), t\in[t_0,T]$.\\
        $J(\ts{\statevar}_{t_0}^T,\target(\tv{\statevar});\tv{w}$ & Learning and control objective function for NODEC. In the current work, we evaluate control goals $\target(\tv{\statevar})$ That are achieved over a system trajectory $\ts{\statevec}_{t_0}^T$.\\
        $\Delta \tv{w}$ & Gradient descent update for neural network parameters.\\
        $\eta$ & Learning rate hyper-parameter for gradient descent.\\
        $h(t)$ & The hidden state evolution function used in the neural ODE paper.\\
        $\text{ODESolve}(\tv{\statevar}(t), t, T, f, u)$ & The function that denotes a numerical ODE solving scheme.\\
    \end{tabularx}
\end{table*}

\begin{table*}[htb!]
    \caption{Nomenclature Part II (Coupled Oscillators) for \Cref{sec:osc}.}
    \label{tab:nodec:nomenclature:two}
    \centering
    \scriptsize
    \begin{tabularx}{\textwidth}{E X}
      $\omega_i$ & Natural frequency for oscillator (node) $i$\\
      $K$ & Coupling constant\\
      $\mathcal{h}(\statevar_i-\statevar_j)$ & Trigonometric function that couples oscillators. Often the sinus function is used, s.t. $h(\cdot) = \sin (\cdot)$.\\
      $\tv{\statevar}^\diamond$ & Synchronized steady state of coupled oscillator system.\\
      $L^\dagger$ & Pseudo-inverse of the graph Laplacian matrix of $G$.\\
      $\tv{b}^{(\textrm{FC})}$ & The feedback control gain vector for the FC baseline.\\
      $r(t)$ & Order parameter, which denotes the synchronization of coupled oscillators.\\
      $\zeta$ & Scaling parameter for feedback control baseline.\\
      $\tau$ & Discrete timestep size for discretizing the time period.\\
      $\Xi$ & Number of time timesteps for discretizing the time period $[0,T]$.\\
      $\xi$ & Timestep index, used to calculate discretized approximations of continuous time metrics.\\
      $r_{\textrm{NODEC}(t)}$ & Order parameter value achieved under NODEC control at time $t$.\\
      $E_{\textrm{NODEC}(t)}$ & Total energy value achieved under NODEC control at time $t$.\\
      $r_{\textrm{FC}(t)}$ & Order parameter value achieved under feedback control baseline at time $t$.\\
      $E_{\textrm{FC}(t)}$ & Total energy value achieved under feedback control baseline control at time $t$.\\
    \end{tabularx}
\end{table*}

\begin{table*}[htb!]
    \caption{Nomenclature Part III (Disease Spreading) for \Cref{sec:spread}.}
    \label{tab:nodec:nomenclature:three}
    \centering
    \scriptsize
    \begin{tabularx}{\textwidth}{E X}
    $S_i(t)$ & Susceptible fraction of individuals in node $i$ at time $t$.\\
    $I_i(t)$ & Infected fraction of individuals in node $i$ at time $t$.\\
    $R_i(t)$ & Recovered fraction of individuals in node $i$ at time $t$.\\
    $Y_i(t)$ & Contained fraction of individuals in node $i$ at time $t$.\\
    $\matr{\statevar}(t)$ & The matrix representation of the state including the state vectors $\tv{S},\tv{I},\tv{R},\tv{S}$ as rows and the node index as columns.\\
    $\target{G}$ & Target sub-graph, i.e. the subset of nodes that we are interested to reduce the peak infection.\\
    $\beta$ & Infection rate parameter.\\
    $\gamma$ & Recovery rate parameter.\\
    $c_j$ & A number sampled from a unoform distribution $c_j \sim \mathcal{U}(0,1)$ to calculate random control.\\
    $\mathcal{b}$ & Control budget. A linear constraint on maximum total control that can be applied on the graph at time $t$.\\
    $\rho(t)$ & Reward signal for reinforcement learning techniques.\\
    $d_i$ & The degree of a node $i$ of the graph.\\
    $\hat{d}$ & The maximum degree value of the graph.\\
    $\Psi$ & Input tensor for convolutional neural network of the GNN.\\
    $\matr{z}$ & Output of hidden layers to be used for message propagation in the graph neural network.\\
    \end{tabularx}
\end{table*}

\end{document}